\documentclass[final]{cvpr}

\usepackage{times}
\usepackage{epsfig}
\usepackage{graphicx}
\usepackage{amsmath}
\usepackage{amssymb}
\usepackage[ruled,vlined]{algorithm2e}

\usepackage{subfig}
\usepackage{placeins}
\usepackage{enumitem}

\usepackage[page]{appendix}

\usepackage{xcolor}
\usepackage{verbatim}
\usepackage{placeins}
\usepackage[pagebackref=true,breaklinks=true,colorlinks,bookmarks=false]{hyperref}

\def\cvprPaperID{4836} 
\def\confYear{CVPR 2021}

\newcommand\imgsize{0.22}
\newcommand*\samethanks[1][\value{footnote}]{\footnotemark[#1]}

\SetCommentSty{mycommfont}

\SetKwInput{KwInput}{Input}                
\SetKwInput{KwOutput}{Output}              

\begin{document}

\title{Rethinking Style Transfer: From Pixels to Parameterized Brushstrokes}


\author{Dmytro Kotovenko\thanks{Both authors contributed equally to this work.} \hspace{15pt} 
Matthias Wright\samethanks \hspace{15pt} 
Arthur Heimbrecht \hspace{15pt}
Björn Ommer \\
IWR,
Heidelberg Collaboratory for Image Processing,
Heidelberg University \\
}

\maketitle

\begin{abstract}
There have been many successful implementations of neural style transfer in recent years.
In most of these works, the stylization process is confined to the pixel domain.
However, we argue that this representation is unnatural because paintings usually consist of brushstrokes rather than pixels. \\
We propose a method to stylize images by optimizing parameterized brushstrokes instead of pixels and further introduce a simple differentiable rendering mechanism. \\
Our approach significantly improves visual quality and enables additional control over the stylization process such as controlling the flow of brushstrokes through user input. \\
We provide qualitative and quantitative evaluations that show the efficacy of the proposed parameterized representation. Code is available at \url{https://github.com/CompVis/brushstroke-parameterized-style-transfer}.
\end{abstract}

\section{Introduction}\label{sec:intro}

\begin{figure}
    \centering
    \def\arraystretch{0.0}
    \setlength{\tabcolsep}{1.0pt}
    \begin{tabular}{c}
    \includegraphics[width=0.49\textwidth]{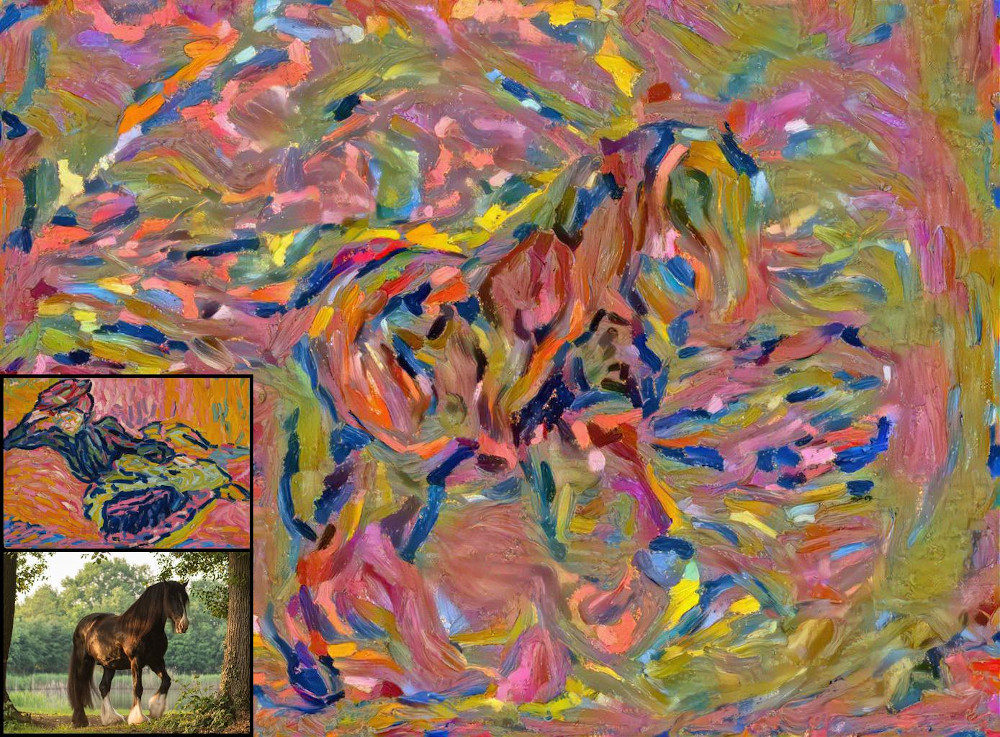} \\[1mm]
    \includegraphics[width=0.49\textwidth]{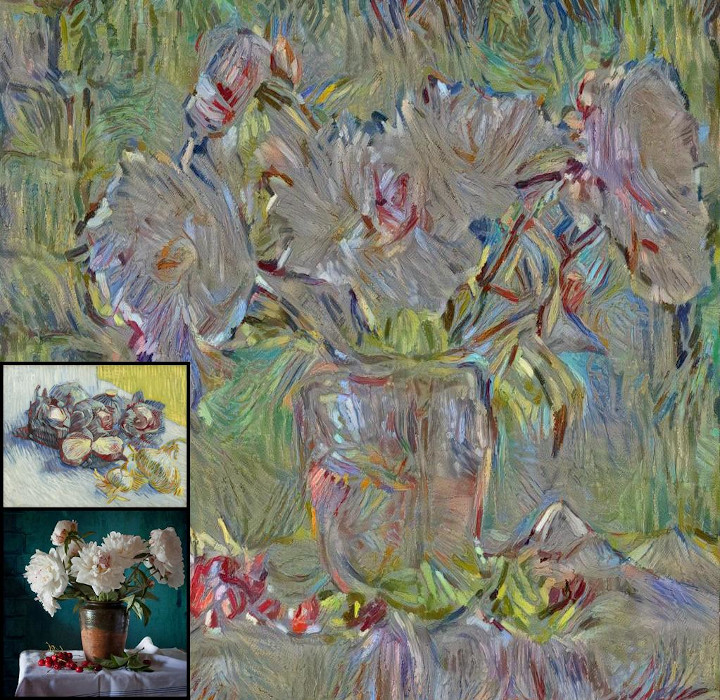} \\
    \end{tabular} \\
    \caption{Stylization results. Top artwork: ``Girl on a Divan'' by Ernst Ludwig Kirchner. Bottom artwork: ``Red Cabbages and Onions'' by Vincent van Gogh.}
    \label{fig:self_portrait_reconstruct}
\end{figure}

Style and texture transfer have been research topics for decades \cite{Hertzmann2001B, Efros2001}.
More recently, the seminal work by Gatys et al. \cite{Gatys2016} reformulated style transfer as the synthesis of an image combining content of one image with style of another image. Since then, a plethora of approaches have explored different aspects of the original problem. There are papers on feed-forward architectures \cite{Johnson2016, Ulyanov2016},  universal feed-forward models \cite{Huang2017, Li2017_a, Li2018ECCV, Li2019CVPR}, disentanglement of style and content \cite{Sanakoyeu2018, Kotovenko2019, Kotovenko2019_a}, ultra-resolution models \cite{Wang2020CVPR}, meta-learning techniques \cite{Shen2018, Zhang2019}, and video style transfer \cite{Chen2020ECCV}. Yet, the initial approach suggested by Gatys et al. \cite{Gatys2016} remains one of the best in terms of image quality, especially in the artistic style transfer scenario, with one style image and one content image.

Recent works have advanced the field of style transfer and produced impressive results by introducing novel losses \cite{Li2017, Risser2017Arxiv, Sanakoyeu2018}, adopting more suitable architectures \cite{Johnson2016, Ulyanov2016, Huang2017, Li2017_a}, imposing regularizations on the final image and intermediate latent representation \cite{Sanakoyeu2018, Kotovenko2019, Kotovenko2019_a, Svoboda2020CVPR}, and even using different training paradigms \cite{Shen2018, Zhang2019}. 
However, they share a key commonality: the stylization process is confined to the pixel domain, almost as if style transfer is a special case of image-to-image translation \cite{Isola2017CVPR, Zhu2017ICCV, Wang2018CVPR, Liu2018NIPS, Huang2018ECCV, Liu2019ICCV, Choi2018CVPR, Choi2020CVPR, Kim2020ICLR}. We argue that the pixel representation is unnatural for the task of artistic style transfer: artists compose their paintings with brushstrokes, not with individual pixels. While position, color, shape, placement and interaction of brushstrokes play an important role in the creation of an artwork, small irregularities appearing on the pixel level like bristle marks, canvas texture or pigments are to some extent arbitrary and random.

With this in mind, we take a step back and rethink the original approach by suggesting a representation that inherently aligns with these characteristics by design. \\
Just like learning to walk in the reinforcement learning setting starts with defining the set of constraints and degrees of freedom for individual joints, we restrict our representation to a collection of brushstrokes instead of pixels. Specifically, we parameterize a brushstroke with a B\'{e}zier curve and additional parameters for color, width, and location. \\
To map these parameterized brushstrokes into the pixel domain we propose a lightweight, explicit, differentiable renderer which serves as a mapping between brushstroke parameters and pixels. Thus, this reparameterization can be seamlessly combined with other style transfer approaches.
One crucial property that this rendering mechanism offers is a spatial relocation ability of groups of pixels. Standard optimization on the pixel level cannot directly move pixels across the image - instead it dims pixels in one area and highlights them in another area. Our model, however, parameterizes brushstrokes with location and shape, thus moving brushstrokes becomes a more natural transformation.

We validate the effectiveness of this reparameterization by coupling the renderer with the model Gatys et al. \cite{Gatys2016} have suggested, see Fig. \ref{fig:approach_diagram}. 
We show that this simple shift of representation along with our rendering mechanism can outperform modern style transfer approaches in terms of stylization quality. This is measured using 1) the deception rate - how similar is the stylized image to the style of an artist 2) human deception rate - whether a human subject can distinguish cropouts of real artworks from cropouts of our stylization. In addition, we illustrate that the brushstroke representation offers more control. A user can control brushstrokes, change the flow of strokes in a neighbourhood.

We further conduct experiments on reconstructions of an image using our rendering mechanism. Huang et al. \cite{Huang2019} train a neural network that successively fits colored quadratic B\'{e}zier curves (brushstrokes) that approximate a target image. Our renderer can be applied to this task as well. It achieves almost 2 times smaller mean squared error (MSE) in the pixel space for a large number of strokes (1000 strokes) and 20\% smaller MSE using 200 strokes.

\begin{figure*}
    \centering
    \def\arraystretch{0.0}
    \setlength{\tabcolsep}{1.0pt}
    \begin{tabular}{ccc}
    \includegraphics[width=0.22\textwidth]{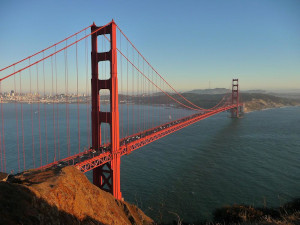} &
    \includegraphics[width=0.32\textwidth]{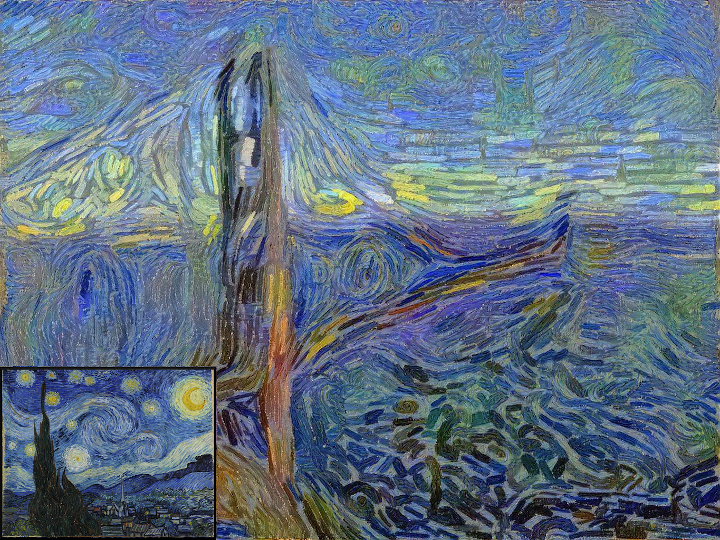} &
    \includegraphics[width=0.32\textwidth]{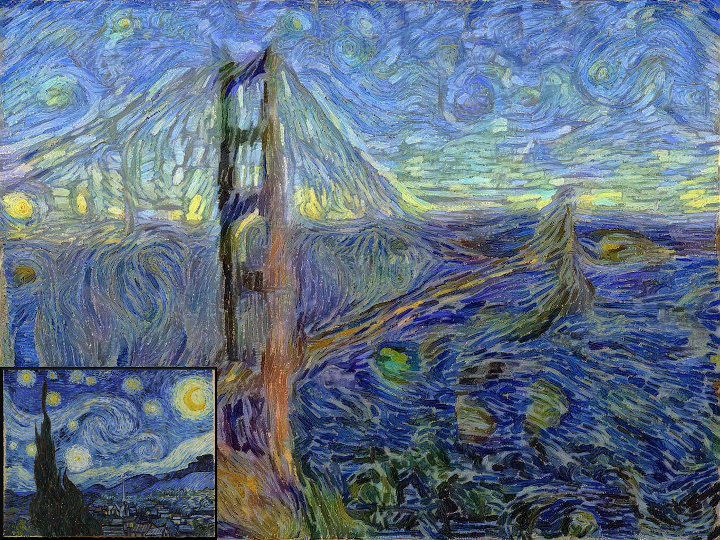} \\
    \vspace{2pt} \\
    (a) Content & (b) 2000 Brushstrokes & (c) 5000 Brushstrokes \\
    \vspace{2pt} \\
    \includegraphics[width=0.22\textwidth]{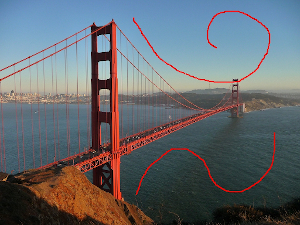} &
    \includegraphics[width=0.32\textwidth]{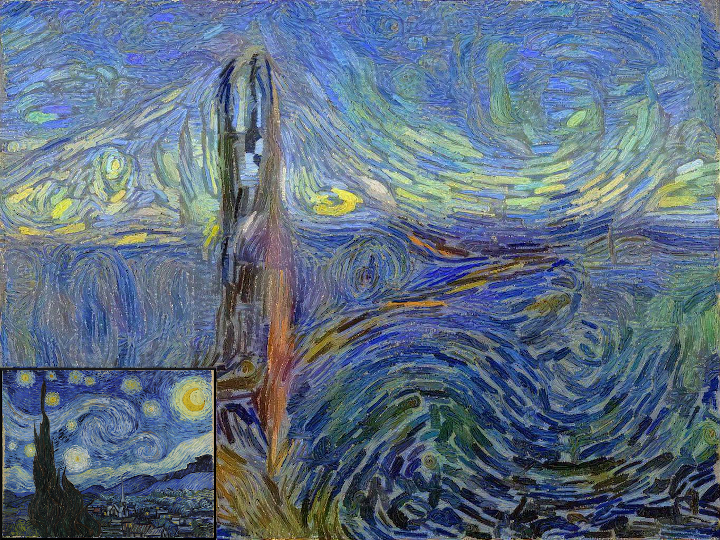} &
    \includegraphics[width=0.32\textwidth]{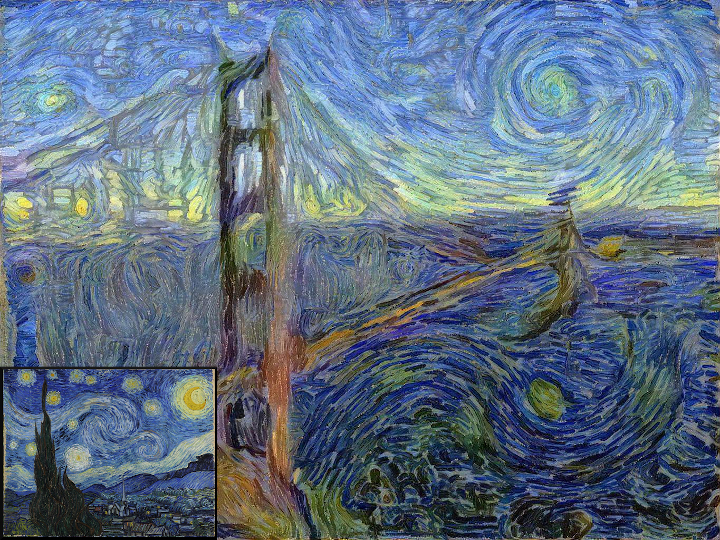} \\
    \vspace{2pt} \\
    (d) Content with User Input & (e) 2000 Brushstrokes and User Input & (f) 5000 Brushstrokes and User Input \\
    \end{tabular} \\
    
    \caption{A user can draw curves on the content image and thus control the flow of the brushstrokes in the stylized image. Note that for the stylizations with user input we also used (a) as content image. The control is imposed on the brushstroke parameters, not the pixels. Images in the middle column are synthesized using 2000 brushstrokes and images in the right column are synthesized with 5000 brushstrokes. See Sec. \ref{sec:drawing} and \ref{sec:add_results} for more experiments.}
    \label{fig:user_control}
\end{figure*}

\section{Related Work}\label{sec:related_work}

\textbf{Style Transfer.}
Initially, Efros and Freeman \cite{Efros2001} performed texture synthesis and transfer using image quilting and Hertzmann et al. \cite{Hertzmann2001B} used a pair of images - one being a filtered version of the other - to learn a filter, which can then be applied to a new image. Wang et al. \cite{Wang2004} introduced a method for synthesizing directional textures. Besides that, there are works studying shape and morphology of images \cite{Yarlagadda2012, Monroy2011, Monroy2014}. \\
More recently, Gatys et al. \cite{Gatys2016} proposed an iterative method for combining the content of one image with the style of another by jointly minimizing content and style losses, where the content loss compares the features of a pretrained VGG network \cite{Simonyan2015} and the style loss compares the feature correlations as given by the Gram matrices.

Several works \cite{Johnson2016, Ulyanov2016} have proposed feed-forward networks to approximate the optimization problem posed by Gatys et al. \cite{Gatys2016} for a fixed style image. \\
Li et al. \cite{Li2017} showed that matching the Gram matrices of feature maps corresponds to minimizing the Maximum Mean Discrepancy with the second order polynomial kernel and also proposed alternative style representations to the Gram matrix such as mean and variance. Dumoulin et al. \cite{Dumoulin2017} introduced conditional instance normalization, which enables the model to learn multiple styles. Huang and Belongie \cite{Huang2017} performed arbitrary real-time style transfer by training a feed-forward network to align the channel-wise mean and standard deviation of the VGG features of a content image to match those of a given style image. Li et al. \cite{Li2017_a} extend this approach by replacing the moment matching between the encoder and decoder with whitening and colouring transformations.

Li et al. \cite{Li2018ECCV} propose a closed-form solution for photorealistic image stylization and Li et al. \cite{Li2019CVPR} learn linear transformations for fast arbitrary style transfer. Sanakoyeu et al. \cite{Sanakoyeu2018} and Kotovenko et al. \cite{Kotovenko2019} propose a style-aware content loss, which also has been used for disentanglement of style and content \cite{Kotovenko2019_a}.

Another line of work draws on meta learning to handle the trade-off between speed, flexibility, and quality \cite{Shen2018, Zhang2019}. \\
Wang et al. \cite{Wang2020CVPR} incorporate model compression to enable ultra-resolution style transfer \cite{Wang2020CVPR}, Xia et al. perform photorealistic style transfer using local affine transforms \cite{Xia2020ECCV}, Chang et al. \cite{Chang2020ECCV} employ domain-specific mappings for style transfer, Chiu and Gurari \cite{Chiu2020ECCV} propose an iterative and analytical solution to the style transfer problem, and Kim et al. \cite{Kim2020ECCV} suggest a method for deformable style transfer that is not restricted to a particular domain.
Yim et al. \cite{Yim2020ECCV} introduce filter style transfer, Wang et al. \cite{Wang2020CVPR_a} propose deep feature perturbation, Svoboda et al. \cite{Svoboda2020CVPR} perform style transfer with a custom graph convolutional layer, and Chen et al. \cite{Chen2020ECCV} employ optical flow to stylize videos.

\textbf{Stroke Based Rendering.}
Stroke based rendering (SBR) aims to represent an image as a collection of parameterized strokes or other shapes that can be explicitly defined by a finite set of parameters. In accordance with other non-photorealistic rendering techniques, the goal is not to reconstruct but rather to render the image into an artistic style. Early works include an interactive method by Haeberli \cite{Haeberli1990}, where the program follows the cursor across the canvas, obtains a color by point sampling the source image, and then paints a brush of that color. Hertzmann \cite{Hertzmann1998} extended this line of research by proposing an automated algorithm that takes a source image and a list of brush sizes, and then paints a series of layers, one for each brush size, on a canvas in order to recreate the source image with a hand-painted appearance. Similar approaches employ segmentation \cite{Gooch2002} or relaxation \cite{Hertzmann2001}. SBR methods are not constrained to static images and have also been used to transform ordinary video segments into animations that possess a hand-painted appearance \cite{Litwinowicz1997}. \\
\textbf{Brush Stroke Extraction}.
Conversely to SBR methods, there have been attempts to detect and extract brush strokes from a given painting. These methods generally utilize edge detection and clustering-based segmentation \cite{Li2012} or other classical computer vision techniques \cite{Berezhnoy2009, Putri2017} and have been used to analyze paintings. \\
\textbf{Drawing Networks.}
Recent work relies on neural networks to predict brush stroke parameters that approximate a given image, using a variety of architectures and training paradigms. These range from supervised training of feed-forward and recurrent architectures \cite{Ha2018, Zheng2019, Nakano2019} to deep reinforcement learning, using recurrent \cite{Ganin2018, Jia2019, Mellor2019} and feed-forward models \cite{Huang2019}. Note that our work is orthogonal to this line of research because we focus on performing style transfer on the level of parameterized brushstrokes.

\section{Background}\label{sec:background}
In the original style transfer formulation, Gatys et al. \cite{Gatys2016} propose an iterative method for combining the content of one image with the style of another by jointly minimizing content and style losses.
The content loss is the Euclidean distance between the rendered image $I_r$ and the content image $I_c$ in the VGG feature space:

\begin{equation}\label{eq:content_loss}
    \mathcal{L}_{\textnormal{content}} = ||\phi_l(\mathit{I}_{\textnormal{r}}) - \phi_l(\mathit{I}_{\textnormal{c}})||_2,
\end{equation}

where $\phi_l(\cdot)$ denotes the $l$-th layer of the VGG-19 network. The style loss is defined as:

\begin{equation}\label{eq:style_loss}
    \mathcal{L}_{\textnormal{style}} = \sum\limits_{l=0}^L w_l E_l
\end{equation}

with

\begin{equation}
    E_l = \frac{1}{N_l^2 M_l^2} ||G_r^l - G_s^l||_F
\end{equation}

where $G_r^l$ and $G_s^l$ are the Gram matrices of $\mathit{I}_{\textnormal{r}}$ and $\mathit{I}_{\textnormal{c}}$ respectively, computed from the $l$-th layer of the VGG-19 network.

\section{Approach}\label{sec:approach}

\begin{figure}
    \centering
    \def\arraystretch{0.0}
    \setlength{\tabcolsep}{1.0pt}
    \begin{tabular}{cc}
    Gatys et al. & Ours\\
    \vspace{1pt} & \\
    \includegraphics[width=0.23\textwidth]{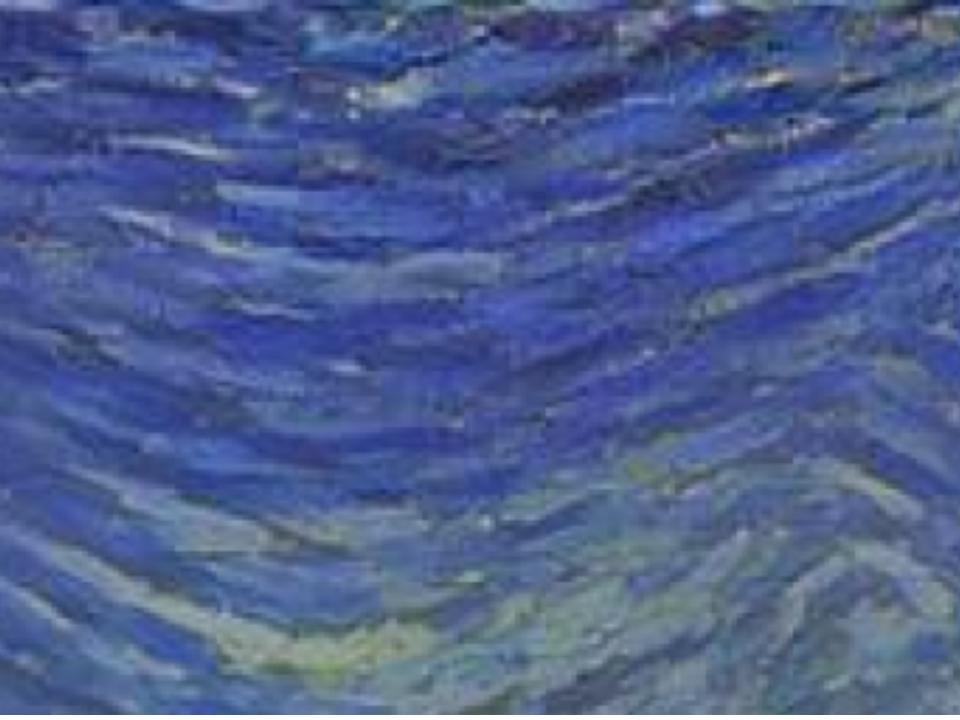} &
    \includegraphics[width=0.23\textwidth]{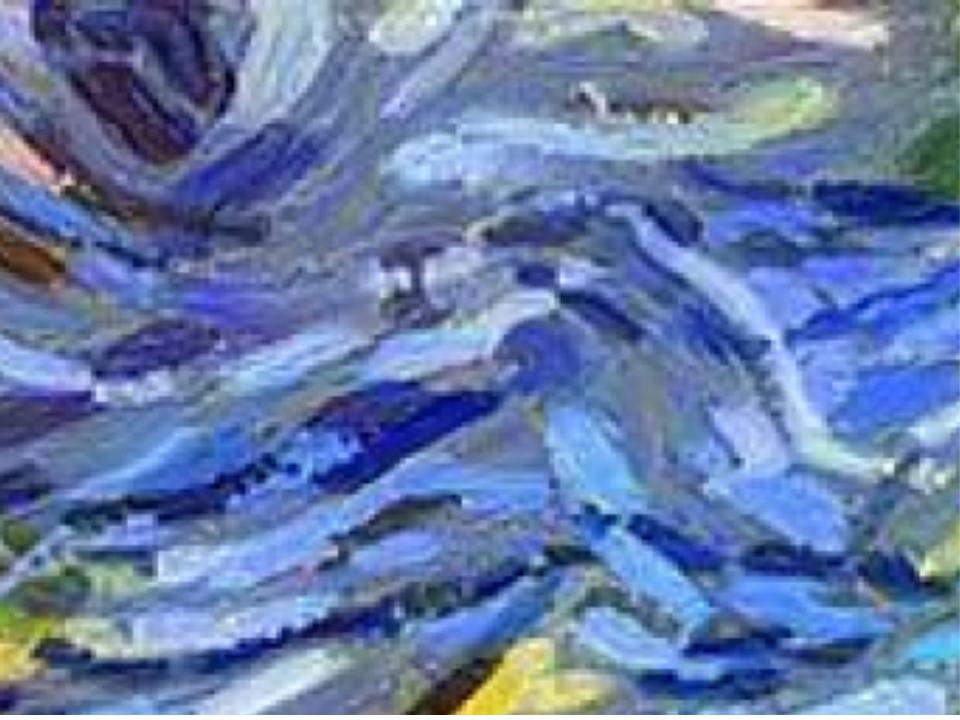} 
    
    \end{tabular} \\
    
    \vspace{1.0pt}
    
    \caption{For Gatys et al. \cite{Gatys2016}, the pixels are adjusted to match the brushstroke pattern. In our approach, the brushstroke pattern is occurring by design. Style image: ``Starry Night'' by Vincent van Gogh. Content image: original image of Tuebingen from the paper \cite{Gatys2016}. Same region of the sky is cropped.}
    \label{fig:patch_compare}
\end{figure}

\begin{figure*}
\begin{center}
   \includegraphics[width=0.99\linewidth]{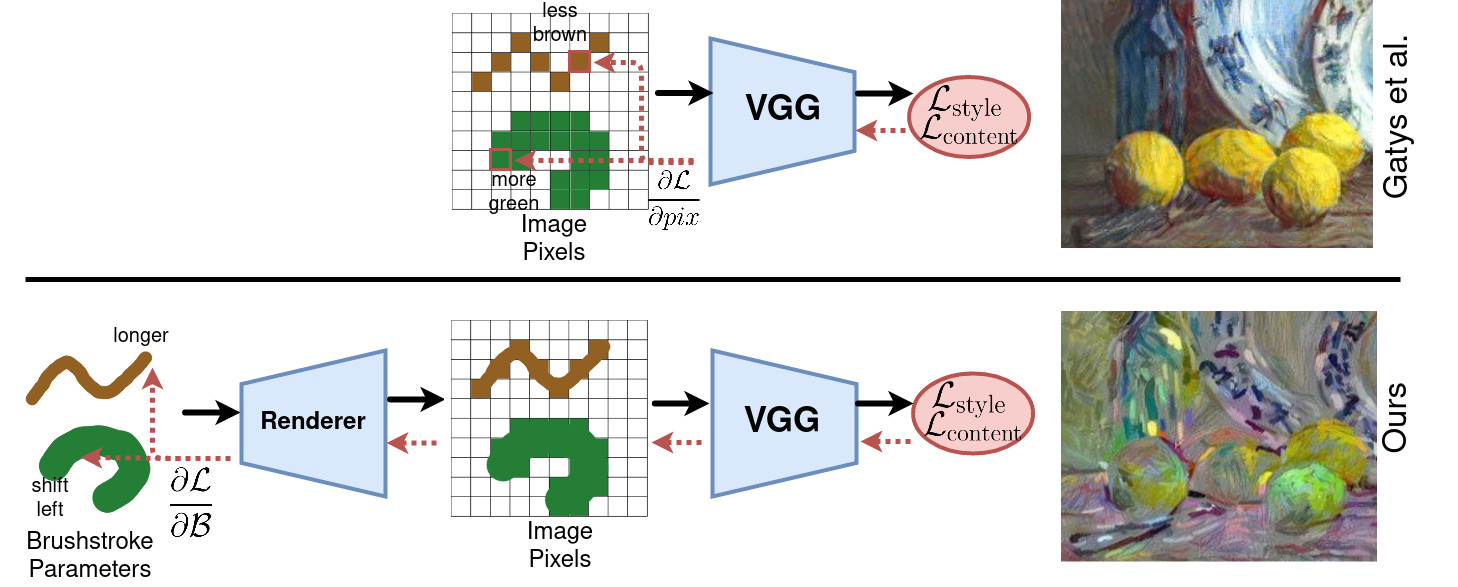}
\end{center}
   \caption{Comparison of our method (bottom row) with Gatys et al. \cite{Gatys2016} (top row). Gatys et al. \cite{Gatys2016} optimize pixels to minimize style and content loss. We directly optimize parameters of the brushstrokes. To do that we have designed a differentiable rendering mechanism that maps brushstrokes onto the canvas. Each brushstroke is parameterized by color, location, width and shape. Brushstroke parameters are updated by gradient backpropagation (red, dashed arrows).}
\label{fig:approach_diagram}
\end{figure*}

\begin{figure}
    \centering
    \def\arraystretch{0.0}
    \setlength{\tabcolsep}{1.0pt}
    \begin{tabular}{cc}
    \includegraphics[width=0.2\linewidth]{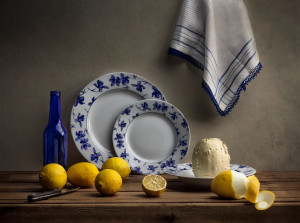}\includegraphics[width=0.1\textwidth]{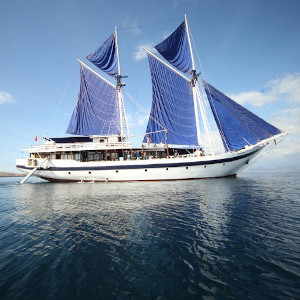} &
    \includegraphics[width=0.22\linewidth]{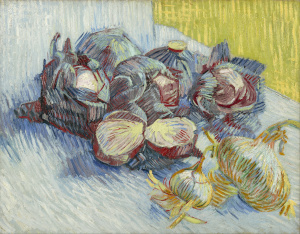}\includegraphics[width=0.12\textwidth]{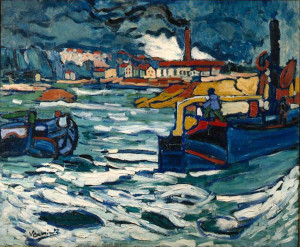} \\
    \vspace{2pt} \\
    (a) Content & (b) Style \\
    \vspace{2pt} \\
    \includegraphics[width=0.25\textwidth]{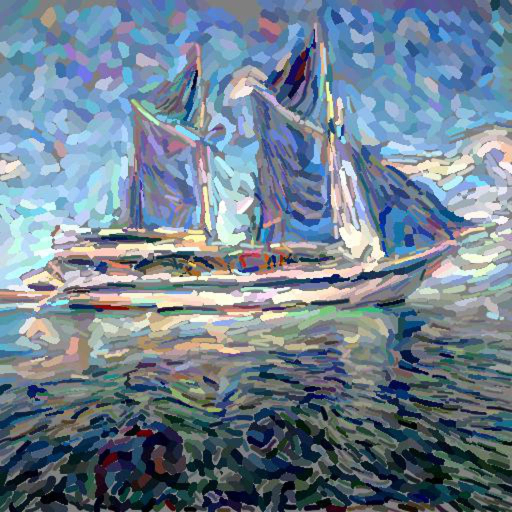} &
    \includegraphics[width=0.25\textwidth]{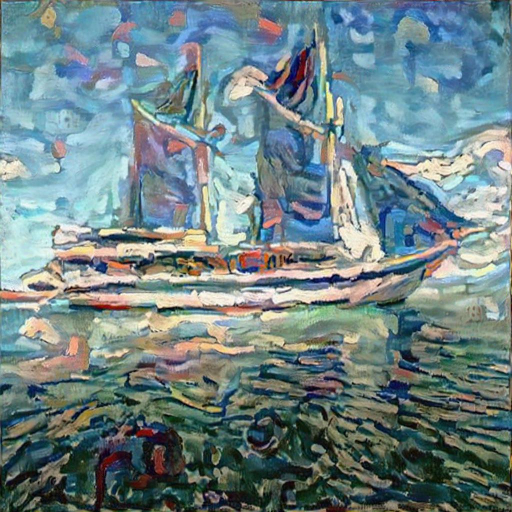} \\
    \vspace{2pt} \\
    \includegraphics[width=0.25\textwidth]{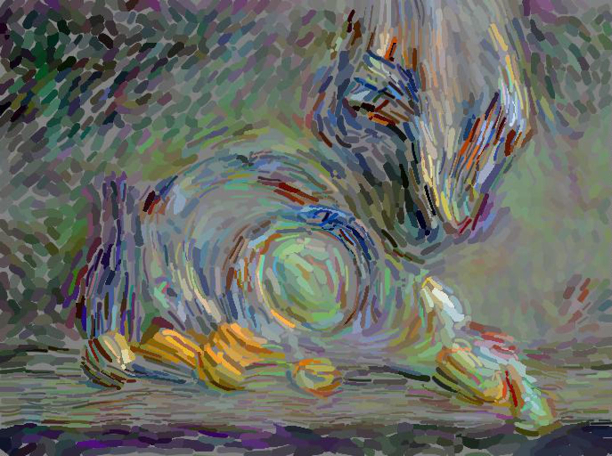} &
    \includegraphics[width=0.25\textwidth]{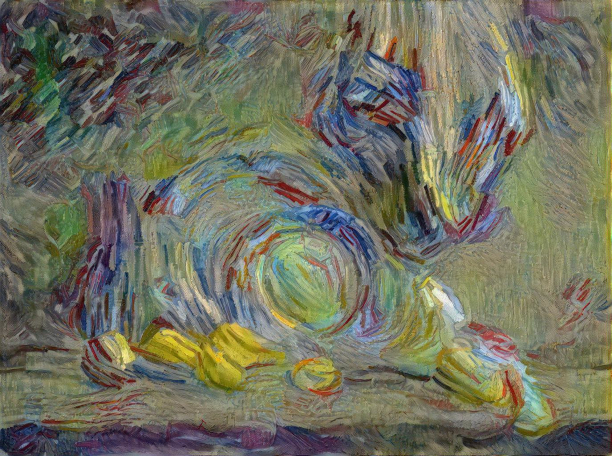} \\
    \vspace{2pt} \\
    (c) Before Pixel Optimization & (d) After Pixel Optimization \\

    \end{tabular} \\
    
    \caption{The effect of the pixel optimization. Brushstrokes are blended together and texture is added. Zoom in for details.}
    \label{fig:before_after_pix_opt}
\end{figure}

The method by Gatys et al. \cite{Gatys2016} adjusts each pixel individually to minimize the content and style losses. However, artworks generally consist of brushstrokes, not pixels. Instead of optimizing on pixels, we therefore optimize directly on parameterized brushstrokes, using the same content and style losses defined in Eq.~\ref{eq:content_loss} and Eq.~\ref{eq:style_loss}, respectively. See Fig.~\ref{fig:approach_diagram} for an overview of our method and Fig.~\ref{fig:patch_compare} for a comparison of the synthesized brushstroke patterns. \\
Our brushstrokes are parameterized by location, color, width, and shape. The shape of a brushstroke is modelled as a quadratic B\'{e}zier curve \cite{Nakano2019, Ganin2018, Huang2019}, which can be parameterized by:
\begin{equation}\label{eq:bezier_curve}
\mathbf{B}(t) = (1 - t)^{2}\mathbf{P}_0 + 2(1 - t)t\mathbf{P}_1 + t^{2}\mathbf{P}_2 \mbox{ , } 0 \le t \le 1.
\end{equation}
A key difficulty here is to find an efficient and differentiable mapping from the brushstroke parameter space into the pixel domain. To this end, we propose a mechanism to construct this mapping explicitly. See Sec.~\ref{sec:neural_renderer} for details. \\
Using our rendering mechanism we can backpropagate gradients from the style and content losses through the rendered pixels directly to the brushstroke parameters. \\
After the optimization is finished, we render the optimized brushstroke parameters to obtain an image $I$ and then apply the standard Gatys et al. \cite{Gatys2016} approach on the pixel level using $I_s$ as style image and $I$ as content image. This final step blends the brushstrokes together and adds some texture. Fig.~\ref{fig:before_after_pix_opt} shows the effect of this pixel optimization.

\subsection{Implementation Details}\label{sec:implement_details}
Similar to Gatys et al. \cite{Gatys2016}, we use layers ``conv4\textunderscore 2'' and ``conv5\textunderscore 2'' for the content loss and layers ``conv1\textunderscore 1'', ``conv2\textunderscore 1'', ``conv3\textunderscore 1'', ``conv4\textunderscore 1'', and ``conv5\textunderscore 1'' for the style loss. \\
We use Adam \cite{adam2014} with learning rate 0.1 for optimization. \\
Similar to Johnson et al. \cite{Johnson2016}, we employ a total variation regularization. 


\subsection{Differentiable Renderer}\label{sec:neural_renderer}

Nowadays, generative models have reached unmatched image quality on a variety of datasets \cite{karras2019style, brock2018large}. Thus, our first attempt to generate brushstrokes followed this line of work. We generated a dataset of brushstrokes simulated in the FluidPaint environment \footnote{\url{https://david.li/paint/}} and trained a network inspired by StyleGAN \cite{karras2019style} to generate images conditioned on brushstroke parameters. Despite achieving satisfactory visual quality, the main limitation of this approach is that it is memory-intensive and can not be efficiently scaled to process a large number of brushstrokes in parallel. This is critical for us since our method relies on an iterative optimization procedure.

Therefore, instead of training a neural network to generate brushstrokes, we explicitly construct a differentiable function which transforms a collection of brushstrokes parameterized by location, shape, width and color into pixel values on a canvas. Formally, the renderer is a function:
\begin{equation}\label{eq:renderer_as_function}
    \mathcal{R}: \mathbb{R}^{N \times F} \rightarrow \mathbb{R}^{H \times W \times 3},
\end{equation}
where $N$ denotes the number of brushstrokes, $F$ the number of brushstroke parameters (12 in our case), and $H$ and $W$ are the height and width of the image to render.
This renderer requires less memory and is also not constrained by the limitations of a brushstroke dataset.

\subsubsection{Motivation and Idea}
Before explaining how our render works, let us start with a simple example. Assume we have a flat disk parameterized with color, radius, and location (1, 1, and 2 scalars respectively) and we want to draw it on a canvas. For the sake of brevity, we assume our images are grayscale but the algorithm trivially generalizes to the RGB space. A grayscale image is a 2D matrix of pixel values. First, we need to decide for every pixel whether or not it belongs to the disk. For this, we simply subtract the disk location from each pixel coordinate and compute the $L_2$ norm to obtain distances $D$ from each pixel to the disk center. Now we have to check if the distance $D$ is smaller then the radius to get a binary mask $M$. To incorporate color, it suffices to multiply the mask by a color value.

If we have two disks, we simply repeat the procedure above for each disk separately and obtain two separate images with disks, namely $I_{1}, I_{2} \in \mathbb{R}^{H \times W \times 3}$. Now, how do we blend $I_{1}, I_{2}$ together? If they do not overlap we can sum the pixel values across disks $I_{1} + I_{2}$. However, if the disks overlap, adding them together will produce artifacts. Therefore, in the overlapping regions, we will assign each pixel to the nearest disk. This can be done by computing the distances $D_1,\ D_2 \in \mathbb{R}^{H \times W}$ from each pixel to each disk center and determine for every pixel the closer disk. We call this object an assignment matrix $A:= \{1\ \text{if}\ D_1 \leq D_2,\ 0\ \text{otherwise}\} \in \mathbb{R}^{H \times W}$. Now the final image $I$ can be computed using the matrices $I_1$, $I_2$ and $A$: $I:=I_1 * A + I_2 * (\mathbf{1} - A) $. The assignment matrix $A$ naturally generalizes to $N$ objects:
\begin{equation}\label{eq:assignment_matrix}
    A(i,j,n) := \begin{cases}
    1 &\text{if}\  D_n(i,j) < D_k(i,j)\ \forall k\neq n,\\
    0 &\text{otherwise}.
    \end{cases}
\end{equation}
It indicates which object is the nearest to the coordinate $(i,j)$. The final image computation for $N$ images of disks $I_1, .. , I_N$ then corresponds to:
\begin{equation}\label{eq:assignment_renderings_blending}
    I(i,j):=\sum_{n=1}^{N} I_n(i,j) * A(i,j,n)
\end{equation}

Hence, the final image is computed by the weighted sum of renderings weighted according to the  assignment matrix $A$. Both the assignment matrix and the individual renderings $I_1,...,I_N$ originate from the distance matrices $D_1,.., D_N$ from each pixel location to the object. Indeed, to render a single object we take its distance matrix, threshold with radius/width and multiply by a color value. The assignment matrix is an indicator function of the smallest distance across distances $D_1,.., D_N$. Thus, the matrix of distances is a cornerstone of our approach. We can effectively render any object for which we can compute the distances from each pixel to the object.

Our initial goal was to render brushstrokes. To render a disk we take a distance matrix $D$, get a mask of points that are closer than the radius and multiply this mask by a color value. The same holds for a B\'{e}zier curve. \\
First, we compute a matrix of distances to the curve $D_\mathbf{B}$ (matrix of distances from every point in a 2D image to the nearest point on the B\'{e}zier curve). \\
Then, we mask points that are closer than the brushstroke width and multiply them by a color value. We approximate the distance from a point $p$ to a B\'{e}zier curve by sampling $S$ equidistant points $p'_1,..,p'_S$ along the curve and computing the minimum pairwise distance between $p$ and $p'_1,..,p'_S$. Note that there exists an analytical solution of this distance for a quadratic B\'{e}zier curve, however, the approximated distance allows the use of arbitrary parametric curves. \\
In the final step, we can compute the individual renderings of brushstrokes and the assignment matrix as in Eq. \ref{eq:assignment_matrix} and blend them together into the final rendering with Eq. \ref{eq:assignment_renderings_blending}.

For the sake of clarity, we have left out two important details in the above explanation. \\
First, the renderer should be differentiable, yet, the computation of the assignment matrix and the masking operation are both discontinuous. To alleviate this problem, we implement a masking operation with a sigmoid function. To make the assignment matrix computation differentiable we replace it with a softmax operation with high temperature. \\
Second, the computation of distances between every brushstroke and every pixel on the canvas is computationally expensive, memory-intensive and also redundant because a brushstroke only affects the nearby area of the canvas. Therefore, we limit the computation of distances from a pixel to all the brushstrokes to only the K nearest brushstrokes, see Sec. \ref{sec:distances} for details.

\begin{algorithm}

\SetAlgoLined
\KwInput{Brushtroke parameters $\mathcal{B} = \{B_1, B_2,...,B_N\}$, temperature parameter $t$, number samples per curve $S$}
\KwOutput{Image $I \in \mathbb{R}^{H \times W \times 3}$}
 init $\mathcal{C} \in \mathbb{R}^{H \times W \times 2}$\ \tcp*{coordinates tensor, $\mathcal{C}(x,y)=(x,y)$ }
 init tensor of brushtrokes colors $c_{strokes}$ from $\mathcal{B}$ parameters \tcp*{shape=[N, 3]}
 init tensor of brushtrokes widths $w_{strokes}$ from $\mathcal{B}$ parameters \tcp*{shape=[N]}
 sample $S$ points $t \in [0; 1]$
 sample points $t$ at each brushtroke $\mathcal{B}_{sampled}:=\{\text{compute}~B_i(t_j)\ \text{with}\ 
 Eq.\ref{eq:bezier_curve}\ | \forall i,j \} $ \tcp*{shape=[N,S,2] }
 
 $D_(x,y,n,s) := ||\mathcal{C}(x,y) - \mathcal{B}_{sampled}(n,s)||_2$  \tcp*{Distances from each sampled point on a stroke to each coordinate, shape=[H,W,N,S]}
 
 $D_{strokes} := min(D, axis=4)$ \tcp*{distance from a coordinate $x,y$ to the nearest point on a curve. shape=[H, W, N]}
 
 $M_{strokes} := \textbf{sigm}(\textbf{max}(t \cdot ||w_{strokes} - D_{strokes}||_2 , axis=4) $ \tcp*{mask of each stroke, shape=[H, W, N]}

 $I_{strokes} := M_{strokes} \cdot c_{strokes}$  \tcp*{rendering of each stroke, shape=[H, W, N, 3]}

 $A:=\textbf{softmax}(t \cdot D_{strokes}, axis=3)$ \tcp*{assignment, shape=[H, W, N]}
 
 $I:=\textbf{einsum}('xyn,xync->xyc', A, I) $ \tcp*{final rendering, see Eq.\ref{eq:assignment_renderings_blending}}

 \caption{Renderer}\label{alg:render_algo}
\end{algorithm}
See Alg.\ref{alg:render_algo}. See Sec. \ref{sec:renderer} for additional technical details of the implementation.

\section{Experiments}\label{sec:experiments}

\subsection{Deception Rate}
To evaluate the quality of the stylization we use a deception rate proposed by Sanakoyeu et al. \cite{Sanakoyeu2018}. The method is based on a network trained to classify paintings into artists. The deception rate is the fraction of stylized images that the network has assigned to the artist, whose artwork has been used for stylization. However, a high deception score indicates high similarity to the target image. But this metric does not indicate how plausible a stylized image is. To measure this quality we conduct the following experiment: we show to a human subject four crop outs. Each one can be either taken from a real artwork or from a generated image. The task is to detect all real crop outs. The experiment is conducted with 10 human subjects, each participant evaluates 200+ tuples. Fake images are randomly sampled from one of three methods: ours, Gatys et al. \cite{Gatys2016}, and AST \cite{Sanakoyeu2018}. For each method we report the proportion of ranking this image as real, see Tab. \ref{tab:deception}.


\begin{table}

    \scriptsize
    \caption{(Left) Deception score. Wikiart test gives the accuracy on real artworks from the test set. 
    Photos correspond to the content images used by each of the methods for style transfer.
    (Right) Human deception rate. The probability of labeling randomly sampled crop out of a specified class as real. 
    Both scores are averaged over $8$ styles.}
    
    \centering
    \begin{tabular}{l||cc}
    \hline\noalign{\smallskip}
    Method & Mean deception score $\uparrow$ & Mean human deception rate $\uparrow$ \\
    \noalign{\smallskip}
    \hline
    \noalign{\smallskip}
    
    AdaIN \cite{Huang2017}  & 0.08 & 0.035 \\
    WCT \cite{Li2017_a}  & 0.11 & 0.091 \\
    Gatys et al. \cite{Gatys2016} & 0.389 & 0.139 \\
    AST \cite{Sanakoyeu2018} & 0.451 & 0.146 \\
    \textbf{Ours} & \textbf{0.588} & \textbf{0.268} \\
    \hline\hline\noalign{\smallskip}
    Wikiart test & 0.687 & - \\
    Photos & 0.002 & -\\
    \hline
      
    \end{tabular}

    \label{tab:deception}

\end{table}

\subsection{Differentiable Renderer}
\begin{figure}
    \centering
    \includegraphics[width=.45\textwidth]{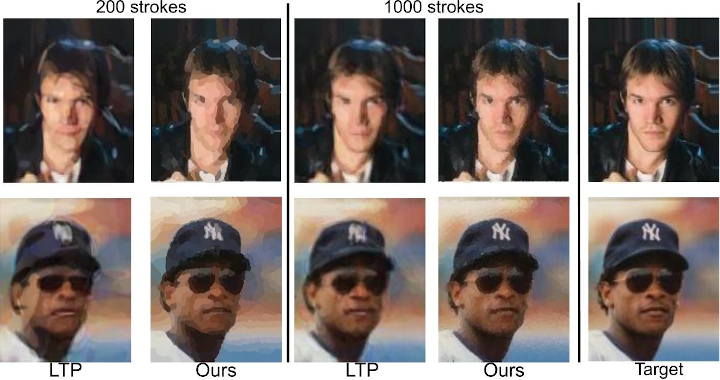}

    \caption{Comparison to the Learning to Paint (LTP) by Huang et al. \cite{Huang2019} on the image reconstruction task. Our method directly minimizes $l_2$ distance between the input target image and image rendered as a collection of brushstrokes. Using our renderer we achieve 20\% lower Mean Squared Error (MSE) for 200 strokes and 49\% lower MSE for 1000 strokes. Please zoom in for details.}
    \label{fig:comarions_to_ltp}
\end{figure}

We compare our simple explicitly constructed renderer to the rendering mechanism proposed by Huang et al. \cite{Huang2019}. Our approach is slower, but it requires no pretraining on specific datasets as opposed to Huang et al. \cite{Huang2019}. We achieve 20\% lower mean squared error (MSE) using 200 strokes, and 49\% lower MSE on 1000 strokes. The comparison has been conducted on the CelebA dataset. See Fig. \ref{fig:comarions_to_ltp} for a visual comparison. 

\subsection{Fitting Brushstrokes to Artwork}\label{sec:fit_to_artwork}
We can fit brushstrokes not only to a photograph but also to an artwork.
This procedure is useful if we want to study the distribution of brushstrokes in an artwork. It has been shown by Li et al. \cite{Li2012} that this information may be helpful to detect forgeries and analyze the style of an artist. In Fig. \ref{fig:self_portrait_reconstruct} we show reconstructions of ``Self-Portrait'' by Vincent van Gogh obtained using our renderer.

We additionally trained a neural network that receives brushstroke parameters as input and generates the corresponding brushstrokes. The network employs an architecture inspired by StyleGAN \cite{karras2019style} and was trained on a dataset obtained using the FluidPaint environment. The brushstroke parameterization is as described in this paper. The trained renderer yields results comparable to our simple renderer but requires more precise hyperparameter tuning an takes more time to optimize on. Since the trained renderer is based on the StyleGAN \cite{karras2019style} architecture, it consumes much more memory and thus fitting hundreds or thousands of brushstrokes cannot be run in parallel. In Fig. \ref{fig:self_portrait_reconstruct} we present results of our renderer and the trained renderer. See Sec. \ref{sec:trained_neural_renderer} for more details.

\begin{figure}
    \centering
    \def\arraystretch{0.0}
    \setlength{\tabcolsep}{1.0pt}
    \begin{tabular}{ccc}
    Trained Renderer & Original & Our Renderer\\
    \vspace{1pt} \\
    \includegraphics[width=0.16\textwidth]{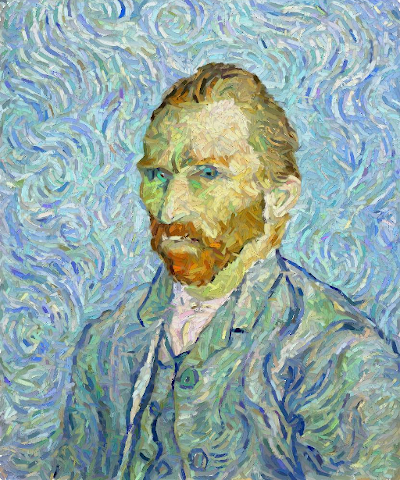} &
    \includegraphics[width=0.16\textwidth]{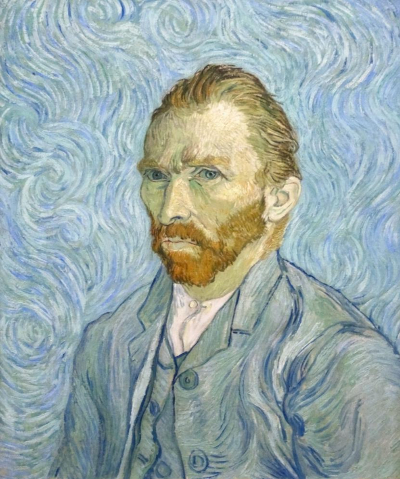} &
    \includegraphics[width=0.16\textwidth]{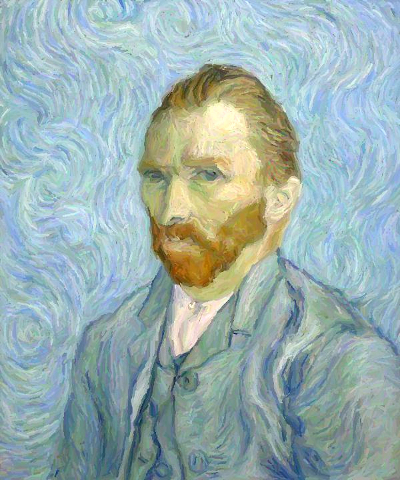}
    \end{tabular} \\
    \caption{Reconstructions of ``Self-Portrait'' by Vincent van Gogh using our brushstroke renderer and a trained renderer. In either case we use 10.000 brushstrokes. }
    \label{fig:self_portrait_reconstruct}
\end{figure}

\subsection{Controlling Brushstrokes}\label{sec:res_control}
To highlight the additional control our brushstroke representation enables over the stylization process, we show how users can control the flow of brushstrokes in the stylized image, see Fig.~\ref{fig:user_control}. A user can draw arbitrary curves on the content image and the brushstrokes in the stylized image will follow these curves. This can be achieved by adding a simple projection loss that enforces brushstrokes along the drawn paths to align with the tangent vectors of the paths. See Sec. \ref{sec:drawing} for details. Fig.~\ref{fig:user_control} further shows the effect the number of brushstrokes has on the stylization.

\section{Conclusion and Future Work}\label{sec:experiments}

In this paper, we have proposed to switch the representation for style transfer from pixels to parameterized brushstrokes. We argue that the latter representation is more natural for artistic style transfer and show how it benefits the visual quality of the stylizations and enables additional control.

We have further introduced an explicit rendering mechanism and show that it can be applied even beyond the field of style transfer.

A limitation of our approach is that it performs best for artistic styles where brushstrokes are clearly visible. This can potentially be alleviated with more sophisticated brushstroke blending procedures and should be investigated in future endeavors.

\section{Acknowledgments}
This work has been supported in part by the German Research Foundation (DFG) within project 421703927.

\begin{figure*}
    \centering
    \def\arraystretch{0.0}
    \setlength{\tabcolsep}{1.0pt}
    \begin{tabular}{cccc}
    \rotatebox{90}{\hspace{1.3cm} Style}
    \includegraphics[width=\imgsize\textwidth]{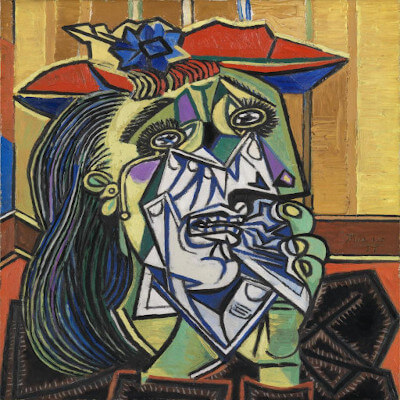} &
    \includegraphics[width=\imgsize\textwidth]{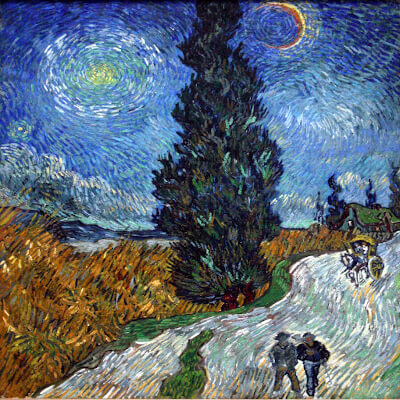} &
    \includegraphics[width=\imgsize\textwidth]{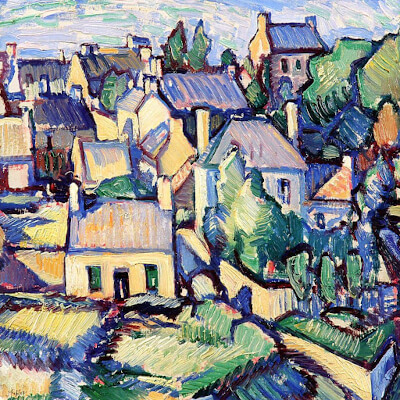} &
    \includegraphics[width=\imgsize\textwidth]{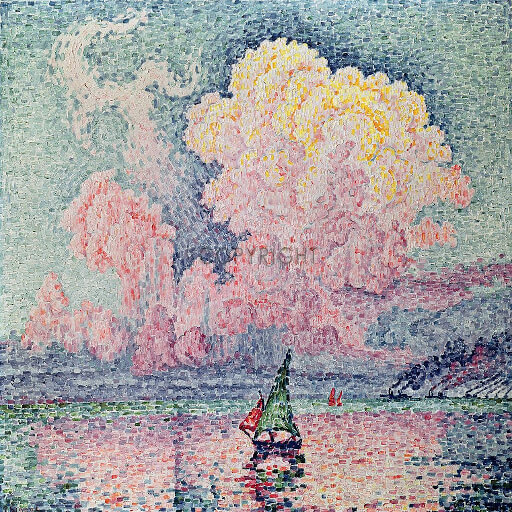} \\
    \vspace{1pt} & \\
    \rotatebox{90}{\hspace{1.1cm} Content}
    \includegraphics[width=\imgsize\textwidth]{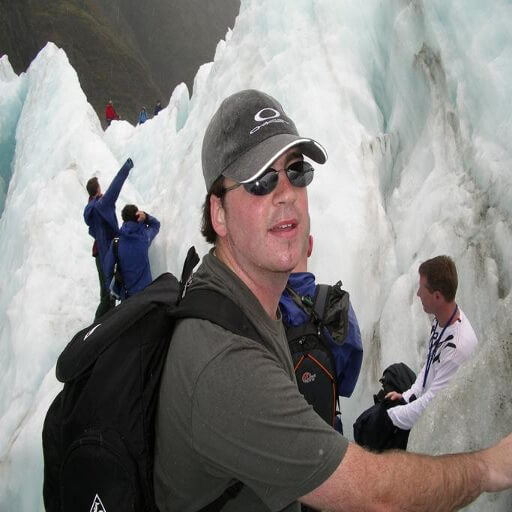} &
    \includegraphics[width=\imgsize\textwidth]{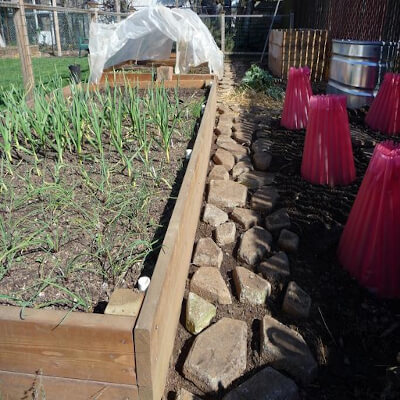} &
    \includegraphics[width=\imgsize\textwidth]{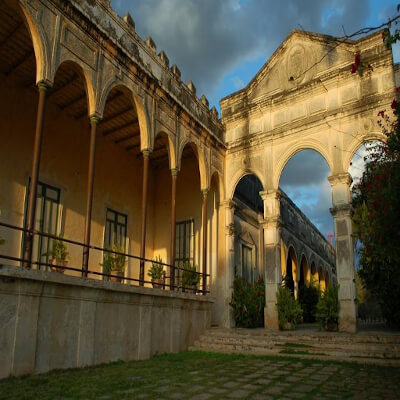} &
    \includegraphics[width=\imgsize\textwidth]{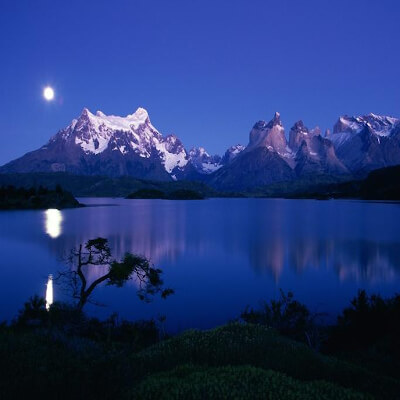} \\
    \vspace{1pt} & \\
    \rotatebox{90}{\hspace{1.3cm} Ours}
    \includegraphics[width=\imgsize\textwidth]{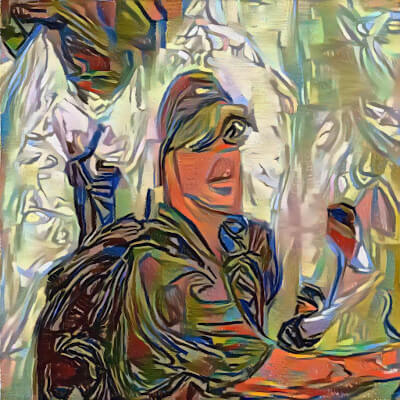} &
    \includegraphics[width=\imgsize\textwidth]{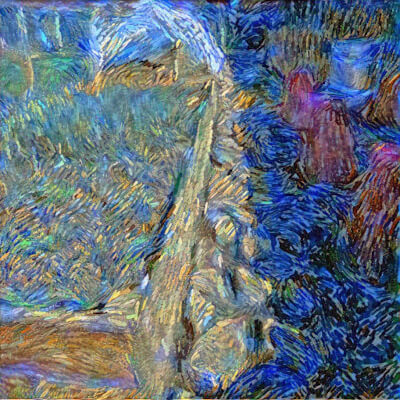} &
    \includegraphics[width=\imgsize\textwidth]{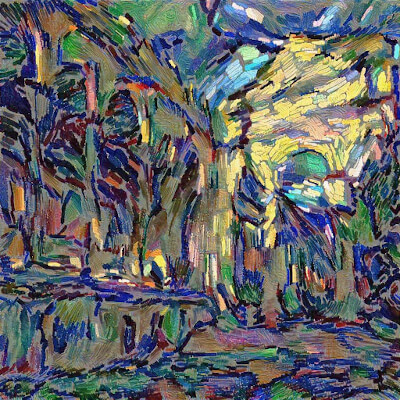} &
    \includegraphics[width=\imgsize\textwidth]{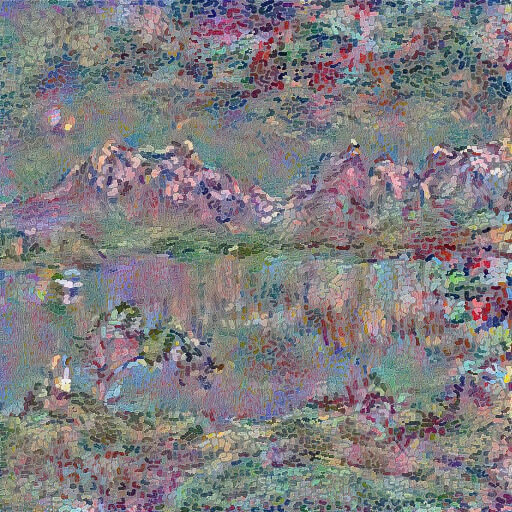} \\
    \vspace{1pt} & \\
    \rotatebox{90}{\hspace{0.32cm} Svoboda et al. \cite{Svoboda2020CVPR}}
    \includegraphics[width=\imgsize\textwidth]{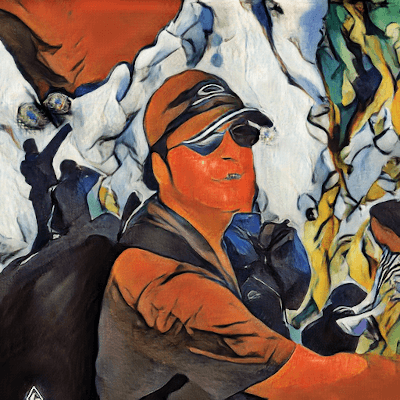} &
    \includegraphics[width=\imgsize\textwidth]{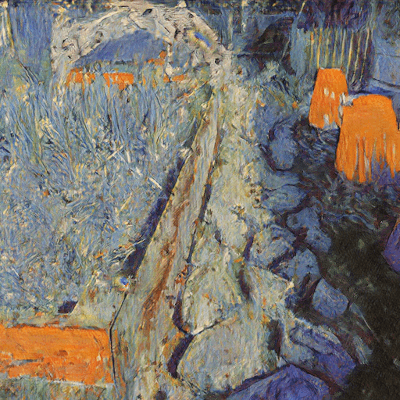} &
    \includegraphics[width=\imgsize\textwidth]{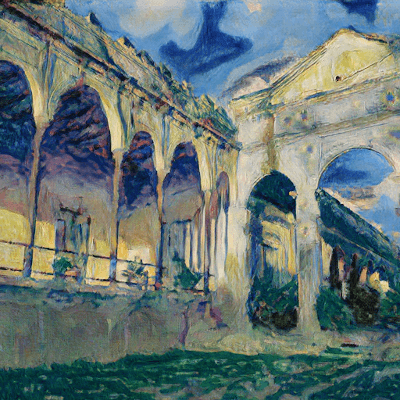} &
    \includegraphics[width=\imgsize\textwidth]{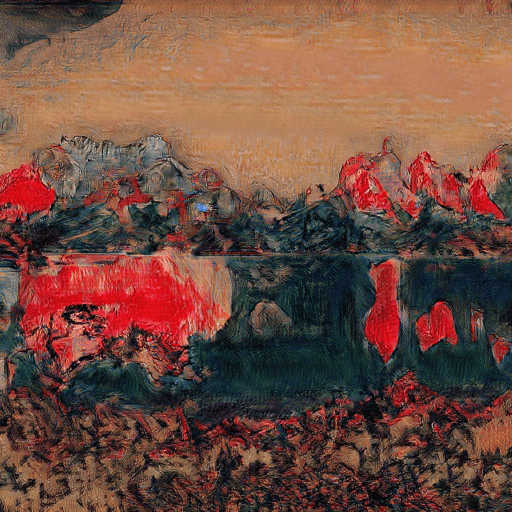} \\
    \vspace{1pt} & \\
    \rotatebox{90}{\hspace{0.6cm} Gatys et al. \cite{Gatys2016}}
    \includegraphics[width=\imgsize\textwidth]{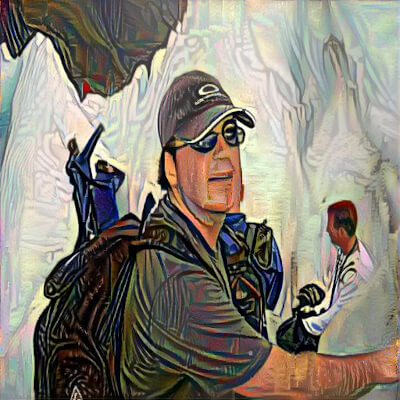} &
    \includegraphics[width=\imgsize\textwidth]{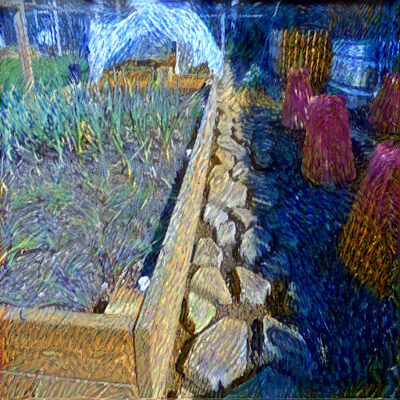} &
    \includegraphics[width=\imgsize\textwidth]{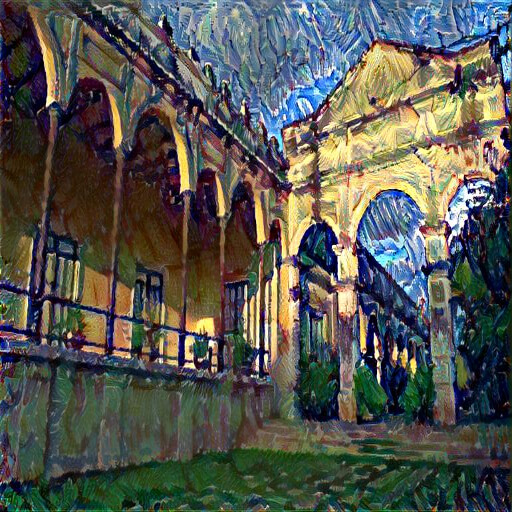} &
    \includegraphics[width=\imgsize\textwidth]{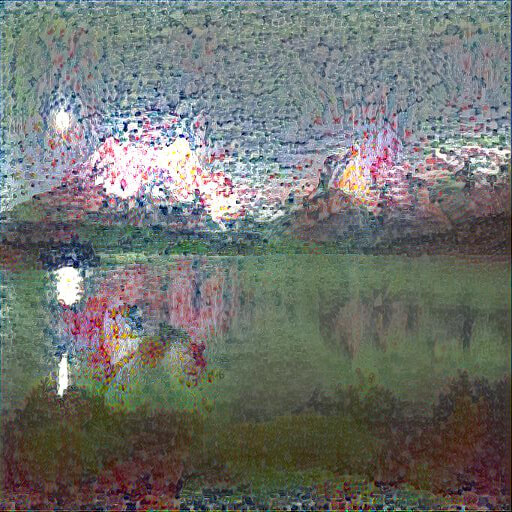} \\
    \vspace{1pt} & \\
    \rotatebox{90}{\hspace{1.2cm} AST \cite{Sanakoyeu2018}}
    \includegraphics[width=\imgsize\textwidth]{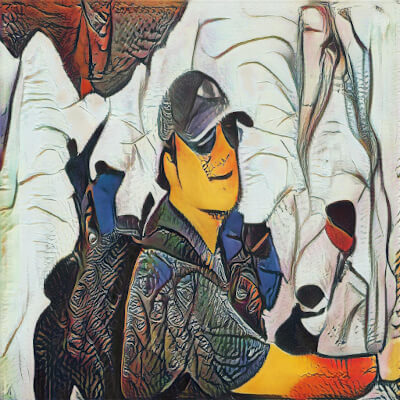} &
    \includegraphics[width=\imgsize\textwidth]{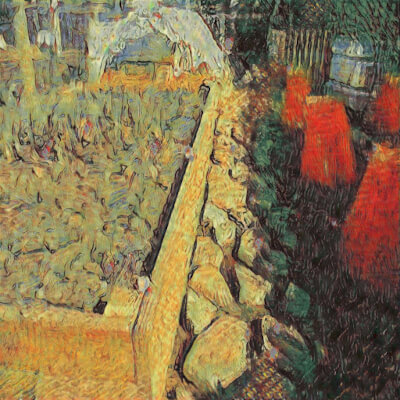} &
    \includegraphics[width=\imgsize\textwidth]{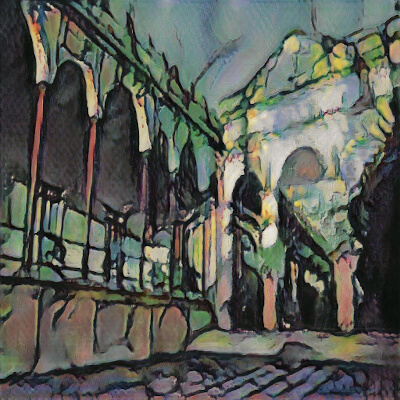} &
    \includegraphics[width=\imgsize\textwidth]{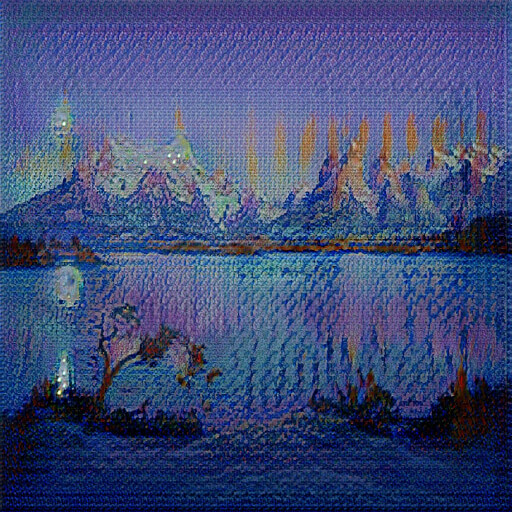} \\
    \end{tabular} \\
    \vspace{3.0pt}
    
    \caption{Comparison with other methods for images used by Svoboda et al. \cite{Svoboda2020CVPR} and AST \cite{Sanakoyeu2018}, please zoom in for details. See Sec. \ref{sec:add_results} for full size images.}
    \label{fig:comparison_w_others}
\end{figure*}

\FloatBarrier

{\small
\bibliographystyle{ieee_fullname}
\bibliography{egbib}

\begin{thebibliography}{10}\itemsep=-1pt

\bibitem{slic_pami}
Radhakrishna Achanta, Appu Shaji, Kevin Smith, Aurelien Lucchi, Pascal Fua, and
  Sabine S\"{u}sstrunk.
\newblock Slic superpixels compared to state-of-the-art superpixel methods.
\newblock {\em IEEE Trans. Pattern Anal. Mach. Intell.}, 34(11):2274--2282,
  2012.

\bibitem{Arjovsky2017}
Martin Arjovsky, Soumith Chintala, and L{\'e}on Bottou.
\newblock Wasserstein generative adversarial networks.
\newblock 2017.

\bibitem{Berezhnoy2009}
Igor~E. Berezhnoy, Eric~O. Postma, and H.~Jaap Herik.
\newblock Automatic extraction of brushstroke orientation from paintings.
\newblock {\em Mach. Vision Appl.}, 20(1):1–9, Jan. 2009.

\bibitem{brock2018large}
Andrew Brock, Jeff Donahue, and Karen Simonyan.
\newblock Large scale gan training for high fidelity natural image synthesis.
\newblock {\em arXiv preprint arXiv:1809.11096}, 2018.

\bibitem{Chang2020ECCV}
Hsin-Yu Chang, Zhixiang Wang, and Yung-Yu Chuang.
\newblock Domain-specific mappings for generative adversarial style transfer.
\newblock In {\em Eur. Conf. Comput. Vis.}, 2020.

\bibitem{Chen2020ECCV}
Xinghao Chen, Yiman Zhang, Yunhe Wang, Han Shu, Chunjing Xu, and Chang Xu.
\newblock Optical flow distillation: Towards efficient and stable video style
  transfer.
\newblock In {\em Eur. Conf. Comput. Vis.}, 2020.

\bibitem{Chiu2020ECCV}
Tai-Yin Chiu and Danna Gurari.
\newblock Iterative feature transformation for fast and versatile universal
  style transfer.
\newblock In {\em Eur. Conf. Comput. Vis.}, 2020.

\bibitem{Choi2018CVPR}
Yunjey Choi, Minje Choi, Munyoung Kim, Jung-Woo Ha, Sunghun Kim, and Jaegul
  Choo.
\newblock Stargan: Unified generative adversarial networks for multi-domain
  image-to-image translation.
\newblock In {\em IEEE Conf. Comput. Vis. Pattern Recog.}, 2018.

\bibitem{Choi2020CVPR}
Yunjey Choi, Youngjung Uh, Jaejun Yoo, and Jung-Woo Ha.
\newblock Stargan v2: Diverse image synthesis for multiple domains.
\newblock In {\em IEEE Conf. Comput. Vis. Pattern Recog.}, 2020.

\bibitem{Dumoulin2017}
Vincent Dumoulin, Jonathon Shlens, and Manjunath Kudlur.
\newblock A learned representation for artistic style.
\newblock In {\em Int. Conf. Learn. Represent.}, 2017.

\bibitem{Efros2001}
Alexei~A. Efros and William~T. Freeman.
\newblock Image quilting for texture synthesis and transfer.
\newblock In {\em Proceedings of the 28th Annual Conference on Computer
  Graphics and Interactive Techniques}, SIGGRAPH '01, page 341–346, New York,
  NY, USA, 2001. Association for Computing Machinery.

\bibitem{Ganin2018}
Yaroslav Ganin, Tejas Kulkarni, Igor Babuschkin, S.M.~Ali Eslami, and Oriol
  Vinyals.
\newblock Synthesizing programs for images using reinforced adversarial
  learning.
\newblock 2018.

\bibitem{Gatys2016}
Leon~A. Gatys, Alexander~S. Ecker, and Matthias Bethge.
\newblock Image style transfer using convolutional neural networks.
\newblock In {\em IEEE Conf. Comput. Vis. Pattern Recog.}, June 2016.

\bibitem{Gooch2002}
Bruce Gooch, Greg Coombe, and Peter Shirley.
\newblock Artistic vision: Painterly rendering using computer vision
  techniques.
\newblock In {\em Proceedings of the 2nd International Symposium on
  Non-Photorealistic Animation and Rendering}, NPAR '02, page 83–ff, New
  York, NY, USA, 2002. Association for Computing Machinery.

\bibitem{Gulrajani2017}
Ishaan Gulrajani, Faruk Ahmed, Martin Arjovsky, Vincent Dumoulin, and Aaron
  Courville.
\newblock Improved training of wasserstein gans.
\newblock In {\em Adv. Neural Inform. Process. Syst.}, 2017.

\bibitem{Ha2018}
David Ha and Douglas Eck.
\newblock A neural representation of sketch drawings.
\newblock In {\em Int. Conf. Learn. Represent.}, 2018.

\bibitem{Haeberli1990}
Paul Haeberli.
\newblock Paint by numbers: Abstract image representations.
\newblock SIGGRAPH '90, page 207–214, New York, NY, USA, 1990. Association
  for Computing Machinery.

\bibitem{Hertzmann1998}
Aaron Hertzmann.
\newblock Painterly rendering with curved brush strokes of multiple sizes.
\newblock In {\em SIGGRAPH '98}, 1998.

\bibitem{Hertzmann2001}
Aaron Hertzmann.
\newblock Paint by relaxation.
\newblock In {\em Computer Graphics International 2001}, CGI '01, page 47–54,
  USA, 2001. IEEE Computer Society.

\bibitem{Hertzmann2001B}
Aaron Hertzmann, Charles~E. Jacobs, Nuria Oliver, Brian Curless, and David~H.
  Salesin.
\newblock Image analogies.
\newblock SIGGRAPH '01, page 327–340, New York, NY, USA, 2001. Association
  for Computing Machinery.

\bibitem{Huang2017}
Xun Huang and Serge Belongie.
\newblock Arbitrary style transfer in real-time with adaptive instance
  normalization.
\newblock In {\em Int. Conf. Comput. Vis.}, 2017.

\bibitem{Huang2018ECCV}
Xun Huang, Ming-Yu Liu, Serge Belongie, and Jan Kautz.
\newblock Multimodal unsupervised image-to-image translation.
\newblock In {\em Eur. Conf. Comput. Vis.}, 2018.

\bibitem{Huang2019}
Zhewei Huang, Wen Heng, and Shuchang Zhou.
\newblock Learning to paint with model-based deep reinforcement learning.
\newblock 2019.

\bibitem{Isola2017CVPR}
Phillip Isola, Jun-Yan Zhu, Tinghui Zhou, and Alexei~A. Efros.
\newblock Image-to-image translation with conditional adversarial nets.
\newblock In {\em IEEE Conf. Comput. Vis. Pattern Recog.}, 2017.

\bibitem{Jia2019}
Biao Jia, Chen Fang, Jonathan Brandt, Byungmoon Kim, and D. Manocha.
\newblock Paintbot: A reinforcement learning approach for natural media
  painting.
\newblock {\em ArXiv}, abs/1904.02201, 2019.

\bibitem{Johnson2016}
Justin Johnson, Alexandre Alahi, and Li Fei-Fei.
\newblock Perceptual losses for real-time style transfer and super-resolution.
\newblock In {\em Eur. Conf. Comput. Vis.}, 2016.

\bibitem{karras2019style}
Tero Karras, Samuli Laine, and Timo Aila.
\newblock A style-based generator architecture for generative adversarial
  networks.
\newblock In {\em IEEE Conf. Comput. Vis. Pattern Recog.}, 2019.

\bibitem{Kim2020ICLR}
Junho Kim, Minjae Kim, Hyeonwoo Kang, and Kwang~Hee Lee.
\newblock U-gat-it: Unsupervised generative attentional networks with adaptive
  layer-instance normalization for image-to-image translation.
\newblock In {\em Int. Conf. Learn. Represent.}, 2020.

\bibitem{Kim2020ECCV}
Sunnie S.~Y. Kim, Nicholas Kolkin, Jason Salavon, and Gregory Shakhnarovich.
\newblock Deformable style transfer.
\newblock In {\em Eur. Conf. Comput. Vis.}, 2020.

\bibitem{adam2014}
Diederik~P. Kingma and Jimmy Ba.
\newblock Adam: A method for stochastic optimization.
\newblock In {\em Int. Conf. Learn. Represent.}, 2014.

\bibitem{Kotovenko2019}
Dmytro Kotovenko, Artsiom Sanakoyeu, Sabine Lang, Pingchuan Ma, and Bj{\"o}rn
  Ommer.
\newblock Using a transformation content block for image style transfer.
\newblock In {\em IEEE Conf. Comput. Vis. Pattern Recog.}, 2019.

\bibitem{Kotovenko2019_a}
Dmytro Kotovenko, Artsiom Sanakoyeu, Sabine Lang, and Bj{\"o}rn Ommer.
\newblock Content and style disentanglement for artistic style transfer.
\newblock In {\em Int. Conf. Comput. Vis.}, 2019.

\bibitem{Li2012}
Jia Li, Lei Yao, Ella Hendriks, and James~Z. Wang.
\newblock Rhythmic brushstrokes distinguish van gogh from his contemporaries:
  Findings via automated brushstroke extraction.
\newblock {\em IEEE Transactions on Pattern Analysis and Machine Intelligence},
  34(6):1159--1176, 2012.

\bibitem{Li2019CVPR}
Xueting Li, Sifei Liu, Jan Kautz, and Ming-Hsuan Yang.
\newblock Learning linear transformations for fast arbitrary style transfer.
\newblock In {\em IEEE Conf. Comput. Vis. Pattern Recog.}, 2019.

\bibitem{Li2017_a}
Yijun Li, Chen Fang, Jimei Yang, Zhaowen Wang, Xin Lu, and Ming-Hsuan Yang.
\newblock Universal style transfer via feature transforms.
\newblock In {\em Adv. Neural Inform. Process. Syst.}, 2017.

\bibitem{Li2018ECCV}
Yijun Li, Ming-Yu Liu, Xueting Li, Ming-Hsuan Yang, and Jan Kautz.
\newblock A closed-form solution to photorealistic image stylization.
\newblock In {\em Eur. Conf. Comput. Vis.}, 2018.

\bibitem{Li2017}
Yanghao Li, Naiyan Wang, Jiaying Liu, and Xiaodi Hou.
\newblock Demystifying neural style transfer.
\newblock In {\em IJCAI}, page 2230–2236. AAAI Press, 2017.

\bibitem{Litwinowicz1997}
Peter Litwinowicz.
\newblock Processing images and video for an impressionist effect.
\newblock In {\em Proceedings of the 24th Annual Conference on Computer
  Graphics and Interactive Techniques}, SIGGRAPH '97, page 407–414, USA,
  1997. ACM Press/Addison-Wesley Publishing Co.

\bibitem{Liu2018NIPS}
Ming-Yu Liu, Thomas Breuel, and Jan Kautz.
\newblock Unsupervised image-to-image translation networks.
\newblock In {\em Adv. Neural Inform. Process. Syst.}, 2017.

\bibitem{Liu2019ICCV}
Ming-Yu Liu, Xun Huang, Arun Mallya, Tero Karras, Timo Aila, Jaakko Lehtinen,
  and Jan Kautz.
\newblock Few-shot unsupervised image-to-image translation.
\newblock In {\em Int. Conf. Comput. Vis.}, 2019.

\bibitem{Mellor2019}
John F.~J. Mellor, Eunbyung Park, Yaroslav Ganin, I. Babuschkin, T. Kulkarni,
  Dan Rosenbaum, Andy Ballard, T. Weber, Oriol Vinyals, and S. Eslami.
\newblock Unsupervised doodling and painting with improved spiral.
\newblock {\em ArXiv}, abs/1910.01007, 2019.

\bibitem{Monroy2014}
Antonio Monroy, Peter Bell, and Björn Ommer.
\newblock Morphological analysis for investigating artistic images.
\newblock {\em Image and Vision Computing}, 32(6):414--423, 2014.

\bibitem{Monroy2011}
Antonio Monroy, Bernd Carqué, and Björn Ommer.
\newblock Reconstructing the drawing process of reproductions from medieval
  images.
\newblock In {\em 2011 18th IEEE International Conference on Image Processing},
  pages 2917--2920, 2011.

\bibitem{Nakano2019}
Reiichiro Nakano.
\newblock Neural painters: A learned differentiable constraint for generating
  brushstroke paintings.
\newblock {\em ArXiv}, abs/1904.08410, 2019.

\bibitem{Putri2017}
Tieta Putri, Ramakrishnan Mukundan, and Kourosh Neshatian.
\newblock Artistic style characterization of vincent van gogh s paintings using
  extracted features from visible brush strokes.
\newblock In {\em Conference on Pattern Recognition Applications and Methods
  (ICPRAM)}, 2017.

\bibitem{Risser2017Arxiv}
Eric Risser, Pierre Wilmot, and Connelly Barnes.
\newblock Stable and controllable neural texture synthesis and style transfer
  using histogram losses.
\newblock {\em ArXiv}, abs/1701.08893, 2017.

\bibitem{Sanakoyeu2018}
Artsiom Sanakoyeu, Dmytro Kotovenko, Sabine Lang, and Bj\"orn Ommer.
\newblock A style-aware content loss for real-time hd style transfer.
\newblock In {\em Eur. Conf. Comput. Vis.}, 2018.

\bibitem{Shen2018}
Falong Shen, Shuicheng Yan, and Gang. Zeng.
\newblock Neural style transfer via meta networks.
\newblock In {\em IEEE Conf. Comput. Vis. Pattern Recog.}, pages 8061--8069,
  2018.

\bibitem{Simonyan2015}
Karen Simonyan and Andrew Zisserman.
\newblock Very deep convolutional networks for large-scale image recognition.
\newblock In {\em Int. Conf. Learn. Represent.}, 2015.

\bibitem{Svoboda2020CVPR}
Jan Svoboda, Asha Anoosheh, Christian Osendorfer, and Jonathan Masci.
\newblock Two-stage peer-regularized feature recombination for arbitrary image
  style transfer.
\newblock In {\em IEEE Conf. Comput. Vis. Pattern Recog.}, 2020.

\bibitem{Ulyanov2016}
Dmitry Ulyanov, Vadim Lebedev, Andrea Vedaldi, and Victor Lempitsky.
\newblock Texture networks: Feed-forward synthesis of textures and stylized
  images.
\newblock volume~48 of {\em Proceedings of Machine Learning Research}, pages
  1349--1357, New York, New York, USA, 20--22 Jun 2016.

\bibitem{Wang2004}
Bin Wang, Wenping Wang, Huaiping Yang, and Jiaguang Sun.
\newblock Efficient example-based painting and synthesis of 2d directional
  texture.
\newblock {\em IEEE Transactions on Visualization and Computer Graphics},
  10(3):266–277, May 2004.

\bibitem{Wang2020CVPR}
Huan Wang, Yijun Li, Yuehai Wang, Haoji Hu, and Ming-Hsuan Yang.
\newblock Collaborative distillation for ultra-resolution universal style
  transfer.
\newblock In {\em IEEE Conf. Comput. Vis. Pattern Recog.}, 2020.

\bibitem{Wang2018CVPR}
Ting-Chun Wang, Ming-Yu Liu, Jun-Yan Zhu, Andrew Tao, Jan Kautz, and Bryan
  Catanzaro.
\newblock High-resolution image synthesis and semantic manipulation with
  conditional gans.
\newblock In {\em IEEE Conf. Comput. Vis. Pattern Recog.}, 2018.

\bibitem{Wang2020CVPR_a}
Zhizhong Wang, Lei Zhao, Haibo Chen, Lihong Qiu, Qihang Mo, Sihuan Lin, Wei
  Xing, and Dongming Lu.
\newblock Diversified arbitrary style transfer via deep feature perturbation.
\newblock In {\em IEEE Conf. Comput. Vis. Pattern Recog.}, 2020.

\bibitem{Xia2020ECCV}
Xide Xia, Meng Zhang, Tianfan Xue, Zheng Sun, Hui Fang, Brian Kulis, and Jiawen
  Chen.
\newblock Joint bilateral learning for real-time universal photorealistic style
  transfer.
\newblock In {\em Eur. Conf. Comput. Vis.}, 2020.

\bibitem{Yarlagadda2012}
Pradeep Yarlagadda and Bj{\"o}rn Ommer.
\newblock From meaningful contours to discriminative object shape.
\newblock In Andrew Fitzgibbon, Svetlana Lazebnik, Pietro Perona, Yoichi Sato,
  and Cordelia Schmid, editors, {\em Computer Vision -- ECCV 2012}, pages
  766--779, Berlin, Heidelberg, 2012. Springer Berlin Heidelberg.

\bibitem{Yim2020ECCV}
Jonghwa Yim, Jisung Yoo, Won joon Do, Beomsu Kim, and Jihwan Choe.
\newblock Filter style transfer between photos.
\newblock In {\em Eur. Conf. Comput. Vis.}, 2020.

\bibitem{Zhang2019}
Chi Zhang, Yixin Zhu, and Song-Chun Zhu.
\newblock Metastyle: Three-way trade-off among speed, flexibility, and quality
  in neural style transfer.
\newblock In {\em AAAI}, 2019.

\bibitem{Zheng2019}
Ningyuan Zheng, Y. Jiang, and Ding jiang Huang.
\newblock Strokenet: A neural painting environment.
\newblock In {\em Int. Conf. Learn. Represent.}, 2019.

\bibitem{Zhu2017ICCV}
Jun-Yan Zhu, Taesung Park, Phillip Isola, and Alexei~A. Efros.
\newblock Unpaired image-to-image translation using cycle-consistent
  adversarial networks.
\newblock In {\em Int. Conf. Comput. Vis.}, 2017.

\end{thebibliography}
}

\clearpage
\appendix
\section{Videos}\label{sec:video}

\FloatBarrier

 \begin{figure}
    \centering
    \includegraphics[width=0.45\textwidth]{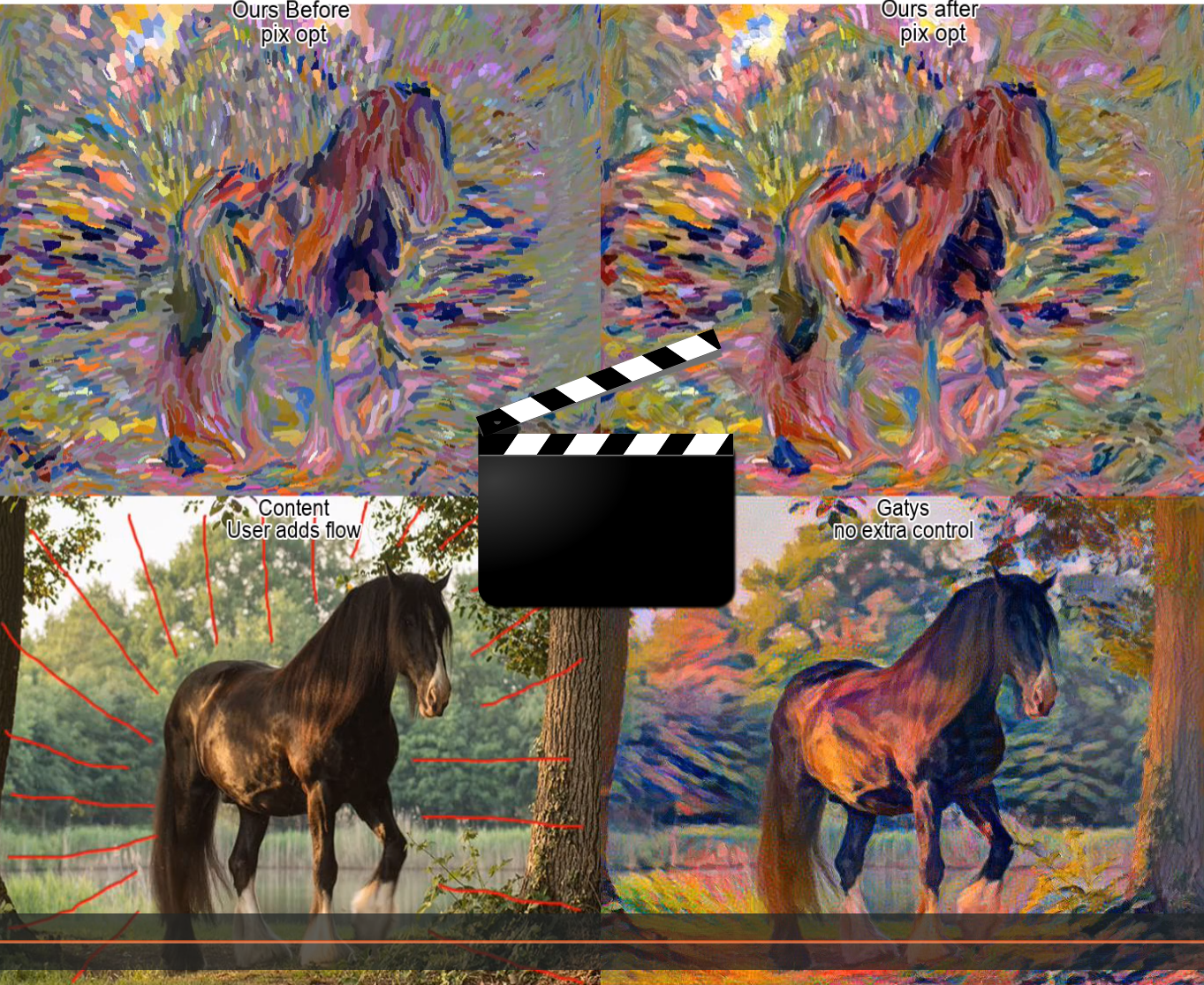}
    \caption{We provide videos that show how the stylization evolves over time. Moreover, we show how flow constraints change the stylization. See Sec.~\ref{sec:video}.}
    \label{fig:horse_teaser}
\end{figure}

In order to give insights into the stylization procedure, we provide two videos, see:
\begin{itemize}
    \item \url{https://heibox.uni-heidelberg.de/f/77c92a1355904de6a6be/}
    \item \url{https://heibox.uni-heidelberg.de/f/c17c112570f646bab081/}
\end{itemize}
 In these videos we show how the stylization evolves over time, before and after pixel optimization. We also compare to Gatys et al. \cite{Gatys2016} and show how the user input (Sec.~\ref{sec:drawing}) influences the stylization. See Fig.~\ref{fig:horse_teaser}.

\section{Controlling the Flow of Brushstrokes with User Input}\label{sec:drawing}
\begin{figure}[!htb]
    \centering
    \def\arraystretch{0.0}
    \setlength{\tabcolsep}{1.0pt}
    \begin{tabular}{c}
    \includegraphics[width=0.44\textwidth]{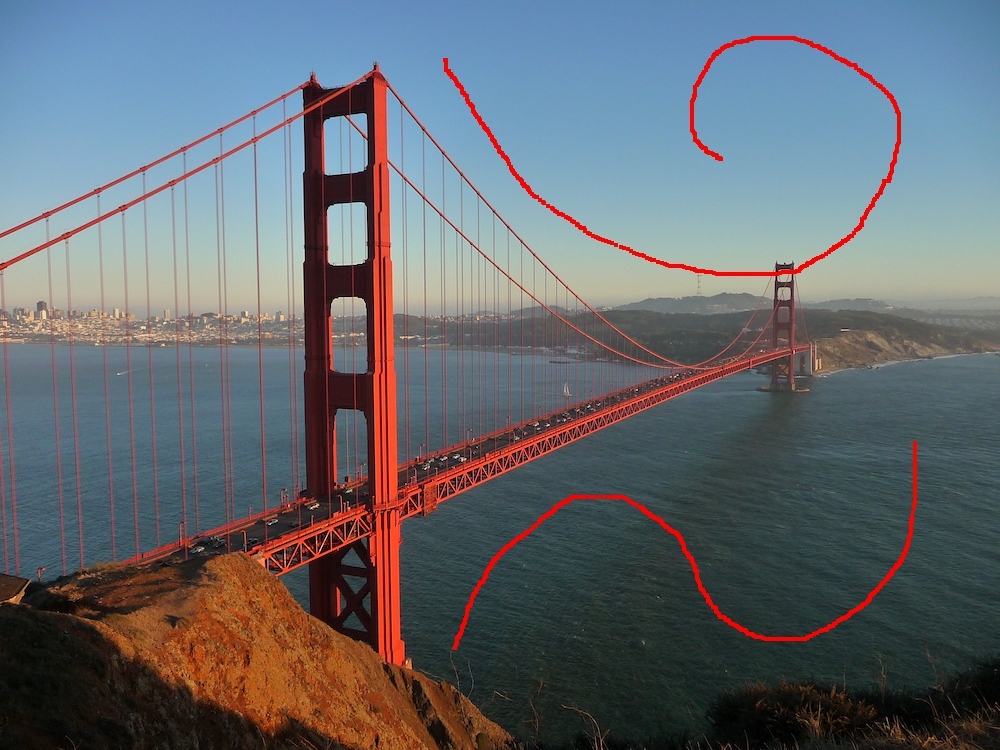} \\
    \includegraphics[width=0.44\textwidth]{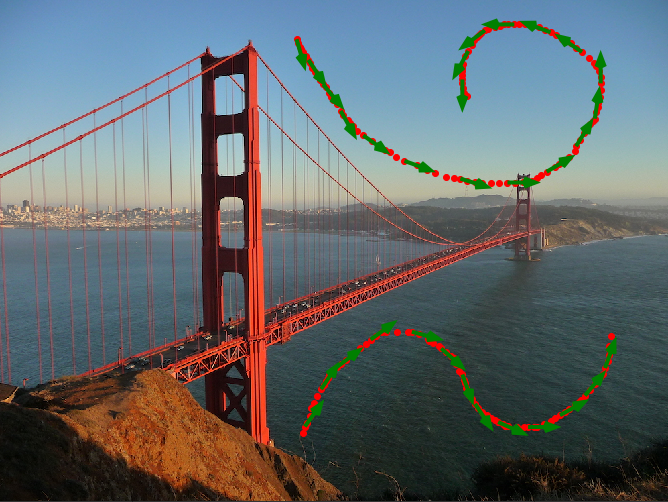}
    \end{tabular} \\
    \vspace{1pt}
    \caption{Top: A user draws arbitrary curves through a user interface. Bottom: A curve is represented as a set of points. For each point we can compute an approximate tangent vector.}
    \label{fig:user_input}
\end{figure}

In Sec. \ref{sec:res_control}, we showed how our method enables us to control the flow of brushstrokes. A user can draw arbitrary curves on the content image through a user interface and the brushstrokes in the stylized image will follow these curves. This can be achieved by adding a simple projection loss, which we will explain in this section. \\
Each drawn curve is represented as a set of points $P_1, P_2,...,P_M$ (do not confuse those with the control points for a B\'{e}zier curve). For each point $P_i$ the approximate tangent vector $\mathbf{v}_i$ is computed as follows:

\begin{equation}\label{eq:tangent_vector}
    \mathbf{v}_i = \frac{\mathbf{\tilde{v}}_i}{||\mathbf{\tilde{v}}_i||}, \quad \mathbf{\tilde{v}}_i = \left(\frac{1}{Q} \sum\limits_{j=1}^Q P_{i + j} \right) - P_i,
\end{equation}
where $Q = 3$ in all our experiments. Fig.~\ref{fig:user_input} shows user drawn curves with corresponding tangent vectors.

As explained in Sec. \ref{sec:approach}, each brushstroke is represented as a quadratic B\'{e}zier curve with additional parameters for location, width, and color. A quadratic B\'{e}zier curve is parameterized by three points: a start point, an end point, and a control point. Roughly speaking, the start and end points determine the curves orientation and the control point determines the curvature.

\begin{figure*}[!htb]
    \centering
    \includegraphics[width=0.9\textwidth]{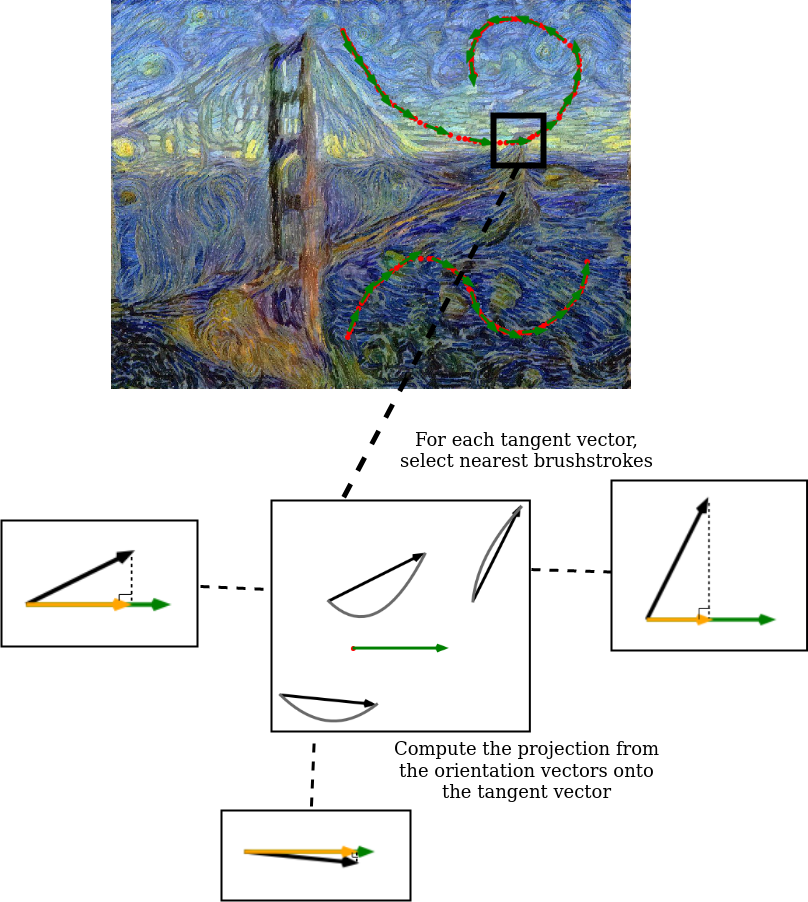}
    \caption{Overview of projection loss for a specific tangent vector. For each tangent vector (green), we select the closest brushstrokes (gray). We then take the orientation vector (black) for each brushstroke and compute the projection (yellow) onto the tangent vector.}
    \label{fig:curve_fit_diagram}
\end{figure*}

The projection loss is computed as follows: 

\begin{enumerate}
    \item Compute for each brushstroke the vector from the start point to the end point of the B\'{e}zier curve, see Fig.~\ref{fig:brushstroke_parametrization}. We will refer to this vector as the \textit{orientation vector} of a brushstroke.
    
    \item For each tangent vector, compute the $L$ nearest brushstrokes on the canvas. $L$ is a hyperparameter and determines the range of the brushstrokes that will be affected by the drawn curves. Fig.~\ref{fig:control_strength} shows the influence of $L$ on the stylization.
    
    \item For each tangent vector, compute the projection of the orientation vectors from the nearest brushstrokes onto the tangent vector. Both the tangent vectors and the orientation vectors are normalized to unit length.
    
    \item The projection loss encourages the absolute value of these projections to be 1. Since all vectors are normalized, the absolute value of the projections will be 1 if and only if the orientation vectors are parallel to the tangent vector. See Fig.~\ref{fig:curve_fit_diagram} for an overview of the whole computation. 
\end{enumerate}

See Fig.~\ref{fig:user_control_1} and ~\ref{fig:user_control_2} for more results.

\section{Renderer}\label{sec:renderer}
\subsection{Brushstroke Parameterization}
\begin{figure}
    \centering
    \includegraphics[width=0.35\textwidth]{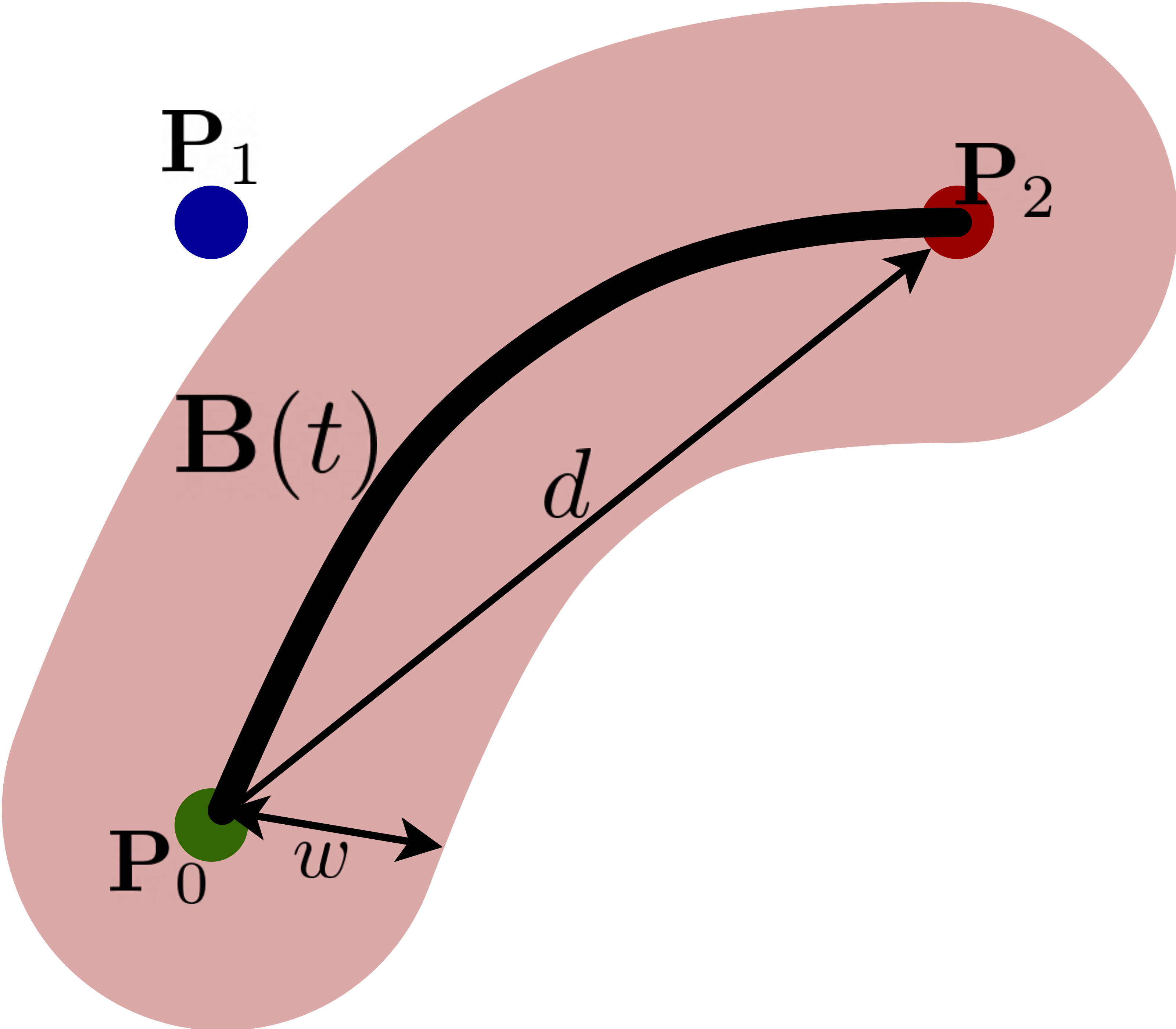}
    \caption{The brushstroke is parameterized by color $rgb \in \mathbb{R}^3$, width $w \in \mathbb{R}$ and B\'{e}zier curve $\mathbf{B}(t)$. The B\'{e}zier curve is defined by points $\mathbf{P}_0,\ \mathbf{P}_1,\ \mathbf{P}_2 \in \mathbb{R}^2$ and position on a curve $t \in [0;1]$. The direction (orientation) vector $d \in \mathbb{R}^2$ is used to simplify the strokes.}
    \label{fig:brushstroke_parametrization}
\end{figure}

Each brushstroke is parameterized by color $rgb \in \mathbb{R}^3$, B\'{e}zier curve $\mathbf{B}(t)$ with $t \in [0;1]$ and width $w \in \mathbb{R}$. B\'{e}zier curve $\mathbf{B}(t)$ is introduced in Eq. \ref{eq:bezier_curve}. From the formulation we see that $\mathbf{B}(t)$ depends on three points $\mathbf{P}_0,\ \mathbf{P}_1,\ \mathbf{P}_2 \in \mathbb{R}^2$ which we further call start point, control point and end point, respectively.
We additionally define the direction (orientation) $d \in \mathbb{R}^2$ of a stroke as $d:=\mathbf{P}_2 - \mathbf{P}_0$. This vector will be used in Sec.~\ref{sec:drawing} to control the flow of the brushstrokes.
This parameterization of a brushstrokes is illustrated in Fig.~\ref{fig:brushstroke_parametrization}.

\subsection{Tensor of Distances}\label{sec:distances}
As described in Alg. \ref{alg:render_algo}, the cornerstone of our rendering mechanism is a tensor of distances $D$ between every sampled point on a brushstroke and each point on a canvas. We use a canvas $\mathcal{C}$ of size $H \times W$, where $H=W=256$. We typically draw $N=5000$ brushstrokes and on every brushstrokes we sample $S=10$ points. This results in a tensor of shape $H \times W \times N \times S$. If we use \texttt{float32} data type taking 4 Bytes, then the distances tensor has size $256*256*5000*10*4=13107200000 \approx 13$GB, which is infeasible in practice. 
However, we actually do not need to compute the distances between every pixel and every brushstroke since a pixel is only affected by a few nearby brushstrokes, say by $K$ nearest brushstrokes (in our implementation we set $K=20$). With this in mind, we can reduce the tensor of distances $D$ of shape $H \times W \times N \times S$ to the size $H \times W \times K \times S$ which requires $\frac{N}{K}=\frac{5000}{20}=250$ times less memory, roughly $52$MB. 

In order to accomplish the tensor size reduction, we need to assign the $K$ nearest brushstrokes to each pixel. However, for this we would need the tensor of distances of size $H \times W \times N$, which is not feasible for large values of $N$. This problem can be circumvented if we compute the distances from each of the $N$ brushstrokes to a sparse subset of ``anchor'' points on the tensor of locations $\mathcal{C} \in \mathbb{R}^{H \times W}$. We create a tensor $\mathcal{C}_{\textnormal{coarse}}$ of size $H' \times W'$ where $H'<H$ and $W'<W$ containing subset of tensor $\mathcal{C}$, in our case we set $H'=0.1 \cdot H$ and $W'=0.1 \cdot W$. Now we can effectively compute the tensor of distances between $\mathcal{C}_{\textnormal{coarse}}$ and each brushstroke, it will have shape $H' \times W' \times N$. We left out dimension $S$ because we only need to roughly estimate the distances of the brushstrokes that are close to the location, so we use the location coordinates of the whole brushstroke. Now we extract the $K$ nearest strokes at every location and obtain a tensor of indices $idcs'$ having shape $H' \times W' \times K$. We then apply nearest neighbor upsampling to $idcs'$ across dimensions $H$ and $W$ and obtain $idcs$, a tensor of shape $H \times W \times K$. Thus, every pixel will have the same nearest neighbors as the nearest ``anchor'' point. This tensor of indices indicates the $K$ nearest brushstrokes for each pixel of the canvas. Now using this tensor of indices $idcs$ and the \texttt{tf.gather} operation in TensorFlow we can effectively assign to every location only the $K$ nearest strokes. We note that the same stroke is assigned to multiple locations but this does not hinder the optimization process because the stroke will just receive more gradient information.

\subsection{Hardware, Runtime, Memory}
Our stylization process consists of two stages. At the first stage we optimize brushstroke parameters, at the second stage we optimize individual pixels.

\textbf{Brushstroke parameters optimization.} We use a canvas of size $H \times W$, where $H=W=256$ and optimize using our renderer for 1000 steps using the Adam Optimizer \cite{adam2014}. It takes around 3 minutes.

\textbf{Pixel optimization.} Now we upsample the canvas with fitted strokes to have the smallest image side of 1024px and keep the input content image aspect ratio. This image with fitted brushstrokes is used as both content image and initialization for the standard Gatys et al. stylization routine. We optimize for another 1000 steps using the Adam optimizer. It takes 4 more minutes to converge.
All the experiments are conducted on NVIDIA TitanXP or NVIDIA 2080Ti graphic cards.

\subsubsection{Number of Brushstrokes}
The memory consumption does not depend on the number of strokes, see Sec. \ref{sec:distances}. However, the run-time reduces linearly as the number of strokes increases, see Tab. \ref{tab:runtime}.\\

\begin{table}[b]
    \setlength{\tabcolsep}{2.3pt}
    \scriptsize

    \caption{Run-time and memory analysis. Experiment conducted on a TITAN Xp GPU.}
    \centering
    \begin{tabular}{l||ccccccccc}
    \hline\noalign{\smallskip}
    \# Strokes & 1K & 5K & 10K & 15K & 20K \\
    \noalign{\smallskip}
    \hline
    \noalign{\smallskip}
    
    Speed [iter/s] & 1.17 $\pm$ 0.01 & 1.16 $\pm$ 0.01 & 1.08 $\pm$ 0.01 & 1.0 $\pm$ 0.01 & 0.93 $\pm$ 0.01 \\
    Memory [GB] & 9550 & 9550 & 9550 & 9550 & 9550 \\

    \hline
      
    \end{tabular}

    \label{tab:runtime}
\end{table}

\section{Trained Renderer}\label{sec:trained_neural_renderer}
\begin{figure}
    \centering
    \includegraphics[width=0.5\textwidth]{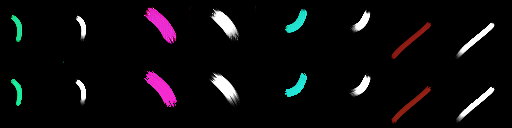}
    \caption{Generated brushstrokes using the trained renderer. Top row: generated brushstrokes. Bottom row: ground truth simulated in the FluidPaint environment. The alpha mask is always white and located to the right of the brushstroke.}
    \label{fig:gener_brushstrokes}
\end{figure}

In order to ablate our renderer, we trained a neural network that receives brushstroke parameters as input and generates the corresponding brushstrokes. The brushstrokes are parameterized as described in Sec. \ref{sec:approach}. The network generates an RGB image of a brushstroke as well as an alpha mask. In order to render $N$ brushstrokes onto a canvas, we first have to generate each brushstroke individually and then blend them together using the alpha masks. See Fig. \ref{fig:gener_brushstrokes} for some generated brushstrokes.

\subsection{Architecture}
The brushstroke generator follows the StyleGAN architecture \cite{karras2019style} and consists of a mapping network $f$ and a synthesis network $g$, here we adopt the notation from the StyleGAN paper. The mapping network $f$ takes in the brushstroke parameters $z$ and processes them using $4$ fully-connected layers to create the latent vector $w$. The synthesis network $g$ consists of 4 blocks, each consisting of an upsampling layer, a 3x3 convolutional layer, and an AdaIN layer \cite{Huang2017}. The latent vector $w$ is injected into the AdaIN layers using learned affine transformations. After every convolutional layers we also inject noise. 
\\
The discriminator follows the StyleGAN architecture \cite{karras2019style} as well, however, it only consists of 8 layers.

\subsection{Training}
For training, we used the Wasserstein GAN loss \cite{Arjovsky2017} with gradient penalty \cite{Gulrajani2017} and a L2 loss (equally weighted). We used the Adam optimizer \cite{adam2014} with learning rate 0.0002.

\section{User Study}
To evaluate the quality of the synthesized images we use two methods. First, we compute the deception score, as suggested by Sanakoyeu et al. \cite{Sanakoyeu2018}. This score indicates how similar is a given stylization to the actual style of the artist. Another way to evaluate the quality of images is to perform a user study. We show to a human subject cropouts from images obtained using different stylization approaches or real artworks and ask them to pick crops from a real artwork. Among the stylization approaches we have Gatys et al. \cite{Gatys2016}, AST by Sanakoyeu et al. \cite{Sanakoyeu2018}, WCT \cite{Li2017_a}, and AdaIN \cite{Huang2017}. At once we show 4 images: each drawn randomly and independently. In Fig.~\ref{fig:user_study_random_patches} we present randomly drawn example trials. Note that in Fig.~\ref{fig:user_study_random_patches} we do not provide images for WCT \cite{Li2017_a} and AdaIN \cite{Huang2017} since those are easy to spot in most cases. 

\section{Additional Results}\label{sec:add_results}
\begin{figure*}[!htb]
    \centering
    \includegraphics[width=1.0\textwidth]{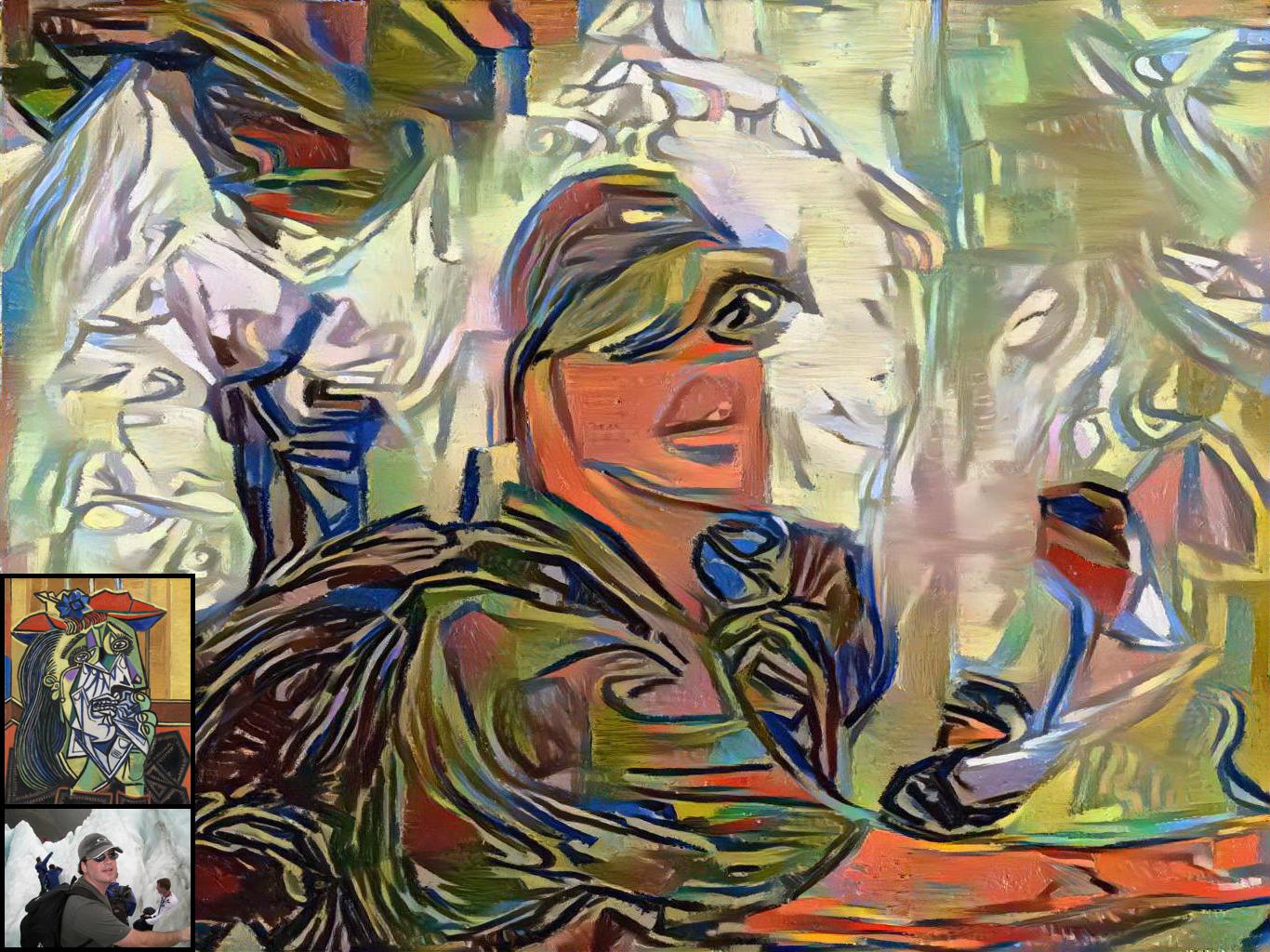} 
    \caption{Style image: ``The Weeping Woman'' by Pablo Picasso.}
    \label{fig:picasso_guy}
\end{figure*}

\begin{figure*}[!htb]
    \centering
    \includegraphics[width=1.0\textwidth]{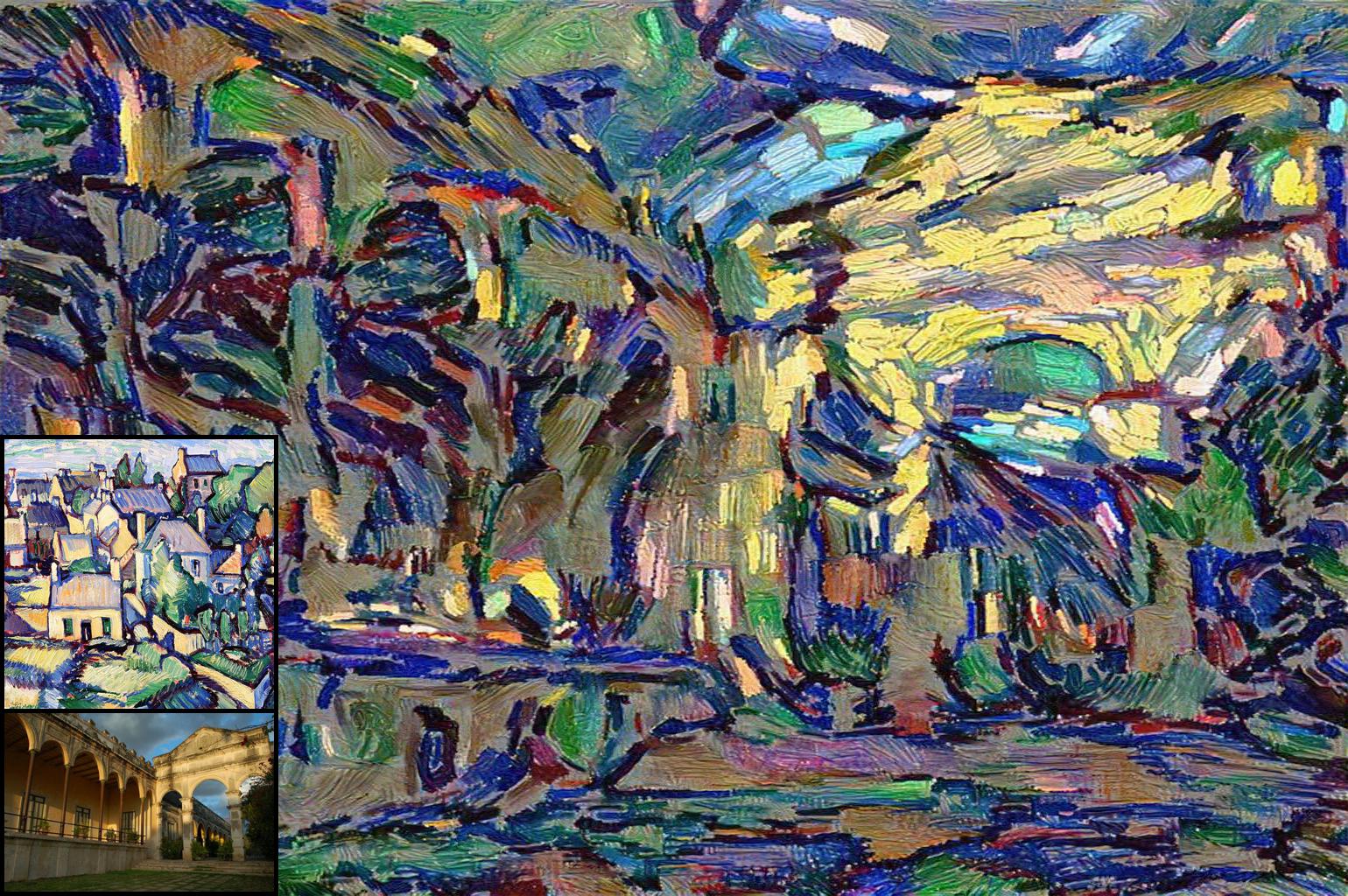} 
    \caption{Style image: ``Ile De Br\'{e}hat'' by Samuel John Peploe.}
    \label{fig:peploe_rome}
\end{figure*}

\begin{figure*}[!htb]
    \centering
    \includegraphics[width=1.0\textwidth]{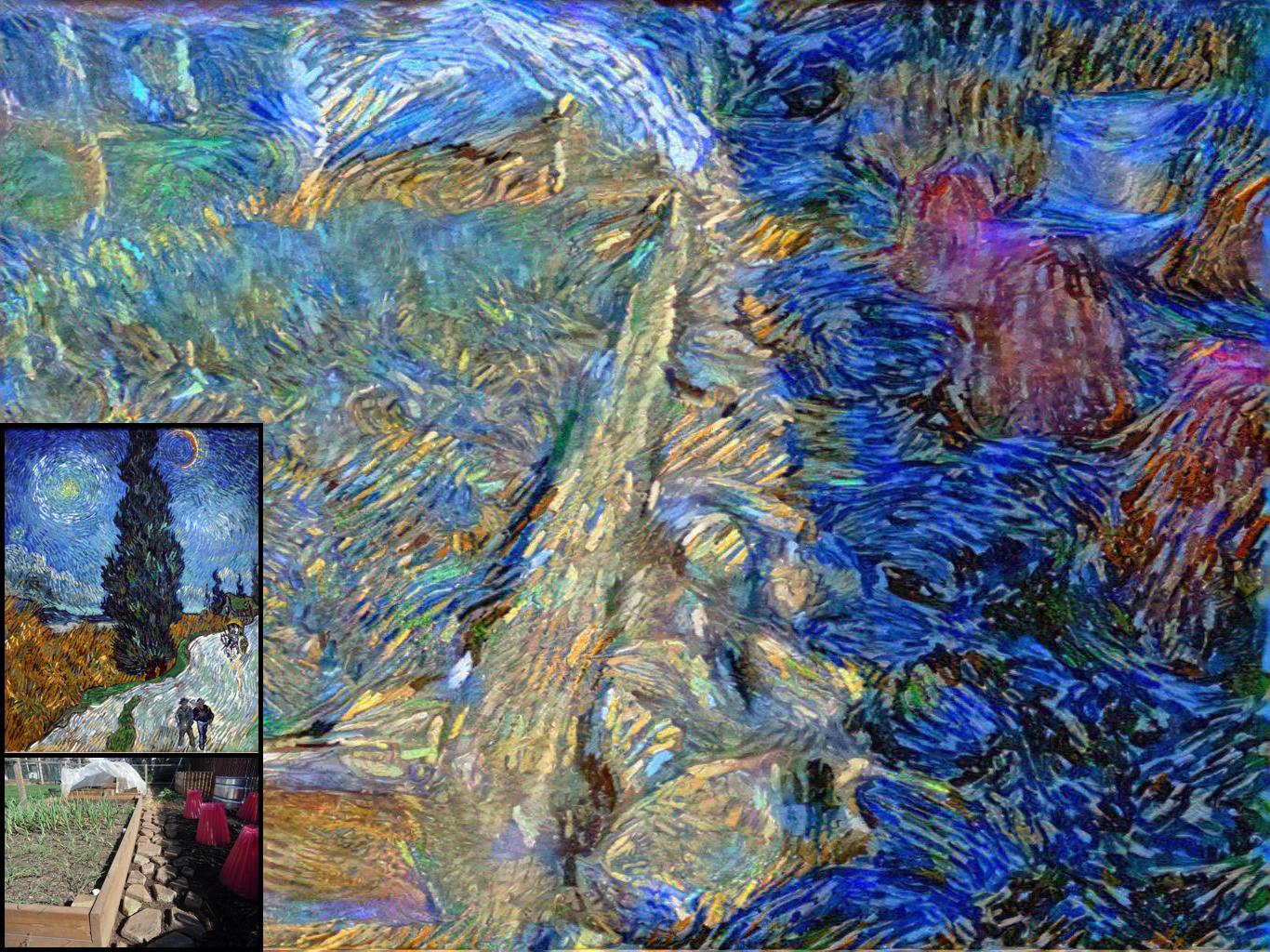} 
    \caption{Style image: ``Road with Cypress and Star'' by Vincent van Gogh.}
    \label{fig:van_gogh_garden}
\end{figure*}

\begin{figure*}[!htb]
    \centering
    \includegraphics[width=1.0\textwidth]{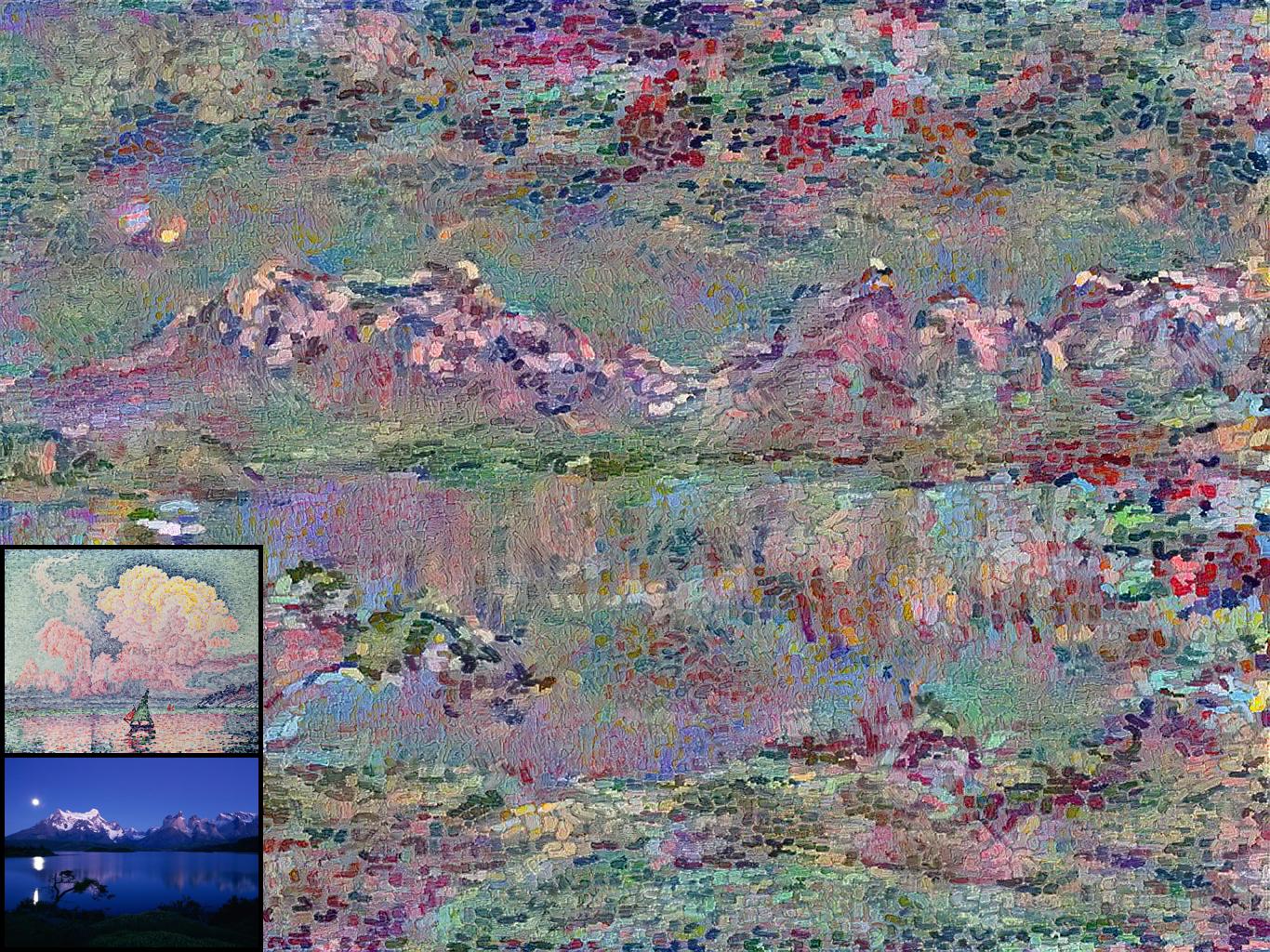} 
    \caption{Style image: ``Antibes, the Pink Cloud'' by Paul Signac.}
    \label{fig:signac_mountains}
\end{figure*}

\begin{figure*}[!htb]
    \centering
    \includegraphics[width=1.0\textwidth]{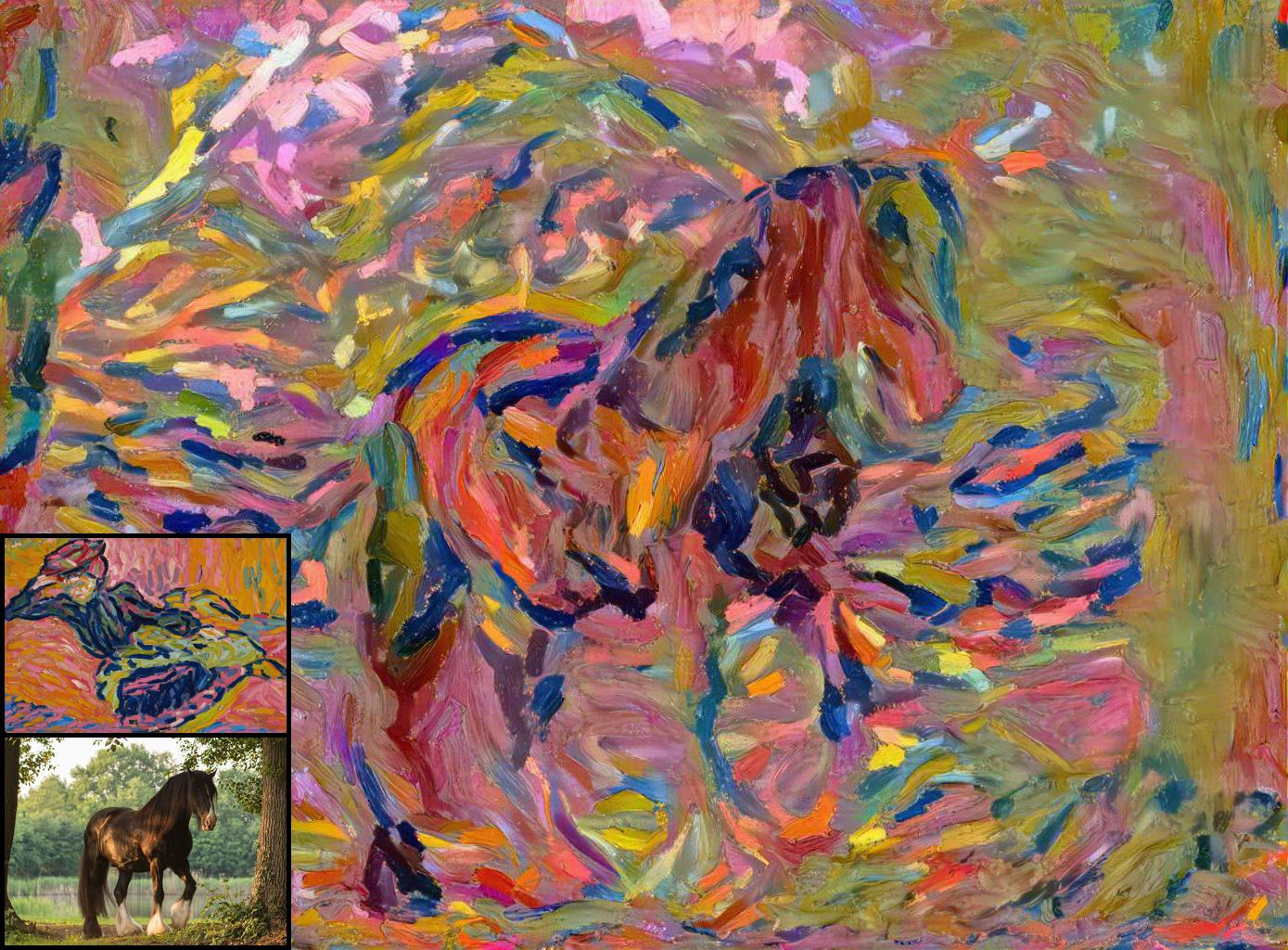} 
    \caption{Style image: ``Girl on a Divan'' by Ernst Ludwig Kirchner.}
    \label{fig:horse_kirchner}
\end{figure*}

\begin{figure*}[!htb]
    \centering
    \includegraphics[width=1.0\textwidth]{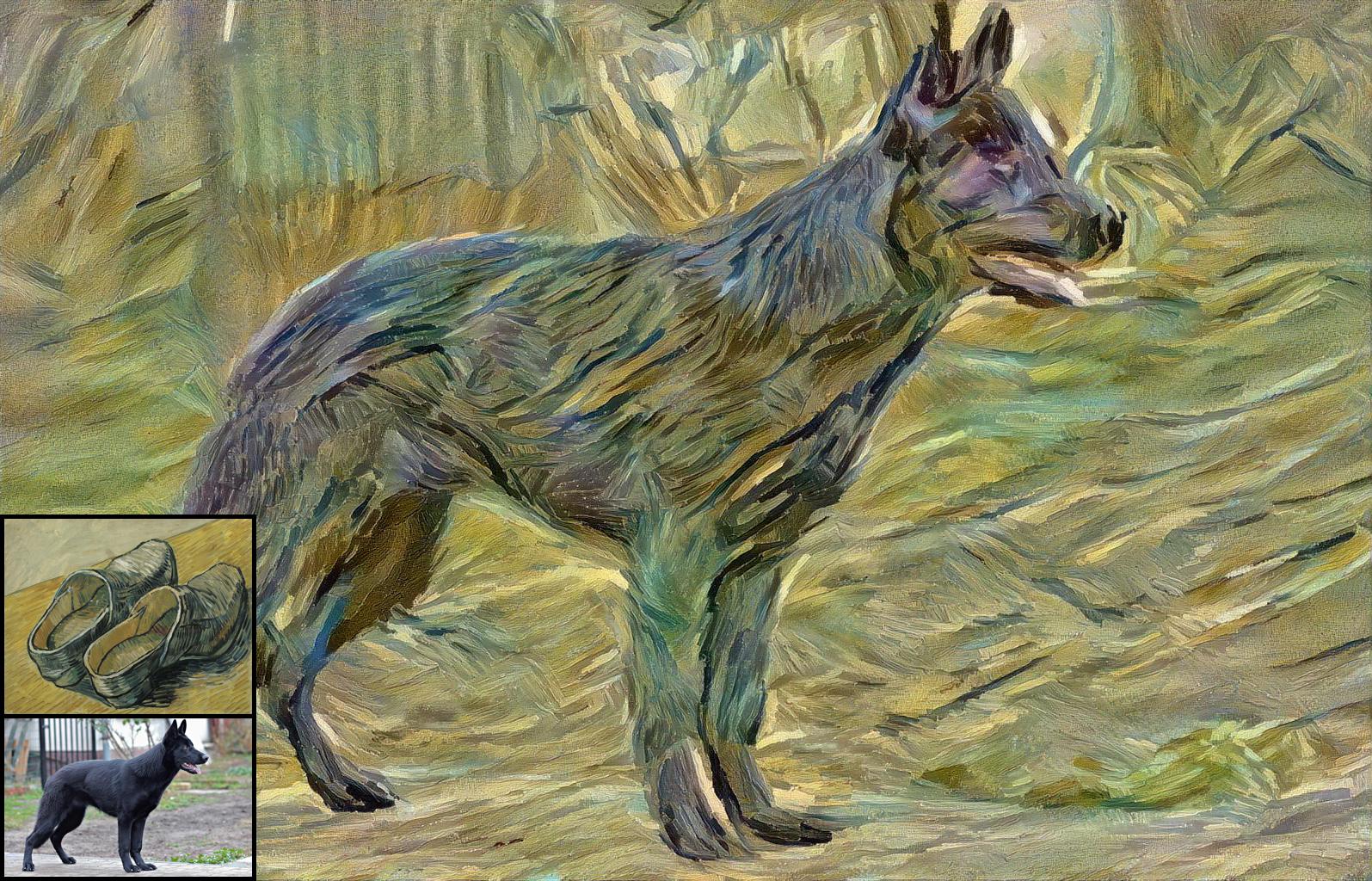} 
    \caption{Style image: ``A Pair of Leather Clogs'' by Vincent van Gogh.}
    \label{fig:black_german_shepard_van_gogh}
\end{figure*}

\begin{figure*}[!htb]
    \centering
    \includegraphics[width=1.0\textwidth]{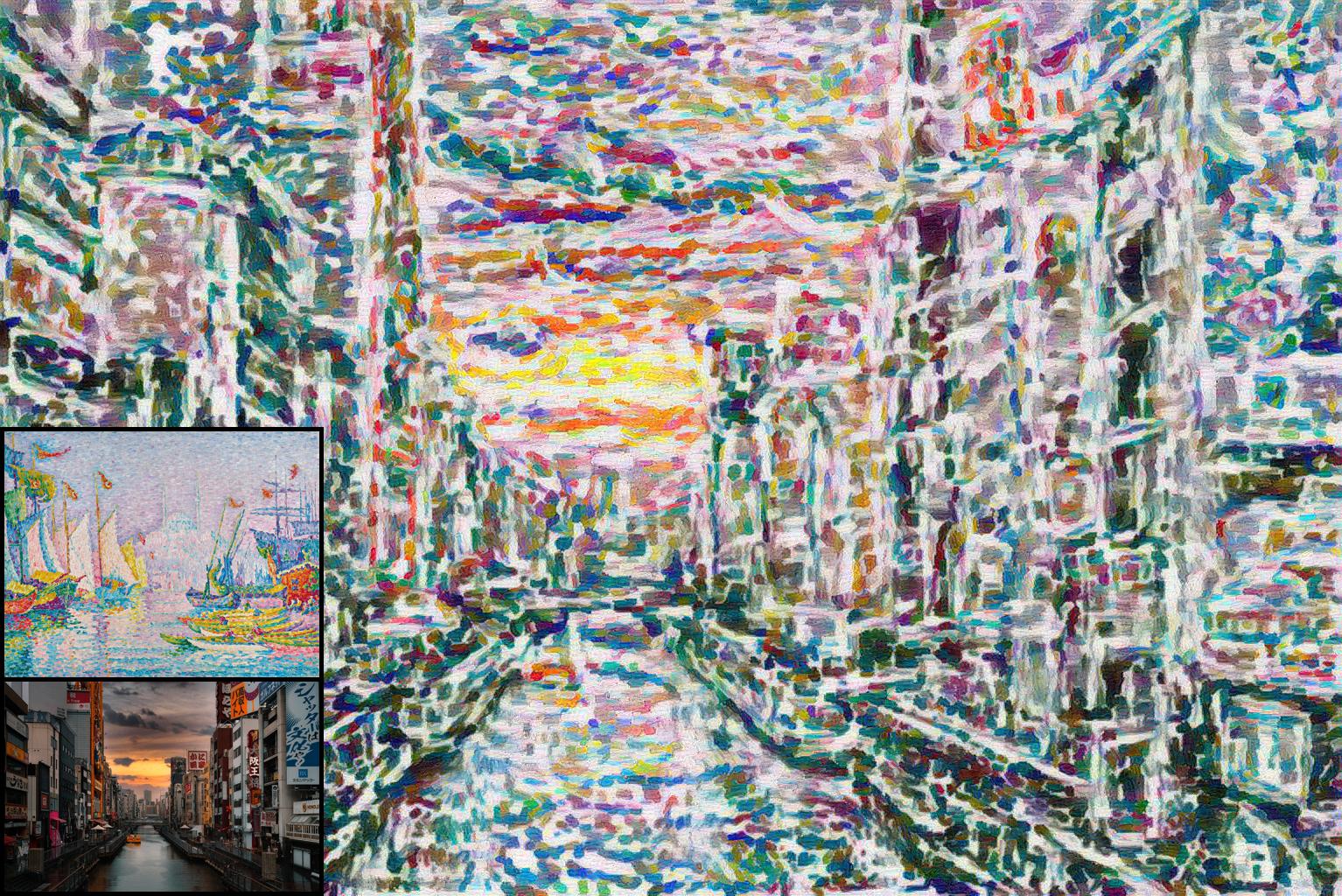} 
    \caption{Style image: ``La Corne d'Or'' by Paul Signac.}
    \label{fig:river_city_signac}
\end{figure*}

\begin{figure*}[!htb]
    \centering
    \includegraphics[width=1.0\textwidth]{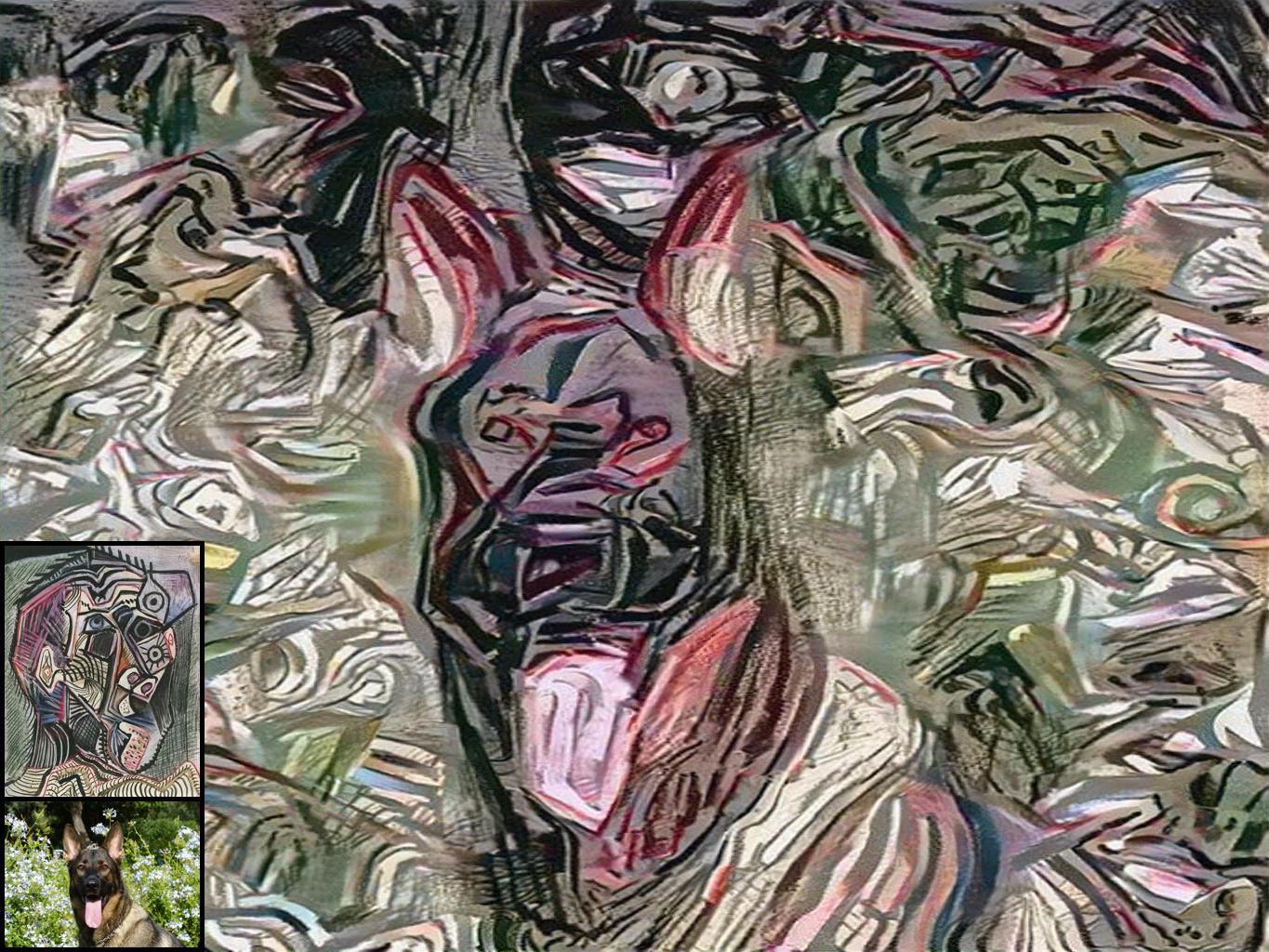} 
    \caption{Style image: ``Self Portrait'' by Pablo Picasso.}
    \label{fig:german_shepard_picasso}
\end{figure*}

\begin{figure*}[!htb]
    \centering
    \includegraphics[width=1.0\textwidth]{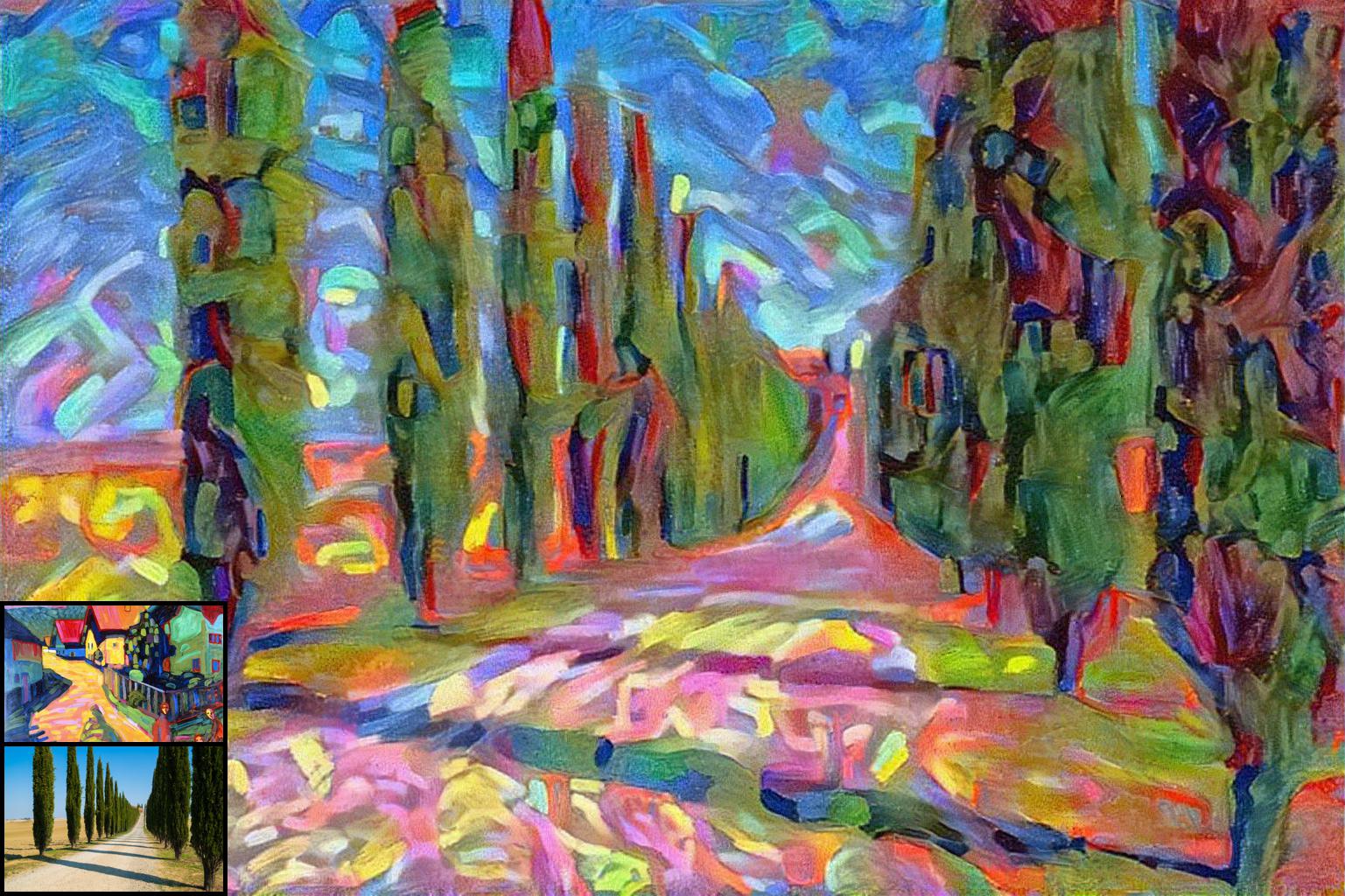} 
    \caption{Style image: ``Murnau Street With Women''
 by Wassily Kandinsky.}
    \label{fig:tuscany_kandinsky_women}
\end{figure*}

\begin{figure*}[!htb]
    \centering
    \includegraphics[width=1.0\textwidth]{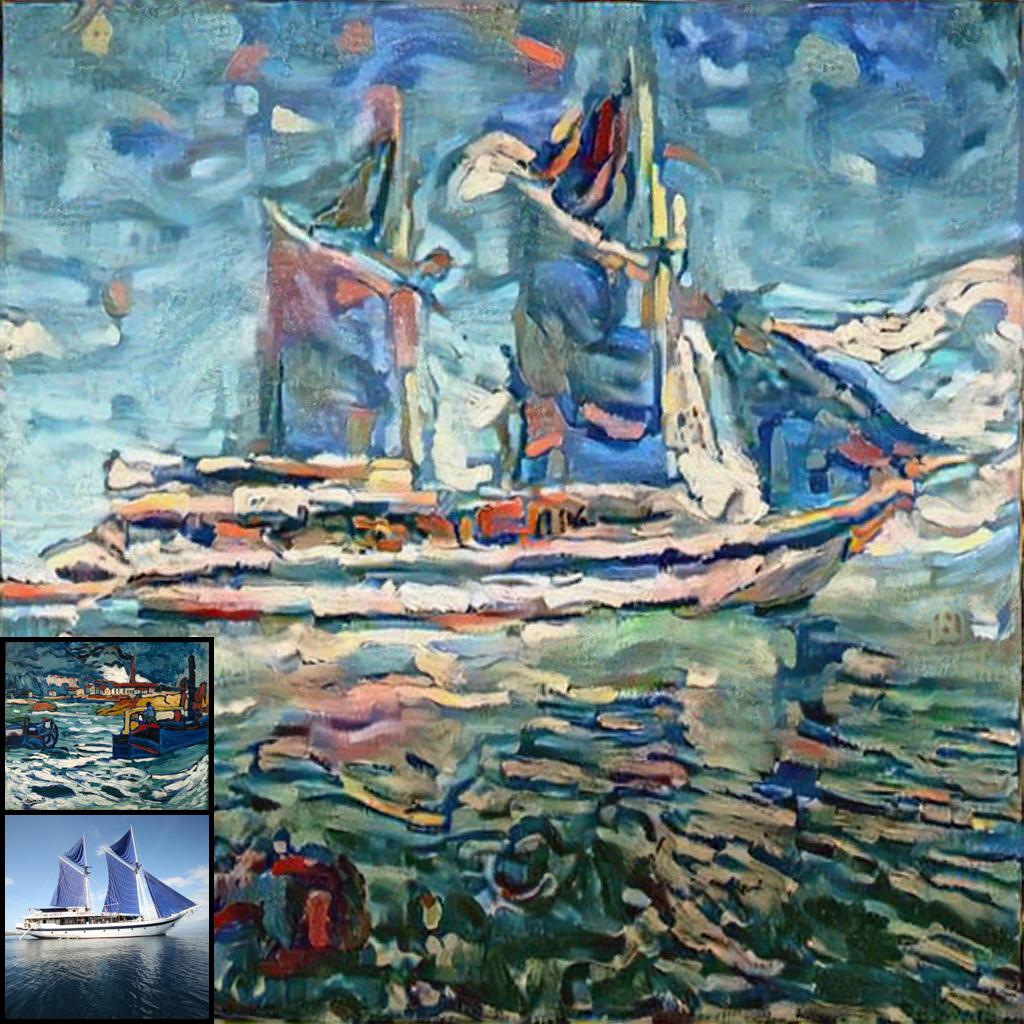}
    \caption{Style image: ``Barges on the Seine'' by Maurice de Vlaminck.}
    \label{fig:sailboat_vlamnick}
\end{figure*}

\begin{figure*}[!htb]
    \centering
    \includegraphics[width=1.0\textwidth]{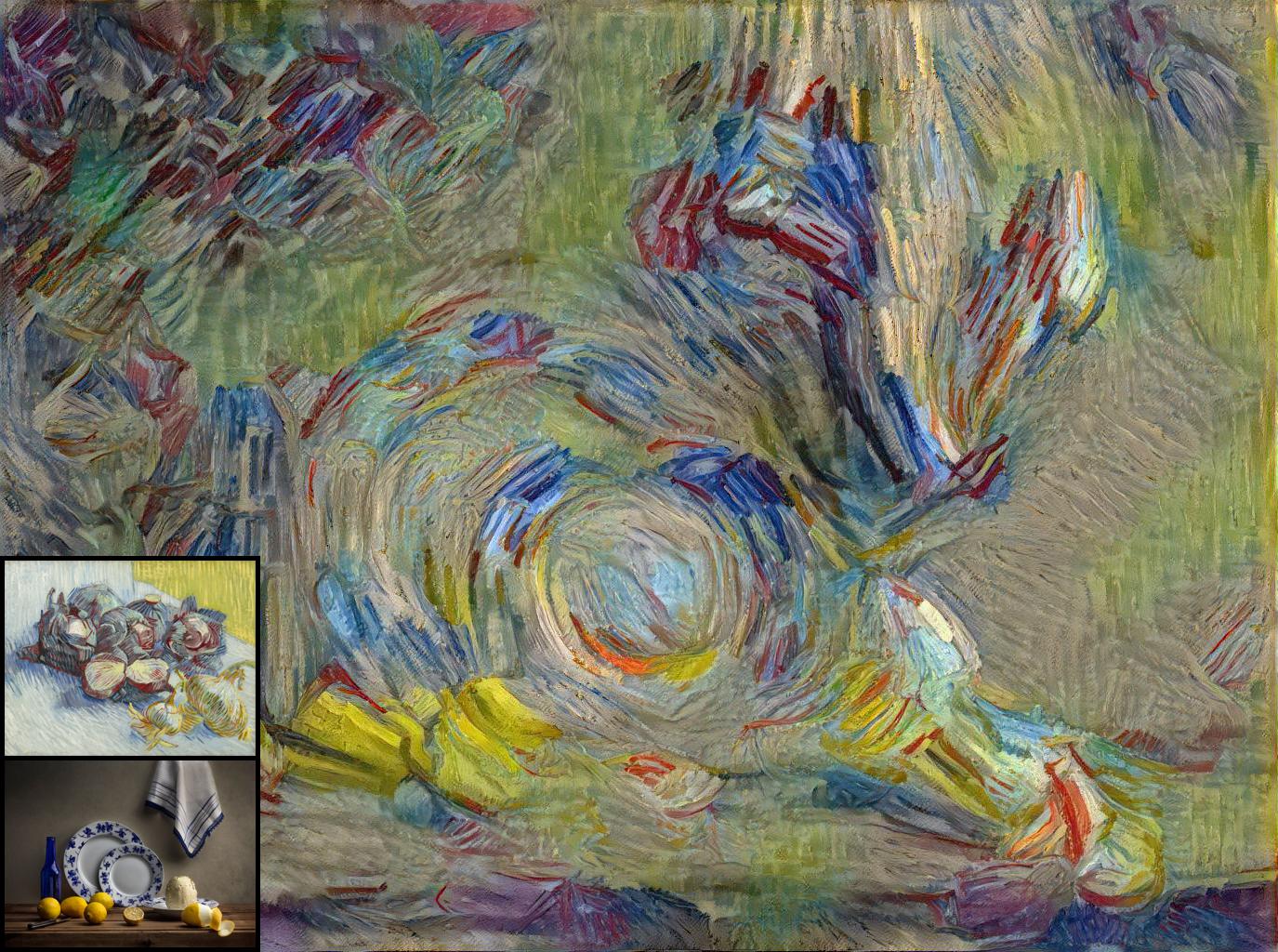}
    \caption{Style image: ``Red Cabbages and Onions'' by Vincent van Gogh.}
    \label{fig:still_life_red_cab_van_gogh}
\end{figure*}

\begin{figure*}[!htb]
    \centering
    \includegraphics[width=1.0\textwidth]{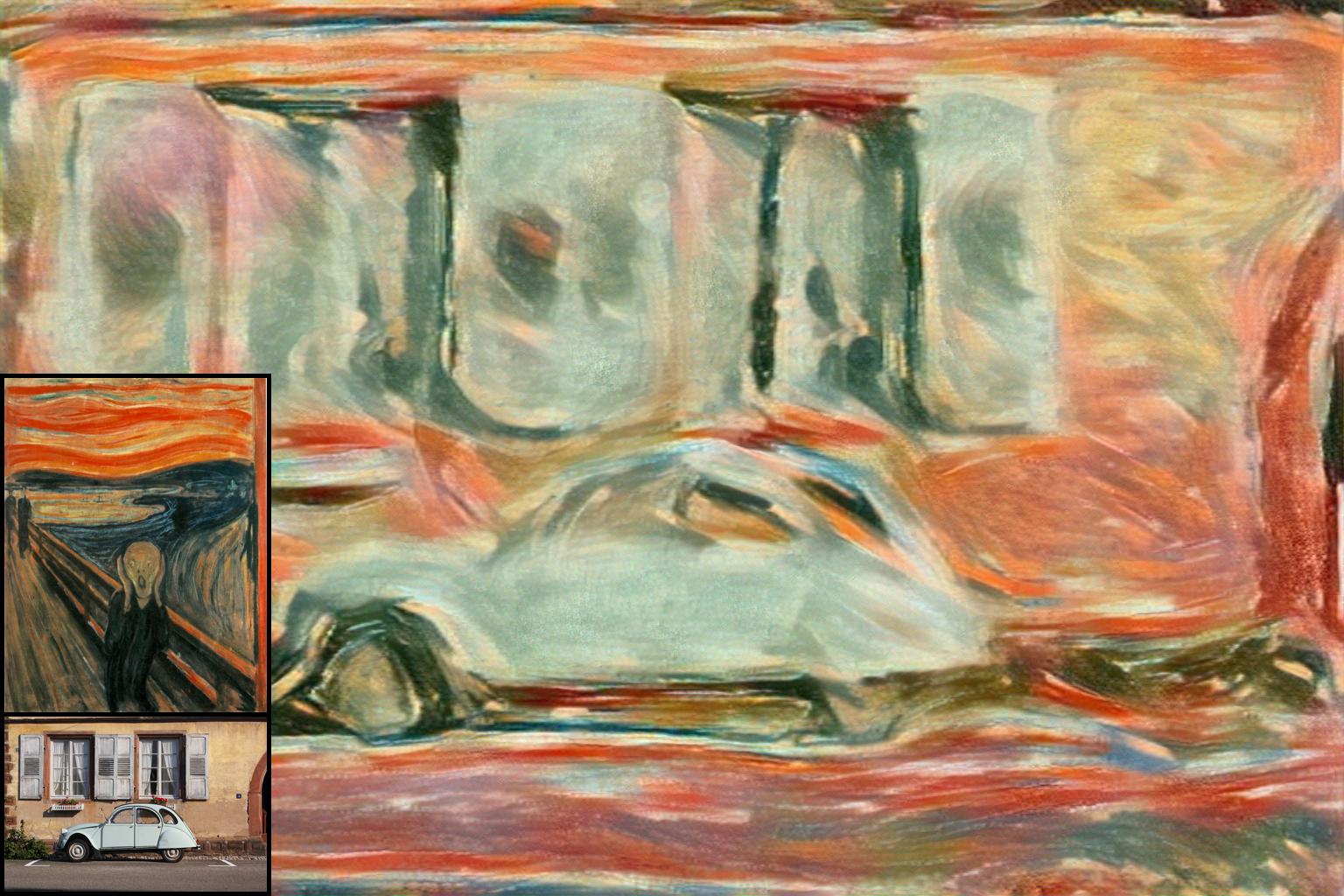}
    \caption{Style image: ``The Scream'' by Edvard Munch.}
    \label{fig:car_munch_scream}
\end{figure*}

\begin{figure*}[!htb]
    \centering
    \includegraphics[width=1.0\textwidth]{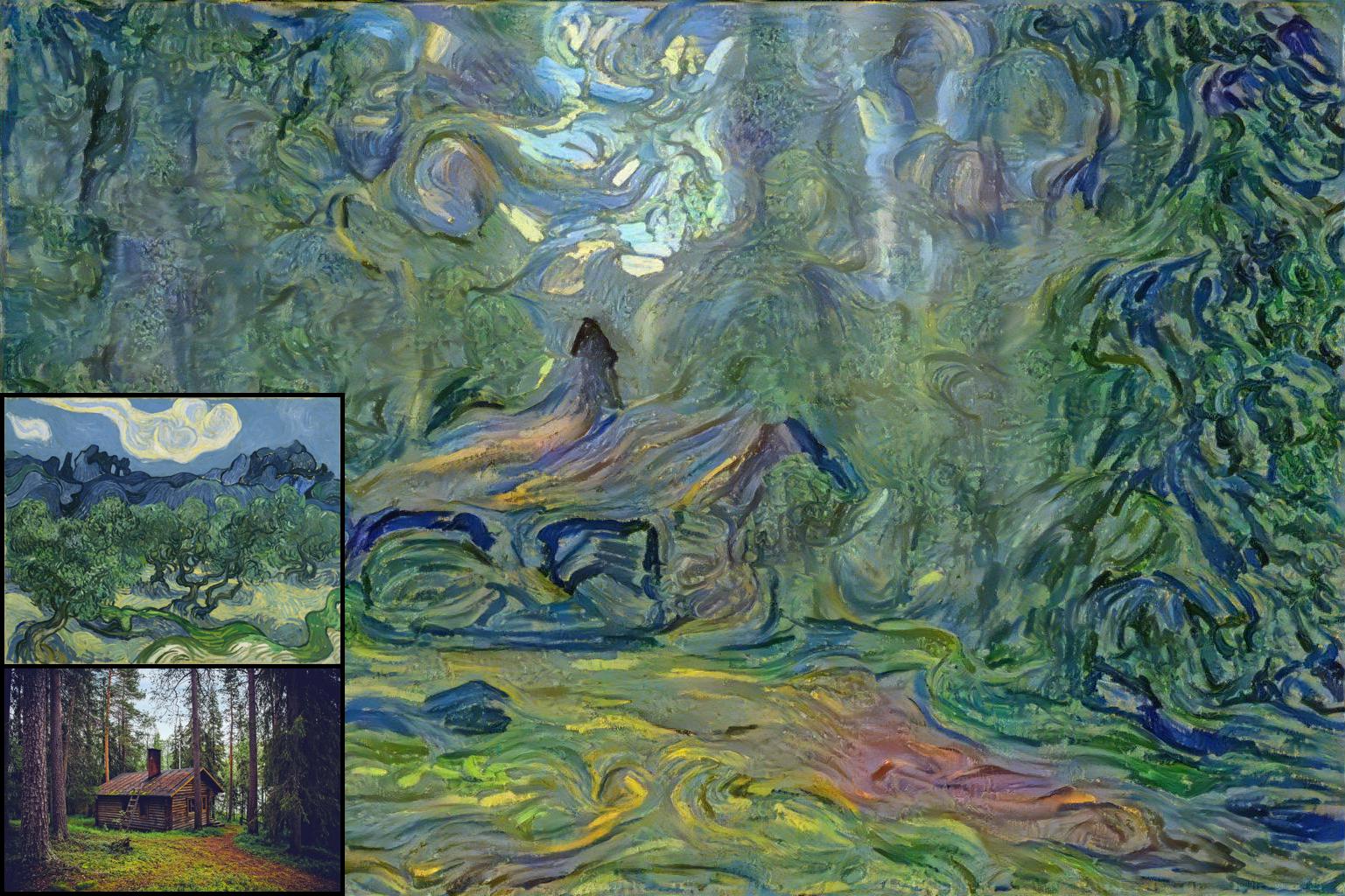}
    \caption{Style image: ``The Olive Trees'' by Vincent van Gogh.}
    \label{fig:cabin_van_gogh_trees}
\end{figure*}

\begin{figure*}[!htb]
    \centering
    \includegraphics[width=1.0\textwidth]{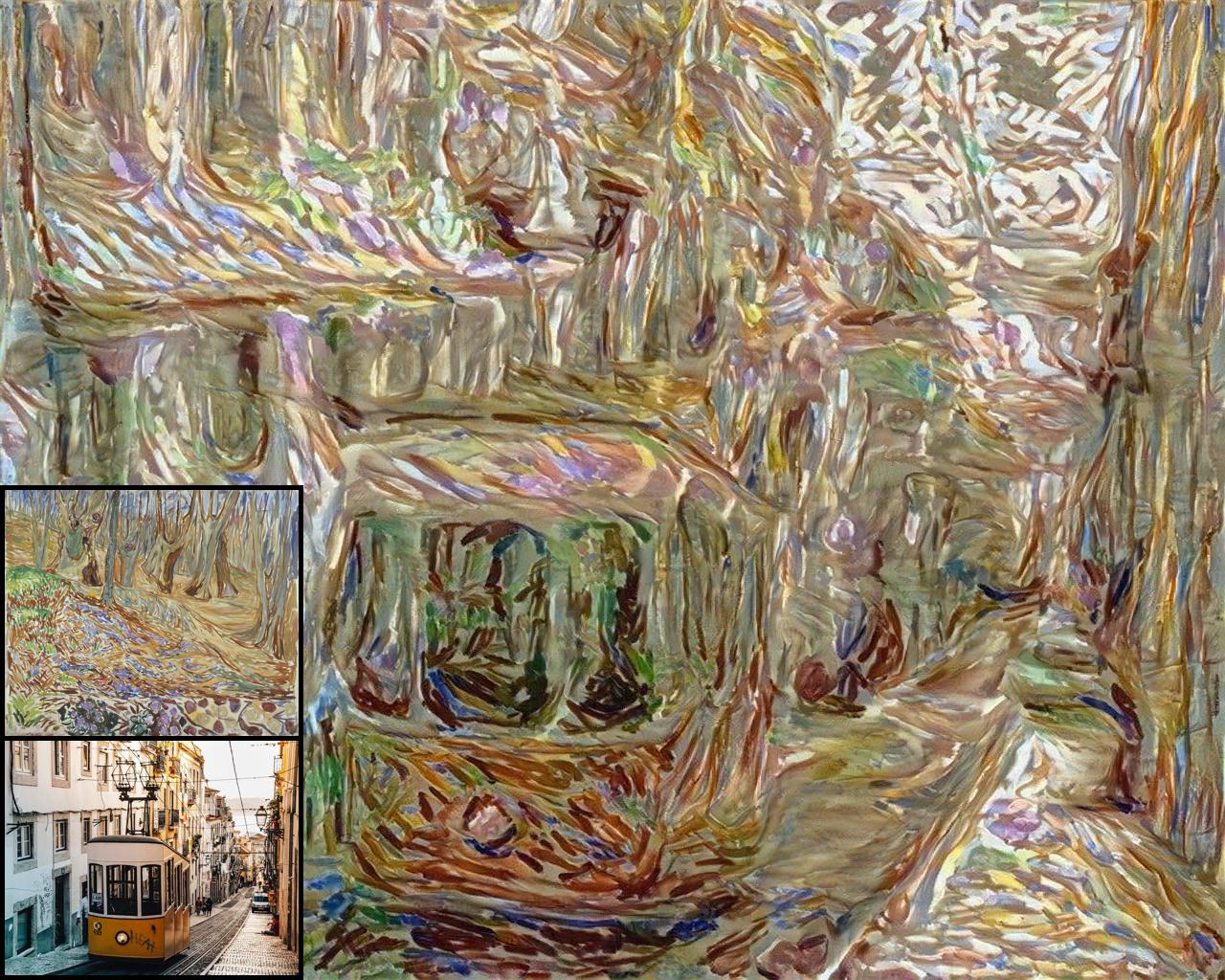}
    \caption{Style image: ``Spring in the Elm Forest'' by Edvard Munch.}
    \label{fig:lisbon_munch_spring}
\end{figure*}

\begin{figure*}[!htb]
    \centering
    \includegraphics[width=1.0\textwidth]{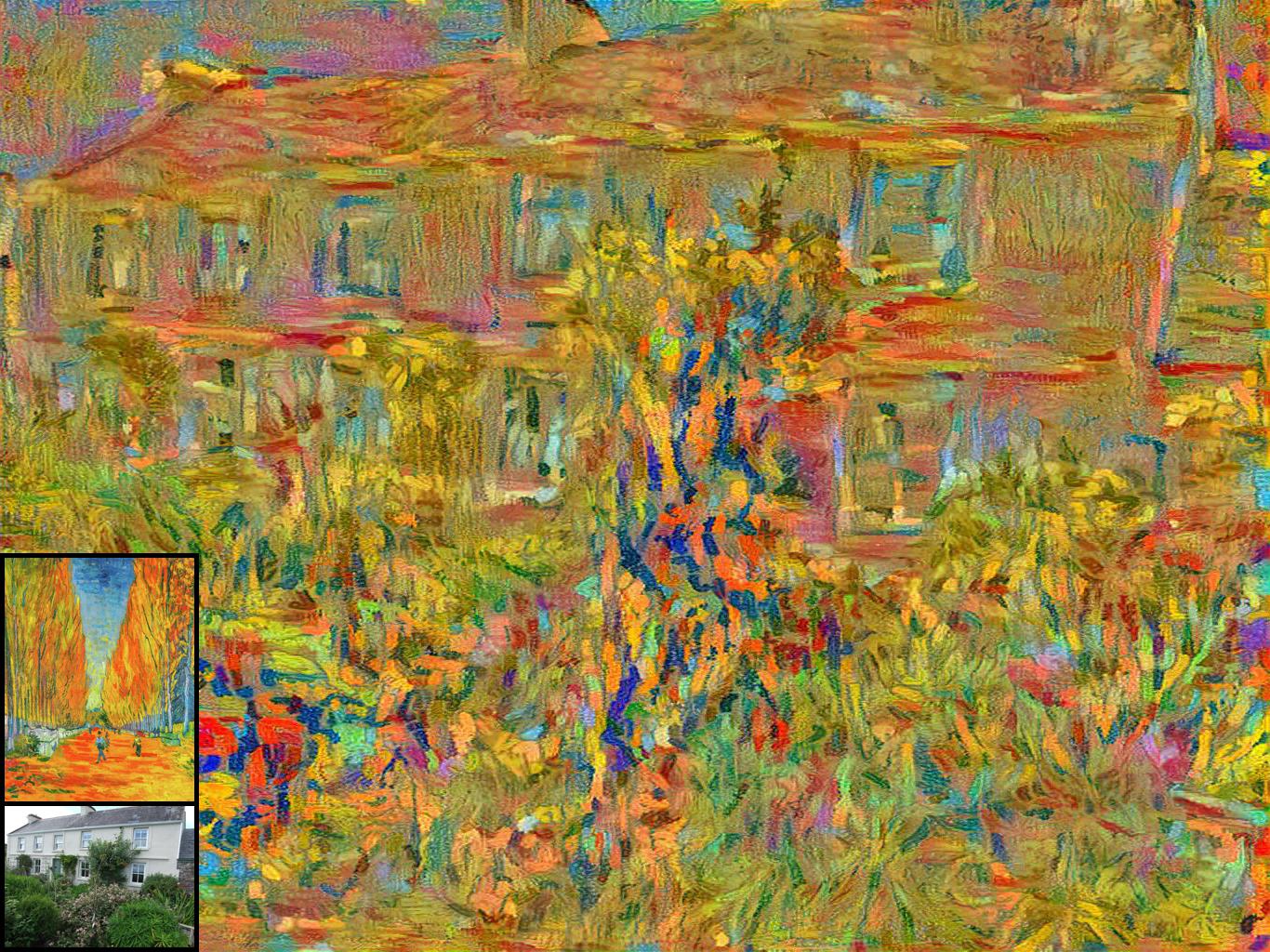}
    \caption{Style image: ``Les Alyscamps'' by Vincent van Gogh.}
    \label{fig:house_van_gogh_alyscamps}
\end{figure*}

\begin{figure*}[!htb]
    \centering
    \includegraphics[width=1.0\textwidth]{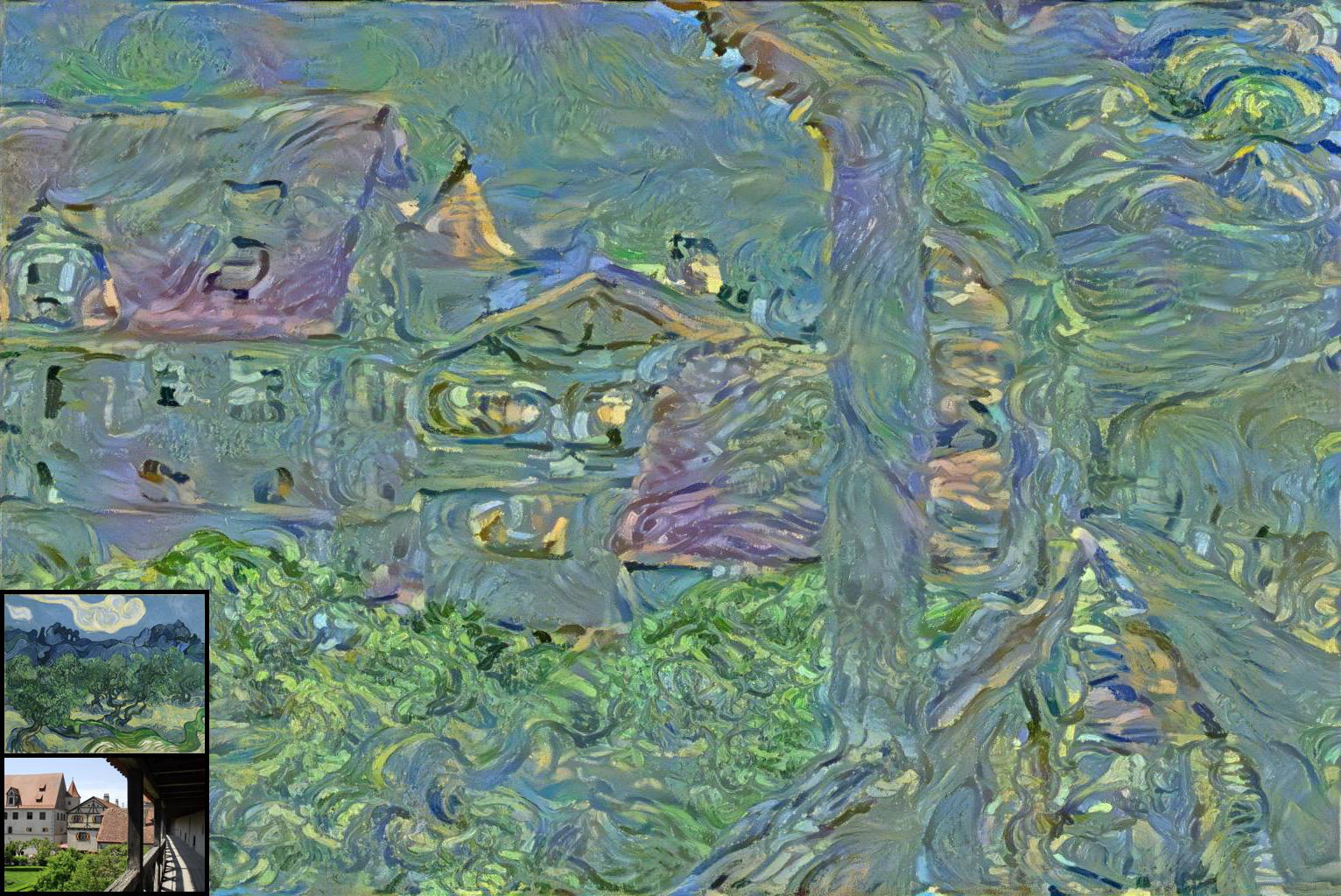}
    \caption{Style image: ``The Olive Trees'' by Vincent van Gogh.}
    \label{fig:other_house_van_gogh_olive_trees}
\end{figure*}

\begin{figure*}[!htb]
    \centering
    \includegraphics[width=1.0\textwidth]{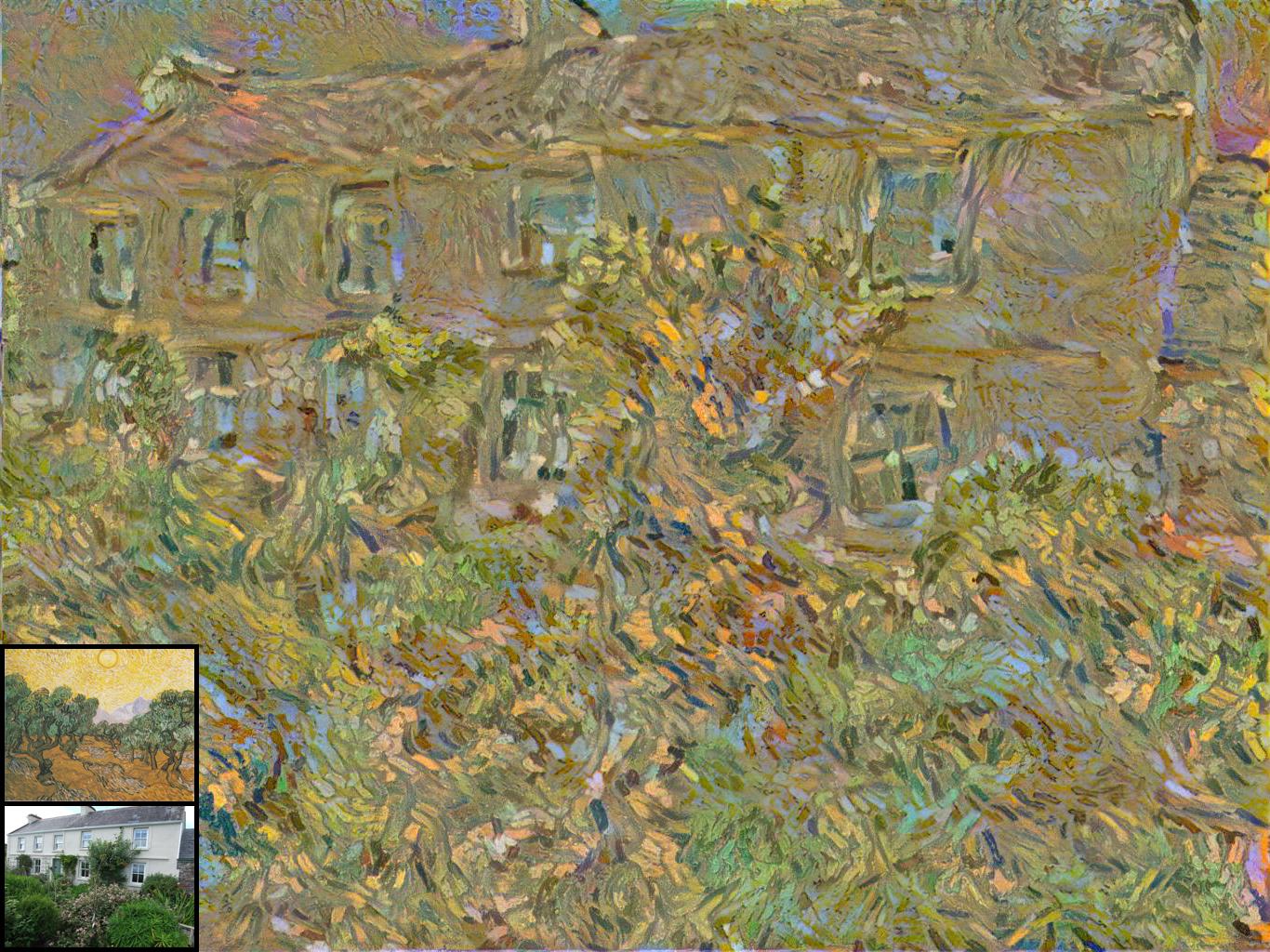}
    \caption{Style image: ``Olive Trees with Yellow Sky and Sun'' by Vincent van Gogh.}
    \label{fig:house_van_gogh_yellow_olive_trees}
\end{figure*}

\begin{figure*}[!htb]
    \centering
    \includegraphics[width=1.0\textwidth]{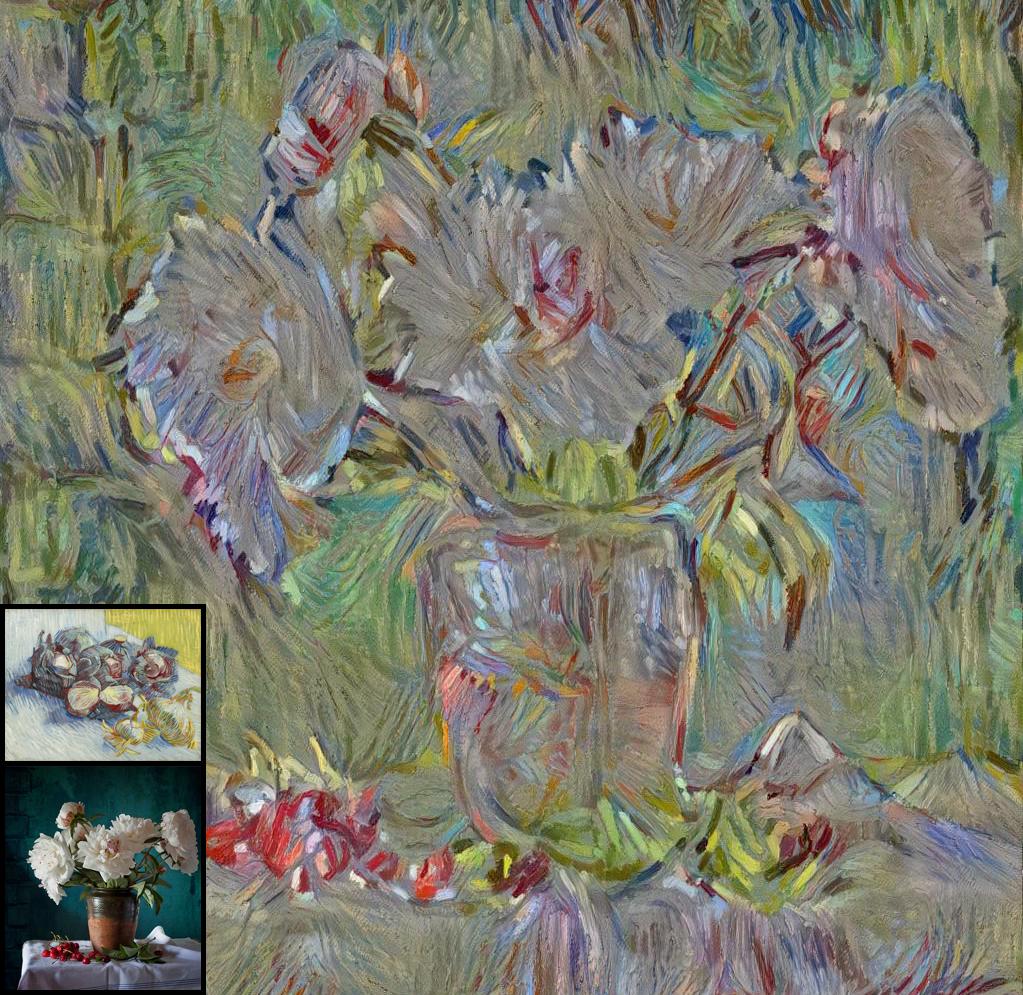}
    \caption{Style image: ``Red Cabbages and Onions'' by Vincent van Gogh.}
    \label{fig:still_life_2_red_van_gogh_cab}
\end{figure*}

We provide additional stylization examples in Fig.~\ref{fig:picasso_guy}, ~\ref{fig:peploe_rome}, ~\ref{fig:van_gogh_garden}, ~\ref{fig:signac_mountains}, ~\ref{fig:horse_kirchner}, ~\ref{fig:black_german_shepard_van_gogh}, ~\ref{fig:river_city_signac}, ~\ref{fig:german_shepard_picasso}, ~\ref{fig:tuscany_kandinsky_women}, ~\ref{fig:sailboat_vlamnick}, ~\ref{fig:still_life_red_cab_van_gogh}, ~\ref{fig:car_munch_scream}, ~\ref{fig:cabin_van_gogh_trees}, ~\ref{fig:lisbon_munch_spring}, ~\ref{fig:house_van_gogh_alyscamps}, ~\ref{fig:other_house_van_gogh_olive_trees}, ~\ref{fig:house_van_gogh_yellow_olive_trees}, ~\ref{fig:still_life_2_red_van_gogh_cab}.

\subsection{Fitting Brushstrokes to Artwork}\label{sec:fitting_artworks}
\begin{figure*}[!htb]
    \centering
    \includegraphics[width=1.0\textwidth]{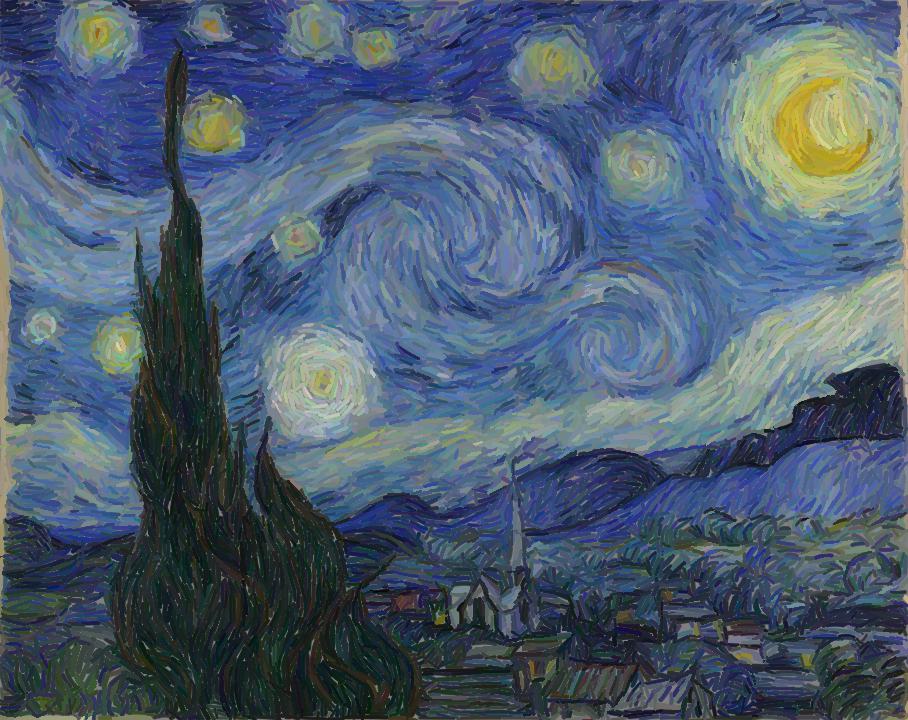} 
    \caption{Brushstroke approximation of ``Starry Night'' by Vincent van Gogh using 12.000 brushstrokes.}
    \label{fig:starry_night_approximation}
\end{figure*}

\begin{figure*}[!htb]
    \centering
    \includegraphics[width=0.9\textwidth]{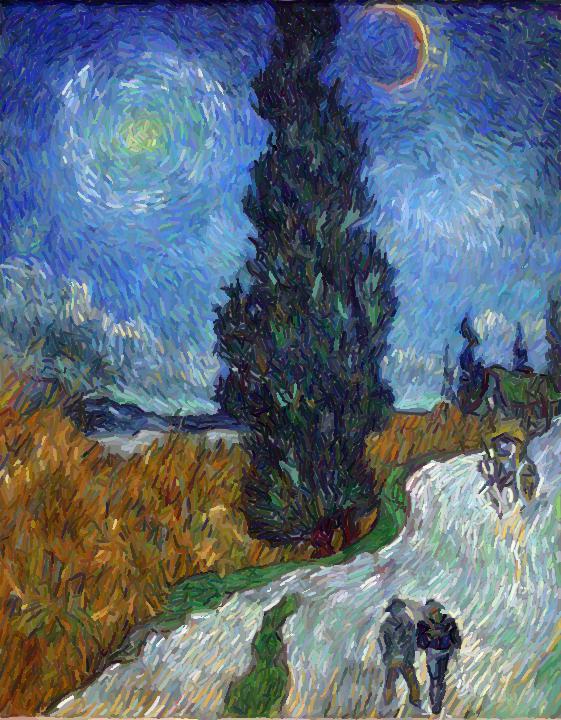} 
    \caption{Brushstroke approximation of ``Road with Cypress and Star'' by Vincent van Gogh using 12.000 brushstrokes.}
    \label{fig:cypress_approximation}
\end{figure*}

\begin{figure*}[!htb]
    \centering
    \includegraphics[width=0.9\textwidth]{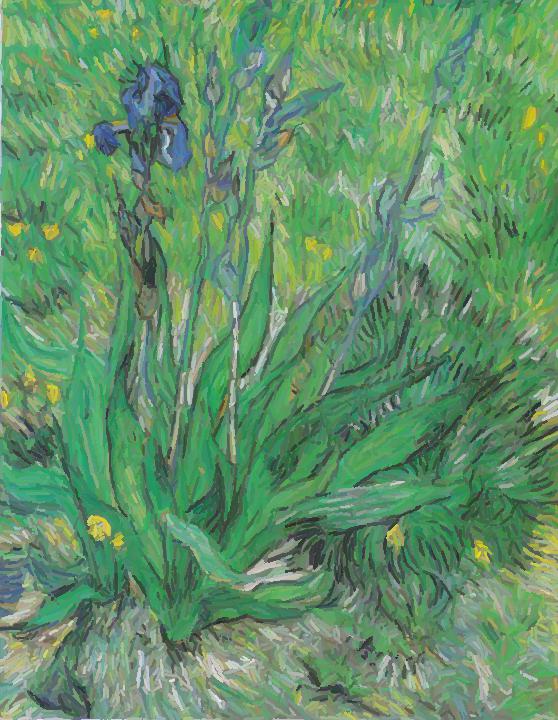} 
    \caption{Brushstroke approximation of ``Iris'' by Vincent van Gogh using 12.000 brushstrokes.}
    \label{fig:iris_approximation}
\end{figure*}

In Sec. \ref{sec:fit_to_artwork} we showed how we can use our renderer to fit brushstrokes to paintings. Note that we use SLIC superpixels \cite{slic_pami} to initialize the brushstrokes for this experiment. We only optimize the brushstroke parameters and did not apply pixel level optimization for this experiment. See Fig.~\ref{fig:starry_night_approximation}, ~\ref{fig:cypress_approximation}, ~\ref{fig:iris_approximation} for additional results.

\begin{figure}
    \centering
    \def\arraystretch{0.0}
    \setlength{\tabcolsep}{1.0pt}
    \begin{tabular}{c}
    \includegraphics[width=0.4\textwidth]{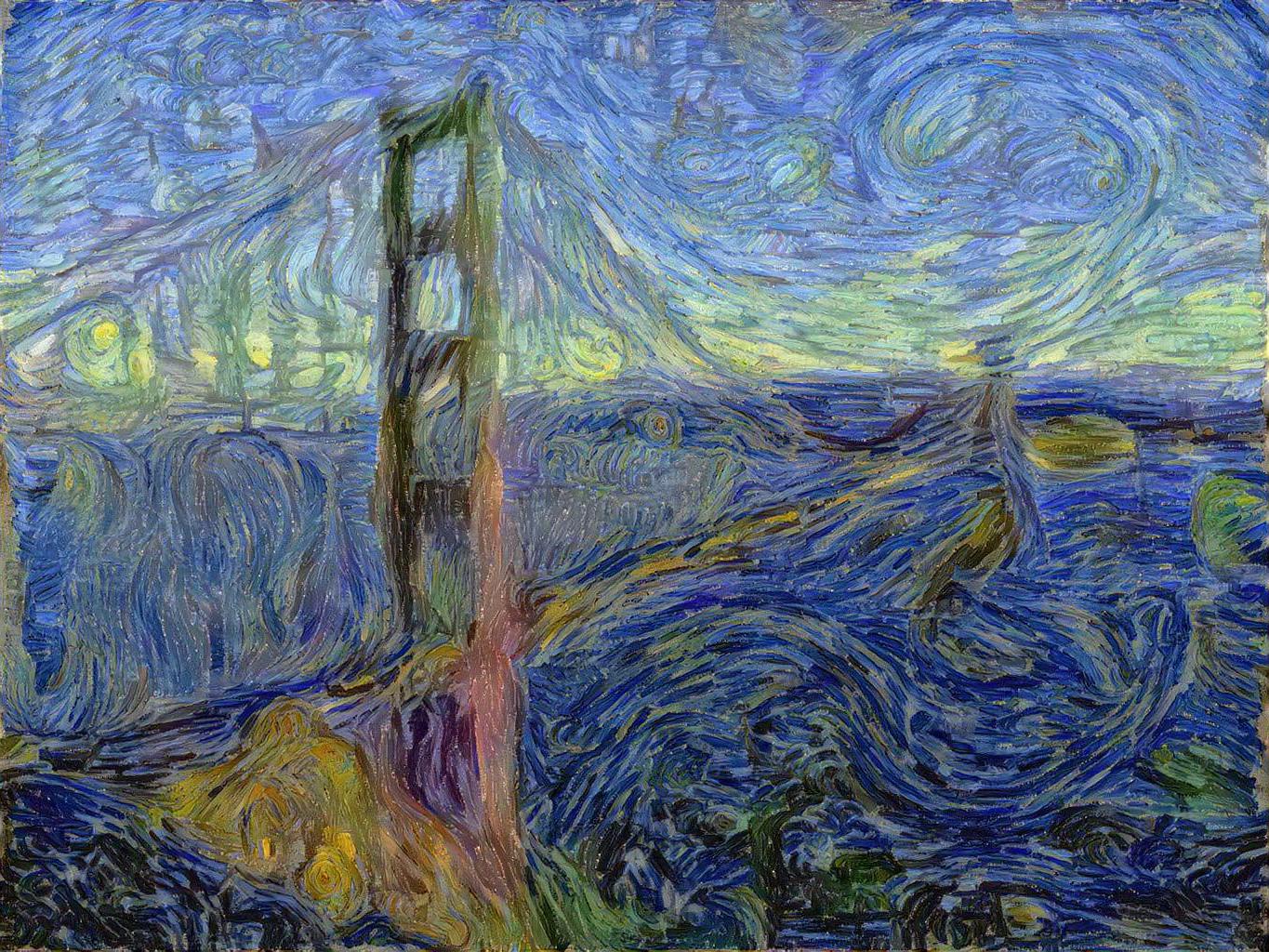} \\
    \vspace{2pt} \\
    L = 10
    \vspace{2pt} \\
    \includegraphics[width=0.4\textwidth]{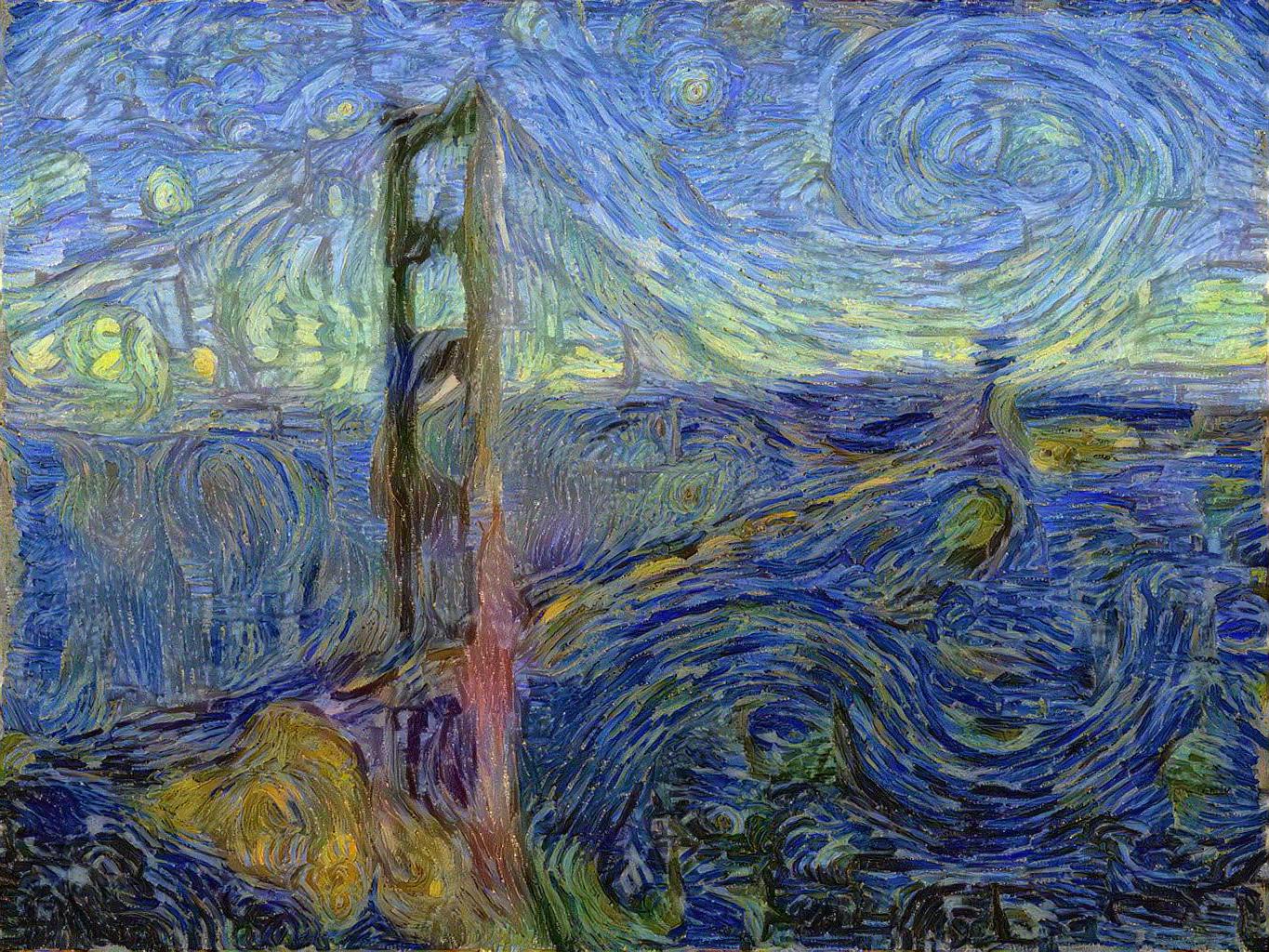} \\
    \vspace{2pt} \\
    L = 30
    \vspace{2pt} \\
    \includegraphics[width=0.4\textwidth]{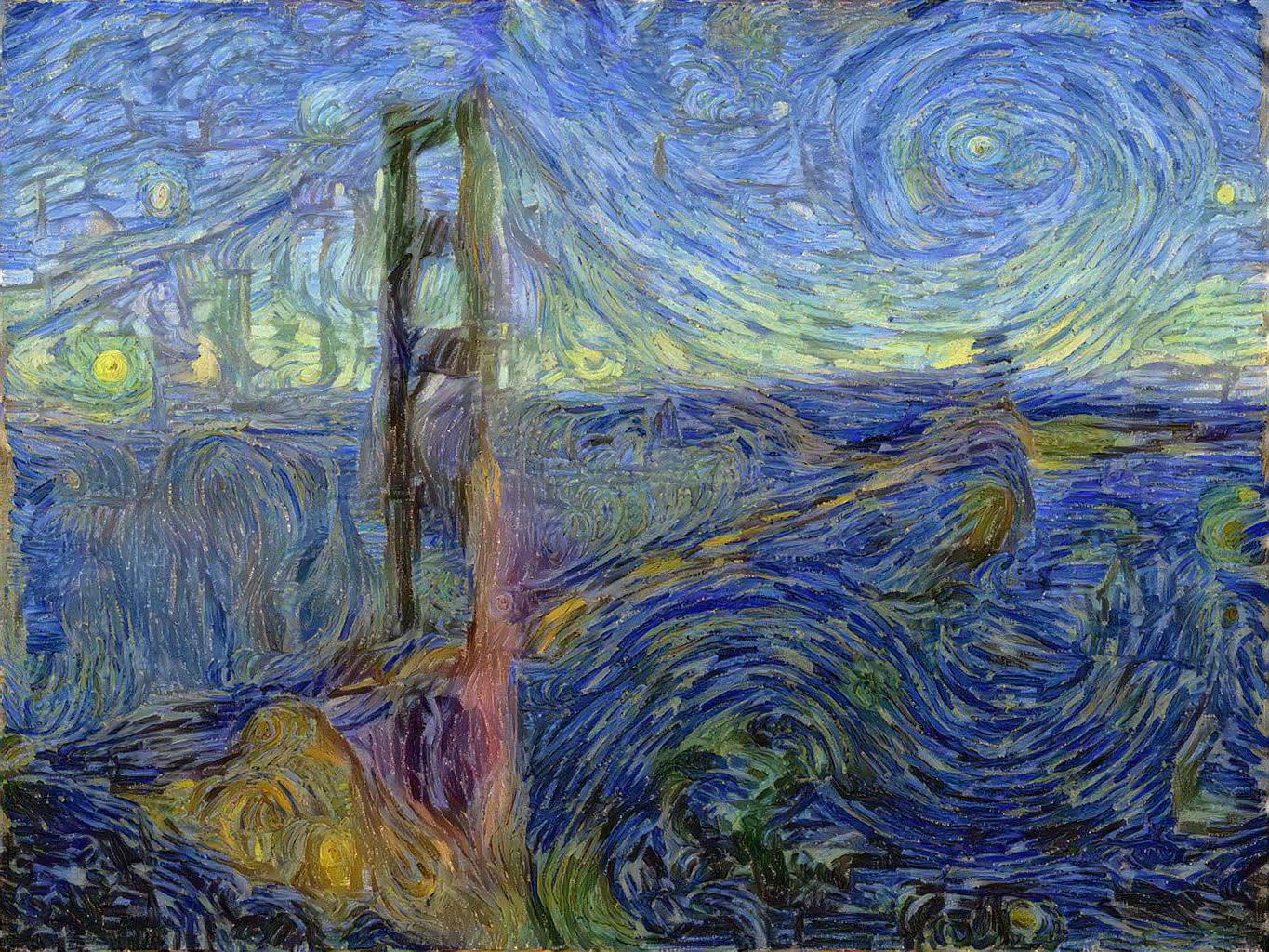} \\
    \vspace{2pt} \\
    L = 60
    \vspace{2pt} \\
    \includegraphics[width=0.4\textwidth]{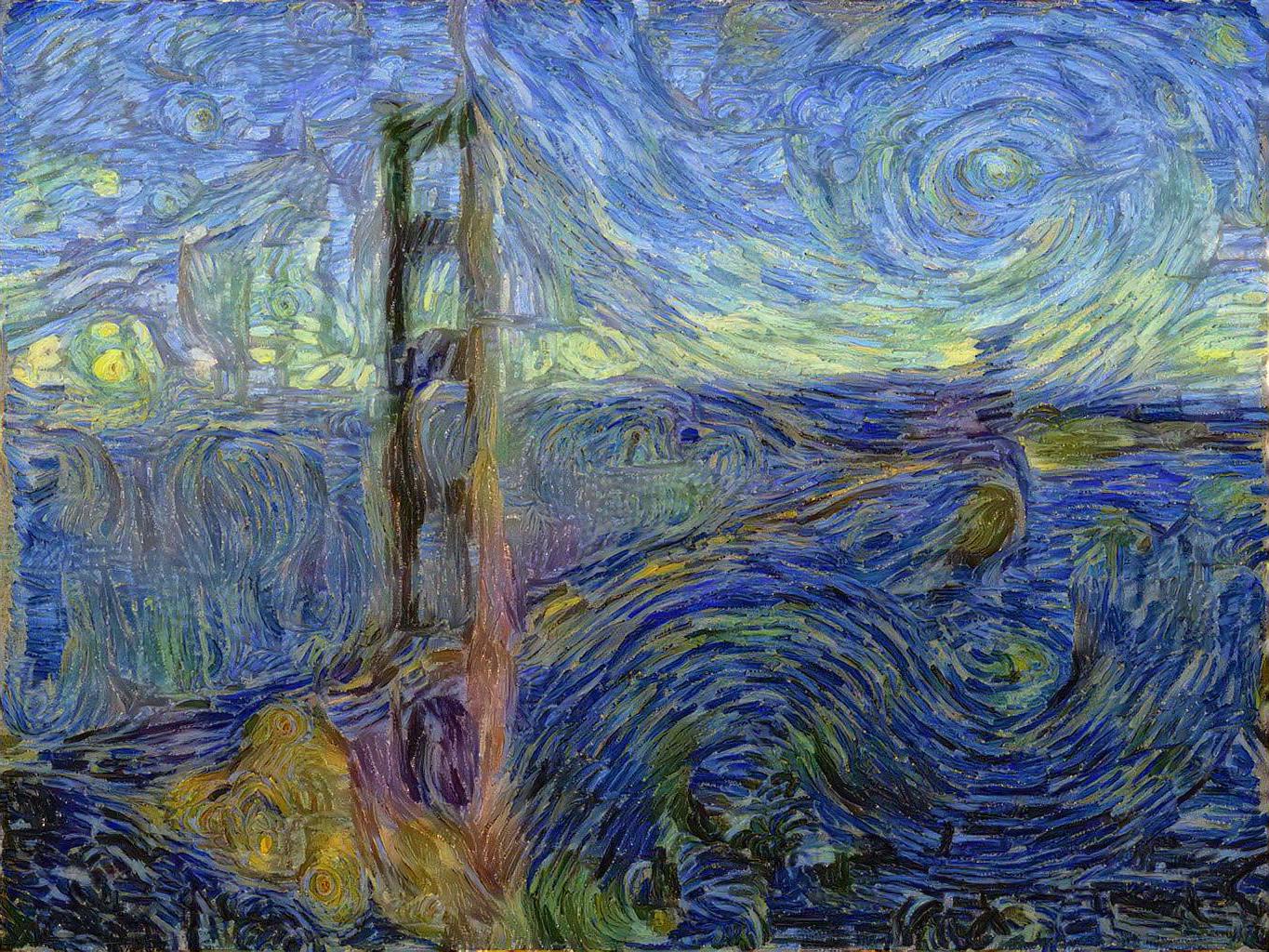} \\
    \vspace{2pt} \\
    L = 100
    \vspace{2pt} \\
    \end{tabular} \\
    
    \caption{Controlling the flow of brushstrokes with different values for $L$. The larger the value $L$ the more strokes around the user input are affected. User input is porvided in Fig. \ref{fig:user_input}.}
    \label{fig:control_strength}
\end{figure}

\begin{figure*}[!htb]
    \centering
    \def\arraystretch{0.0}
    \setlength{\tabcolsep}{1.0pt}
    \begin{tabular}{cc}
    \includegraphics[width=0.3\textwidth]{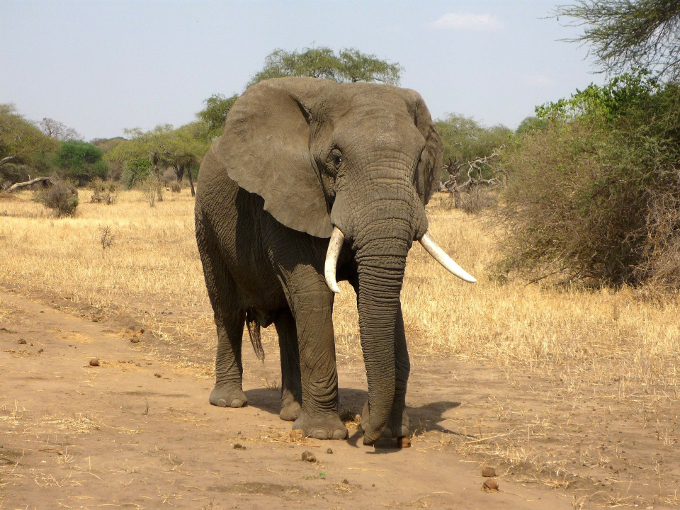} &
    \includegraphics[width=0.6\textwidth]{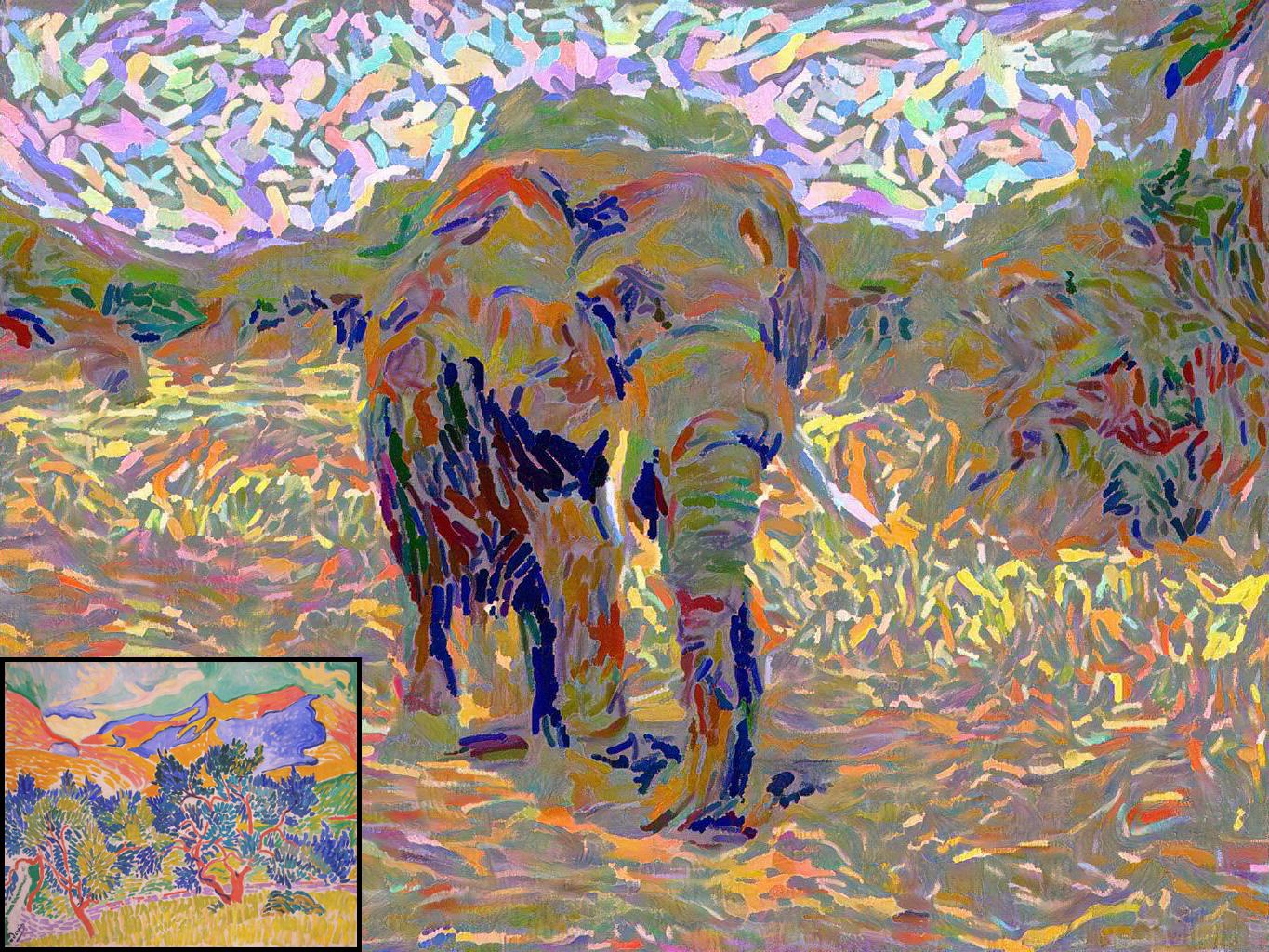} \\
    \vspace{2pt} \\
    (a) Content & (b) Stylized without user input \\
    \vspace{2pt} \\
    \includegraphics[width=0.3\textwidth]{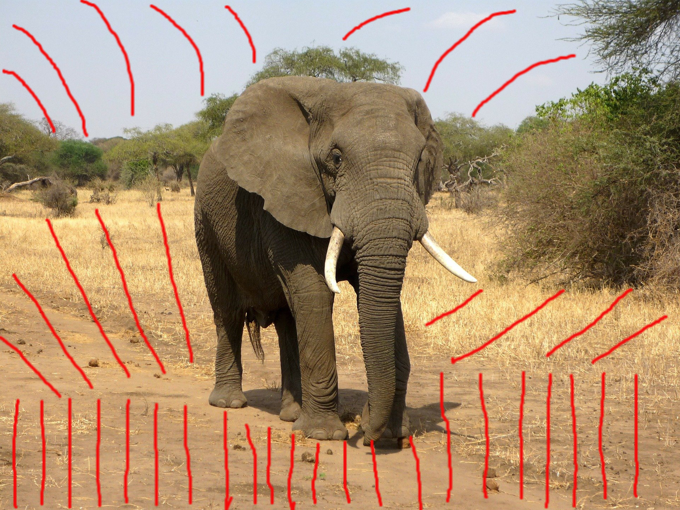} &
    \includegraphics[width=0.6\textwidth]{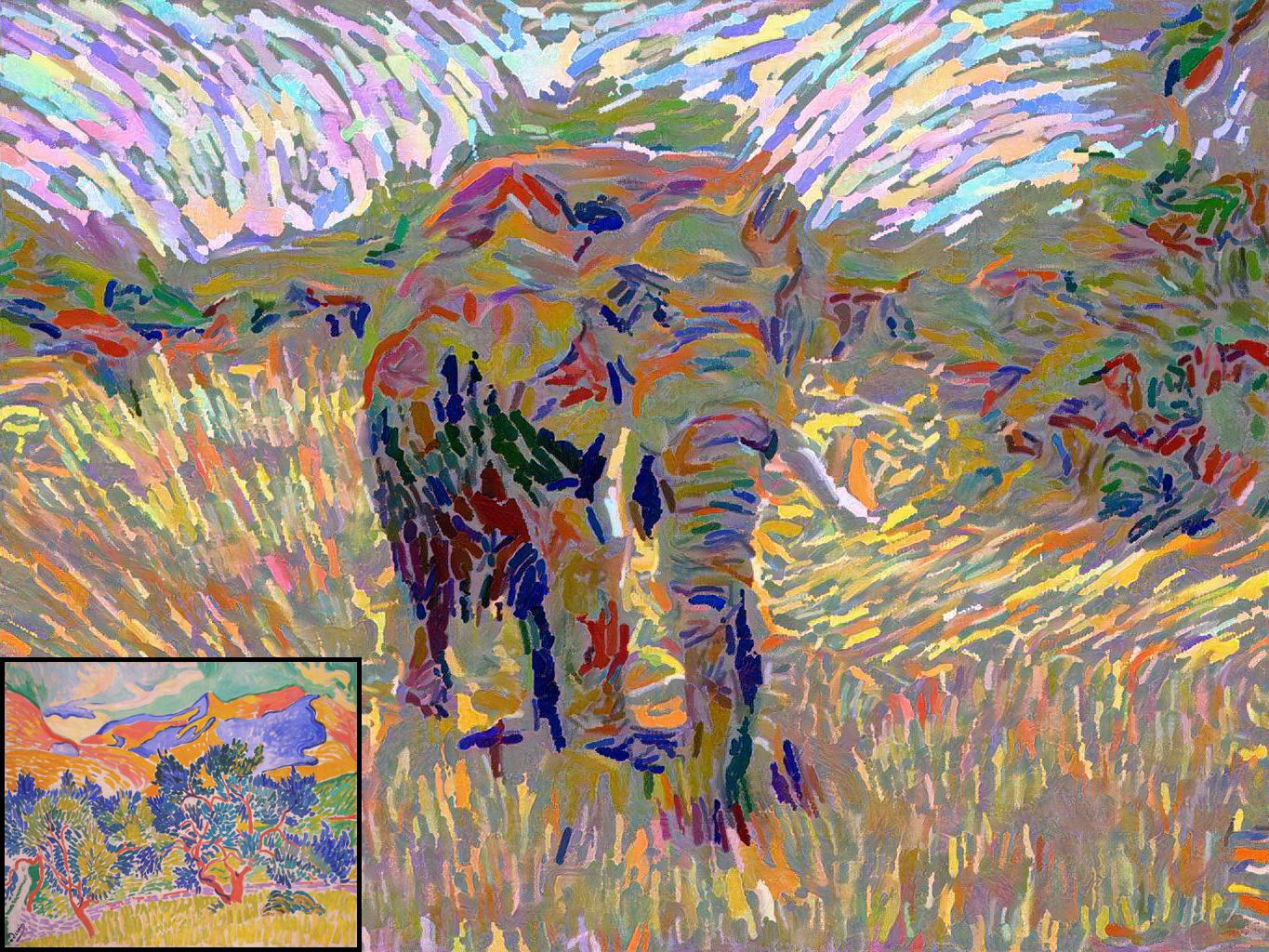} \\
    \vspace{2pt} \\
    (d) Content with user input & (e) Stylized with user input \\
    \vspace{2pt} \\
    \end{tabular} \\
    
    \caption{A user can draw curves on the content image and thus control the flow of the brushstrokes in the stylized image. Note that for the stylization with user input we also used (a) as content image. The control is imposed on the brushstroke parameters, not the pixels.}
    \label{fig:user_control_1}
\end{figure*}

\begin{figure*}[!htb]
    \centering
    \def\arraystretch{0.0}
    \setlength{\tabcolsep}{1.0pt}
    \begin{tabular}{cc}
    \includegraphics[width=0.3\textwidth]{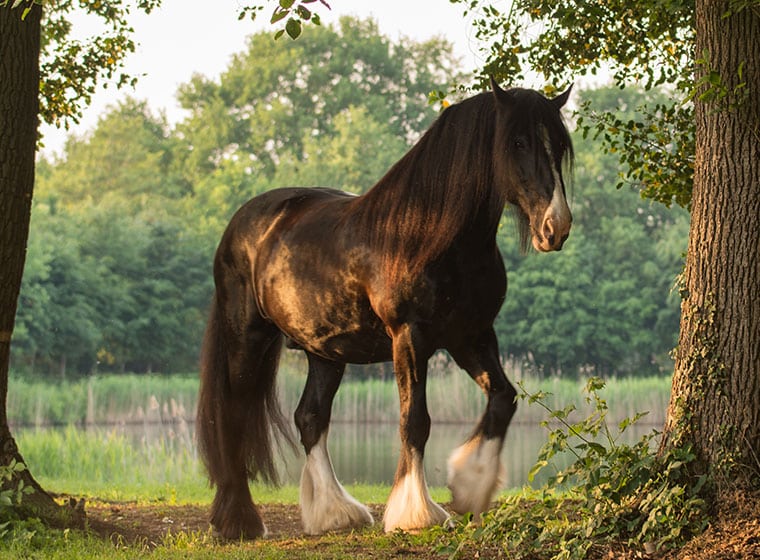} &
    \includegraphics[width=0.6\textwidth]{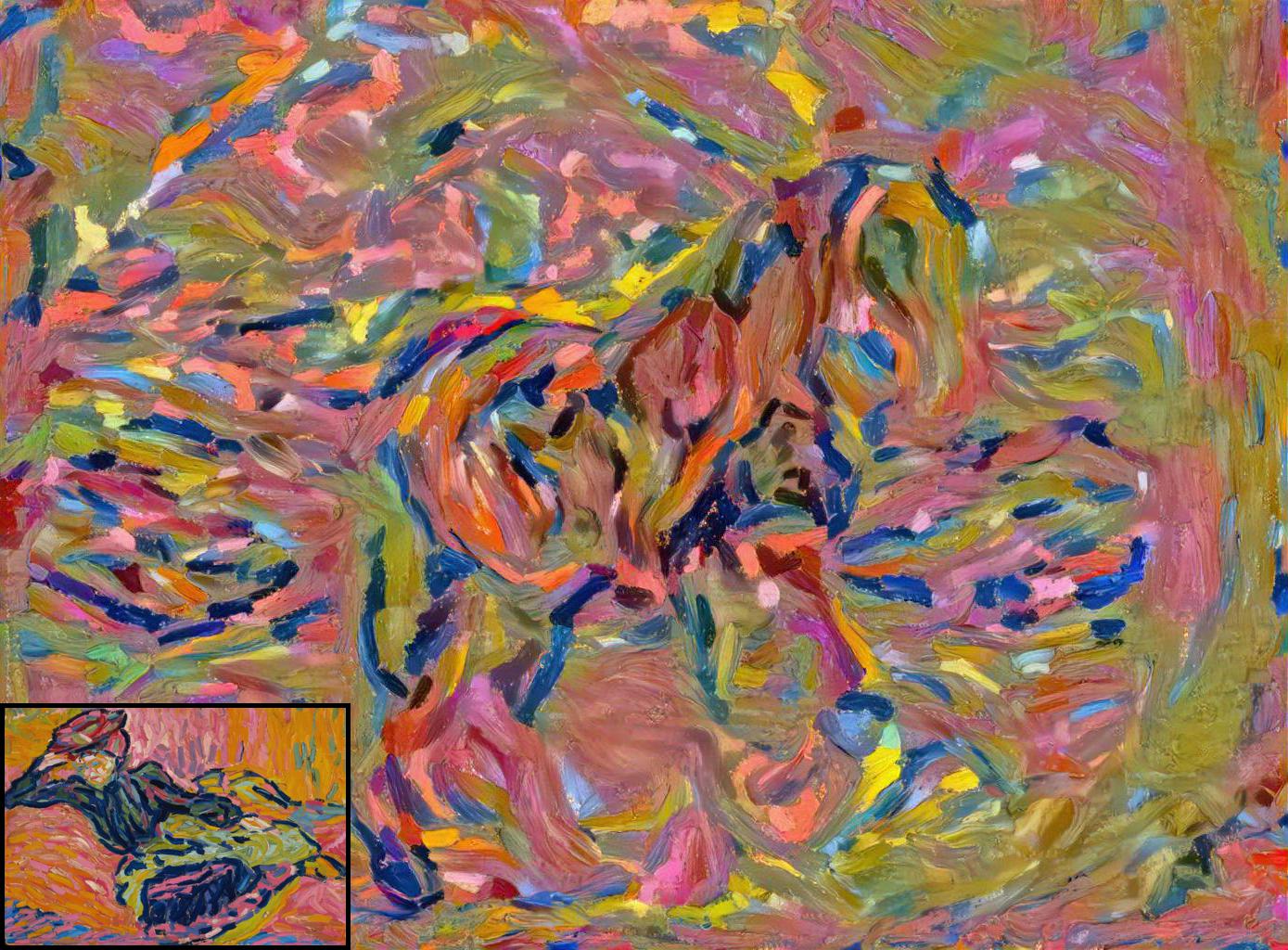} \\
    \vspace{2pt} \\
    (a) Content & (b) Stylized without user input \\
    \vspace{2pt} \\
    \includegraphics[width=0.3\textwidth]{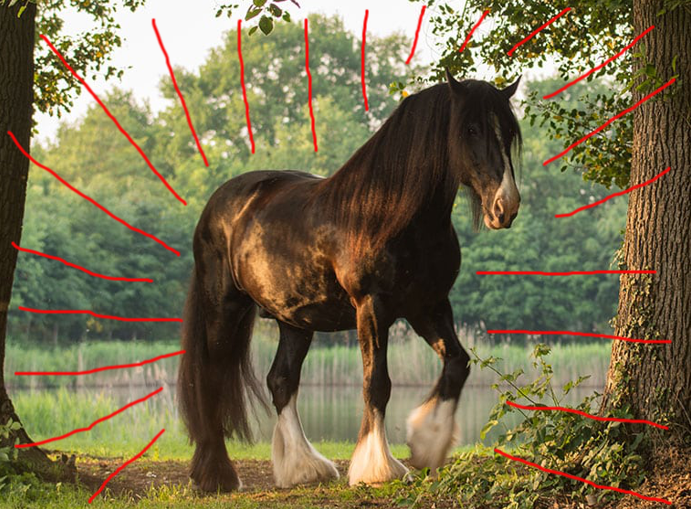} &
    \includegraphics[width=0.6\textwidth]{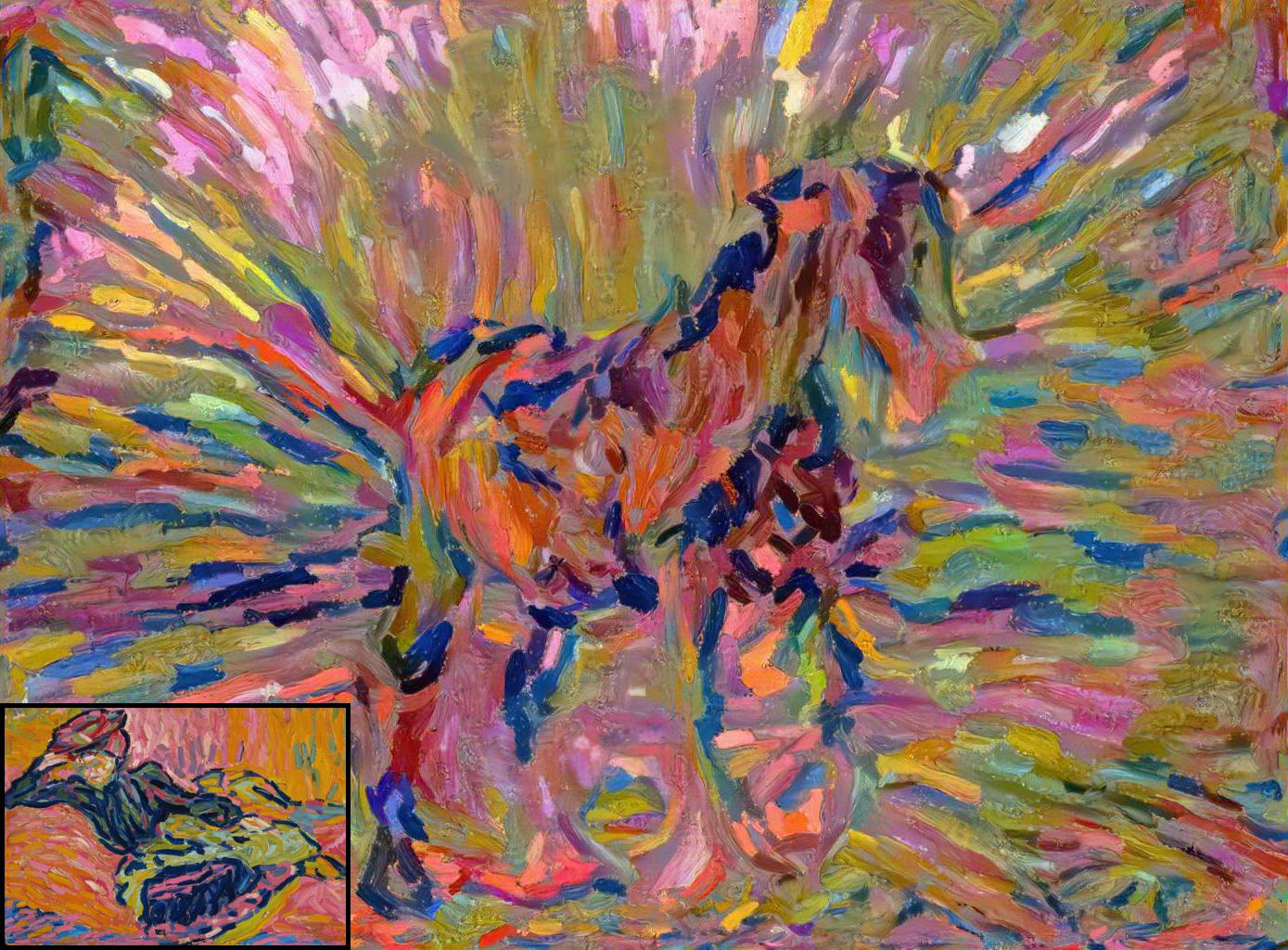} \\
    \vspace{2pt} \\
    (d) Content with user input & (e) Stylized with user input \\
    \vspace{2pt} \\
    \end{tabular} \\
    
    \caption{A user can draw curves on the content image and thus control the flow of the brushstrokes in the stylized image. Note that for the stylization with user input we also used (a) as content image. The control is imposed on the brushstroke parameters, not the pixels.}
    \label{fig:user_control_2}
\end{figure*}

\begin{figure*}[!htb]
    \centering
    \includegraphics[width=0.6\textwidth]{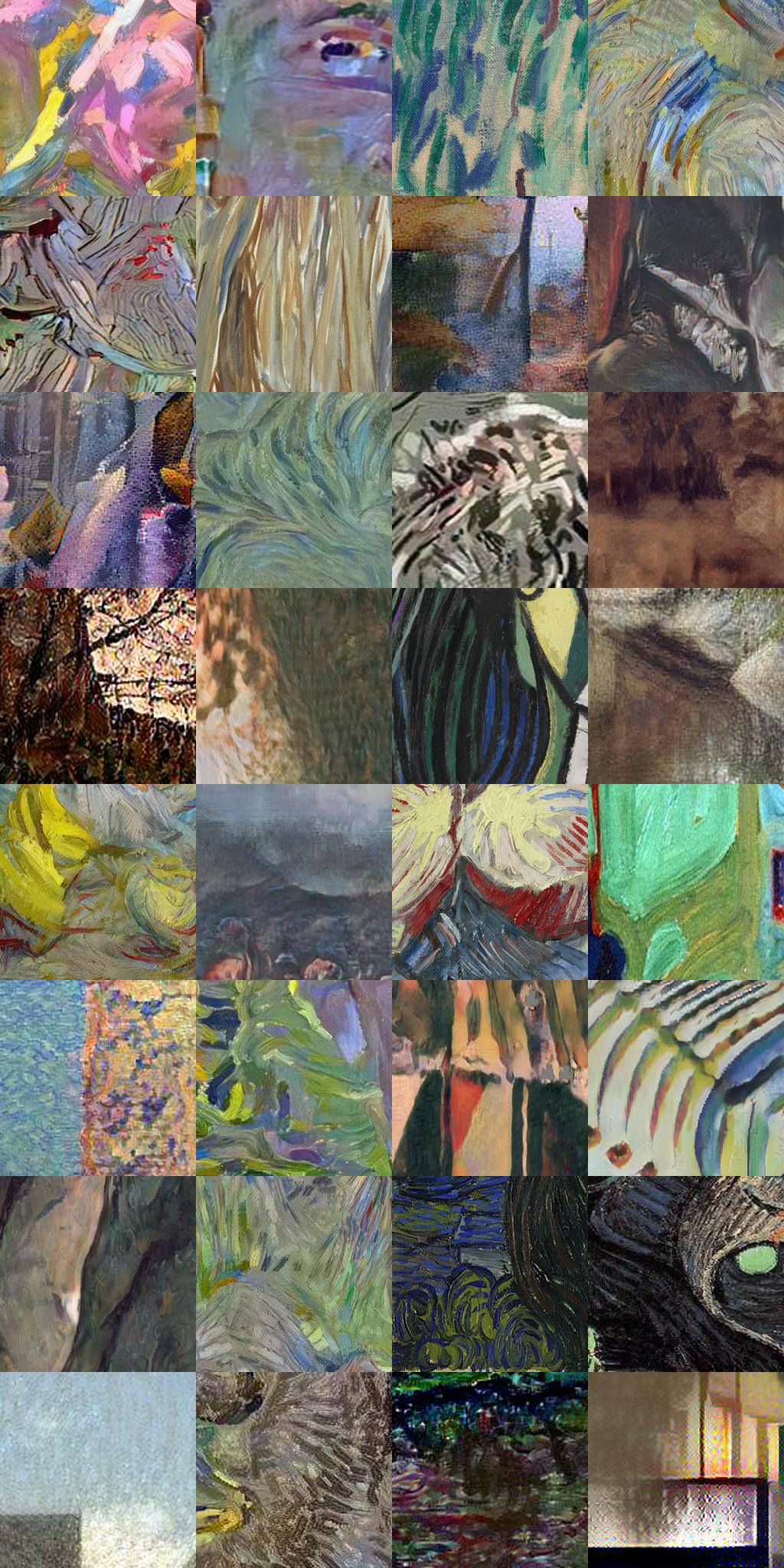}
    \caption{Randomly sampled patches from the user study. In each round we show one row and ask the users to mark all the patches cropped out of real artworks. Each crop in a row is drawn from either a real artwork (`real'), a stylization by Gatys et al. \cite{Gatys2016} (`Gatys'), a stylization by Sanakoyeu et al. \cite{Sanakoyeu2018} (`AST'), or from our method (`ours'). In this table we have restricted ourselves to only those 4 classes to make this quiz more difficult for the reader. Try to guess which are real. The answers are on the last page.}
    \label{fig:user_study_random_patches}
\end{figure*}

\FloatBarrier
\clearpage

\let\thefootnote\relax\footnotetext{Solution to Fig.~\ref{fig:user_study_random_patches}: \\
ours, ours, real, ours  \\
ours, real, Gatys, AST  \\
Gatys, ours, ours, AST  \\
Gatys, AST, real, Gatys  \\
ours, AST, real, real  \\
Gatys, ours, AST, AST  \\
AST, ours, real, real  \\
Gatys, ours, Gatys, Gatys }


\end{document}